\documentclass[11pt]{article}
\usepackage{smile}
\usepackage{natbib}
\usepackage{multirow}
\usepackage{algorithm}
\usepackage{color}

\usepackage{fullpage}

\usepackage{kpfonts}
\DeclareMathAlphabet{\mathsf}{OT1}{cmss}{m}{n}
\SetMathAlphabet{\mathsf}{bold}{OT1}{cmss}{bx}{n}

\newcommand{\sgn}{\mathop{\mathrm{sign}}}

\providecommand{\norm}[1]{\|#1\|}

\providecommand{\norm}[1]{\|#1\|}

\newcommand{\prox}{\mathrm{prox}}
\newcommand*{\Mc}{\cM_{\perp}}

\newcommand*{\Sc}{\overline{\cS}}

\newcommand*{\Ic}{\overline{\cI}}

\newcolumntype{L}[1]{>{\raggedright\arraybackslash}p{#1}}
\newcolumntype{C}[1]{>{\centering\arraybackslash}p{#1}}
\newcolumntype{R}[1]{>{\raggedleft\arraybackslash}p{#1}}

\begin{document}
	
	\title{\huge \bf On Fast Convergence of Proximal Algorithms for SQRT-Lasso Optimization: Don't Worry About its Nonsmooth Loss Function\thanks{Xingguo Li, Jarvis Haupt and Mingyi Hong are affiliated with Department of Electrical and Computer Engineering at University of Minnesota; Raman Arora is affiliated with Department of Computer Science at Johns Hopkins University; Han Liu is affiliated with Department of Electrical Engineering and Computer Science at Northwestern University; Haoming Jiang and Tuo Zhao is affiliated with School of Industrial and Systems Engineering at Georgia Institute of Technology; Tuo Zhao is the corresponding author. Emails: {\tt lixx1661@umn.edu, tourzhao@gatech.edu}. }}

	\author{Xingguo Li, Haoming Jiang, Jarvis Haupt, Raman Arora,\\ Han Liu, Mingyi Hong and Tuo Zhao}
	
	\date{}
	
	\maketitle
	

\begin{abstract}
Many machine learning techniques sacrifice convenient computational structures to gain estimation robustness and modeling flexibility. However, by exploring the modeling structures, we find these ``sacrifices'' do not always require more computational efforts. To shed light on such a ``free-lunch'' phenomenon, we study the square-root-Lasso (SQRT-Lasso) type regression problem. Specifically, we show that the nonsmooth loss functions of SQRT-Lasso type regression ease tuning effort and gain adaptivity to inhomogeneous noise, but is not necessarily more challenging than Lasso in computation. We can directly apply proximal algorithms (e.g. proximal gradient descent, proximal Newton, and proximal quasi-Newton algorithms) without worrying about the nonsmoothness of the loss function. Theoretically, we prove that the proximal algorithms enjoy fast local convergence with high probability. 
Our numerical experiments also show that when further combined with the pathwise optimization scheme, the proximal algorithms significantly outperform other competing algorithms.

\end{abstract}
	
\vspace{-0.25in}
\section{Introduction}
\vspace{-0.1in}

Many statistical machine learning methods can be formulated as optimization problems in the following form
\begin{align}\label{general-form}
\min_{\theta}\cL(\theta) + \cR(\theta),
\end{align}
where $\cL(\theta)$ is a loss function and $\cR(\theta)$ is a regularizer. When the loss function is smooth and has a Lipschitz continuous gradient, \eqref{general-form} can be efficiently solved by simple proximal gradient descent and proximal Newton algorithms (also requires a Lipschitz continuous Hessian matrix of $\cL(\theta)$). Some statistical machine learning methods, however, sacrifice convenient computational structures to gain estimation robustness and modeling flexibility \cite{wang2013l1,belloni2011square,liu2015calibrated}. Taking SVM as an example, the hinge loss function gains estimation robustness, but sacrifices the smoothness (compared with the square hinge loss function). However, by exploring the structure of the problem, we find that these ``sacrifices" do not always require more computational efforts.


\noindent{\bf Advantage of SQRT-Lasso over Lasso.} To shed light on such a ``free-lunch" phenomenon, we study the high dimensional square-root (SQRT) Lasso regression problem \cite{belloni2011square,sun2012scaled}. 
Specifically, we consider a sparse linear model in high dimensions,
\begin{align*}
 y =  X \theta^* + \epsilon,
\end{align*}
where $ X \in \RR^{n \times d}$ is the design matrix, $ y\in\RR^n$ is the response vector, $\epsilon\sim N( 0,\sigma^2 I_n)$ is the random noise, and $\theta^*$ is the sparse unknown regression coefficient vector. To estimate $\theta^*$, \cite{tibshirani1996regression} propose the well-known Lasso estimator by solving
\begin{align}\label{lasso-formulation}
\overline{\theta}^{\sf Lasso} = \argmin_{\theta} \frac{1}{n} \norm{ y- X\theta}_2^2 + \lambda_{\sf Lasso}\norm{\theta}_1,
\end{align}
where $\lambda_{\sf Lasso}$ is the regularization parameter. Existing literature shows that given
\begin{align}\label{lasso-tuning}
\lambda_{\sf Lasso} \asymp {\sigma}\sqrt{\frac{\log d}{n}},
\end{align}
$\overline{\theta}^{\sf Lasso}$ is minimax optimal for parameter estimation in high dimensions. Note that the optimal regularization parameter for Lasso in \eqref{lasso-tuning}, however, requires the prior knowledge of the unknown parameter $\sigma$. This requires the regularization parameter to be carefully tuned over a wide range of potential values to get a good finite-sample performance.

To overcome this drawback, \cite{belloni2011square} propose the SQRT-Lasso estimator by solving
\begin{align}\label{sqrt-lasso-formulation}
\overline{\theta}^{\sf SQRT} = \argmin_{\theta\in\RR^d} \frac{1}{\sqrt{n}} \norm{ y- X\theta}_2 + \lambda_{\sf SQRT}\norm{\theta}_1,
\end{align}
where $\lambda_{\sf SQRT}$ is the regularization parameter. They further show that $\overline{\theta}^{\sf SQRT}$ is also minimax optimal
in parameter estimation, but the optimal regularization parameter is
\begin{align}\label{sqrt-tuning}
\lambda_{\sf SQRT} \asymp {\sqrt{\frac{\log d}{n}}}.
\end{align}
Since \eqref{sqrt-tuning} no longer depends on $\sigma$, SQRT-Lasso eases tuning effort.


\noindent{\bf Extensions of SQRT-Lasso.} Besides the tuning advantage, the regularization selection for SQRT-Lasso type methods is also adaptive to inhomogeneous noise. For example, \cite{liu2015calibrated} propose a multivariate SQRT-Lasso for sparse multitask learning. Given a matrix $ A \in \RR^{d \times d}$, let $ A_{*k}$ denote the $k$-th column of $ A$, and $ A_{i*}$ denote the $i$-th row of $ A$. Specifically, \cite{liu2015calibrated} consider a multitask regression model
\begin{align*}
 Y =  X\Theta^*+ W,
\end{align*}
where $ X \in \RR^{n \times d}$ is the design matrix, $ Y\in\RR^{n\times m}$ is the response matrix, $ W_{*k}\sim N( 0,\sigma_k^2 I_n)$ is the random noise, and $\Theta^*\in\RR^{d\times m}$ is the unknown row-wise sparse coefficient matrix, i.e., $\Theta^*$ has many rows with all zero entries. To estimate $\Theta^*$, \cite{liu2015calibrated} propose a calibrated multivariate regression (CMR) estimator by solving
\begin{align*}
\overline{\theta}^{\sf CMR} = \argmin_{\theta\in\RR^{d\times m}} \frac{1}{\sqrt{n}} \sum_{k=1}^m\norm{ Y_{*k}- X\Theta_{*k}}_2 + \lambda_{\sf CMR}\norm{\Theta}_{1,2},
\end{align*}
where $\norm{\Theta}_{1,2}=\sum_{j=1}^d\norm{\Theta_{j*}}_{2}$. \cite{liu2015calibrated} further shows that the regularization of CMR approach is adaptive to $\sigma_k$'s for each regression task, i.e., $ Y_{*k} =  X\Theta^*_{*k}+ W_{*k}$, and therefore CMR achieves better performance in parameter estimation and variable selection than its least square loss based counterpart. With a similar motivation, \cite{liu2017tiger} propose a node-wise SQRT-Lasso approach for sparse precision matrix estimation. Due to space limit, please refer to \cite{liu2017tiger} for more details.

\noindent{\bf Existing Algorithms for SQRT-Lasso Optimization.} Despite of these good properties, in terms of optimization, \eqref{sqrt-lasso-formulation} for SQRT-Lasso is computationally more challenging than \eqref{lasso-formulation} for Lasso. The $\ell_2$ loss in \eqref{sqrt-lasso-formulation} is not necessarily differentiable, and does not have a Lipschitz continuous gradient, compared with the least square loss in \eqref{lasso-formulation}. A few algorithms have been proposed for solving \eqref{sqrt-lasso-formulation} in existing literature, but none of them are satisfactory when $n$ and $d$ are large. \cite{belloni2011square} reformulate \eqref{sqrt-lasso-formulation} as a second order cone program (SOCP) and solve by an interior point method with a computational cost of $\cO(nd^{3.5}\log(\epsilon^{-1}))$, where $\epsilon$ is a pre-specified optimization accuracy; \cite{li2015flare} solve \eqref{sqrt-lasso-formulation} by an alternating direction method of multipliers (ADMM) algorithm with a computational cost of $\cO(nd^2/\epsilon)$; \cite{sun2012scaled} propose to solve the variational form of \eqref{sqrt-lasso-formulation} by an alternating minimization algorithm, and \cite{ndiaye2016efficient} further develop a coordinate descent subroutine to accelerate its computation. However, no iteration complexity is established in \cite{ndiaye2016efficient}. Our numerical study shows that their algorithm only scales to moderate problems. Moreover, \cite{ndiaye2016efficient} require a good initial guess for the lower bound of $\sigma$. When the initial guess is inaccurate, the empirical convergence can be slow.

\begin{table*}[htb!] 
		\label{summary}
		\caption{Comparison with existing algorithms for solving SQRT-Lasso. SOCP: Second-order Cone Programming; TRM: Trust Region Newton; VAM: Variational Alternating Minimization; ADMM: Alternating Direction Method of Multipliers; VCD: Coordinate Descent; Prox-GD: Proximal Gradient Descent; Prox-Newton: Proximal Newton.}
		{\small
		\begin{center}
			\begin{tabular}{lccclc}
				\toprule
			&Algorithm	 &Theoretical Guarantee &Empirical Performance\\
			\midrule
			\cite{belloni2011square} &SOCP + TRM	&$\cO(nd^{3.5}\log(\epsilon^{-1}))$ &Very Slow\\
			\midrule
			\cite{sun2012scaled} &VAM	&N.A. &Very Slow\\
			\midrule
			\cite{li2015flare} &ADMM	&$\cO(nd^2/\epsilon)$ &Slow\\
			\midrule
			\cite{ndiaye2016efficient}	&VAM + CD&N.A.  &Moderate\\
			\midrule
			This paper	&Pathwise Prox-GD &$\cO(nd\log(\epsilon^{-1}))$  &Fast\\
			\midrule
			This paper	&Pathwise Prox-Newton + CD &$\cO(snd\log\log(\epsilon^{-1}))$  &Very Fast\\
			\bottomrule
			\end{tabular}
			\end{center}
			}
			{\bf Remark}: \cite{ndiaye2016efficient} requires a good initial guess of $\sigma$ to achieve moderate performance. Otherwise, its empirical performance is similar to ADMM.
	\end{table*}

\begin{figure}[t]
	\centering
	\includegraphics[width=1\linewidth]{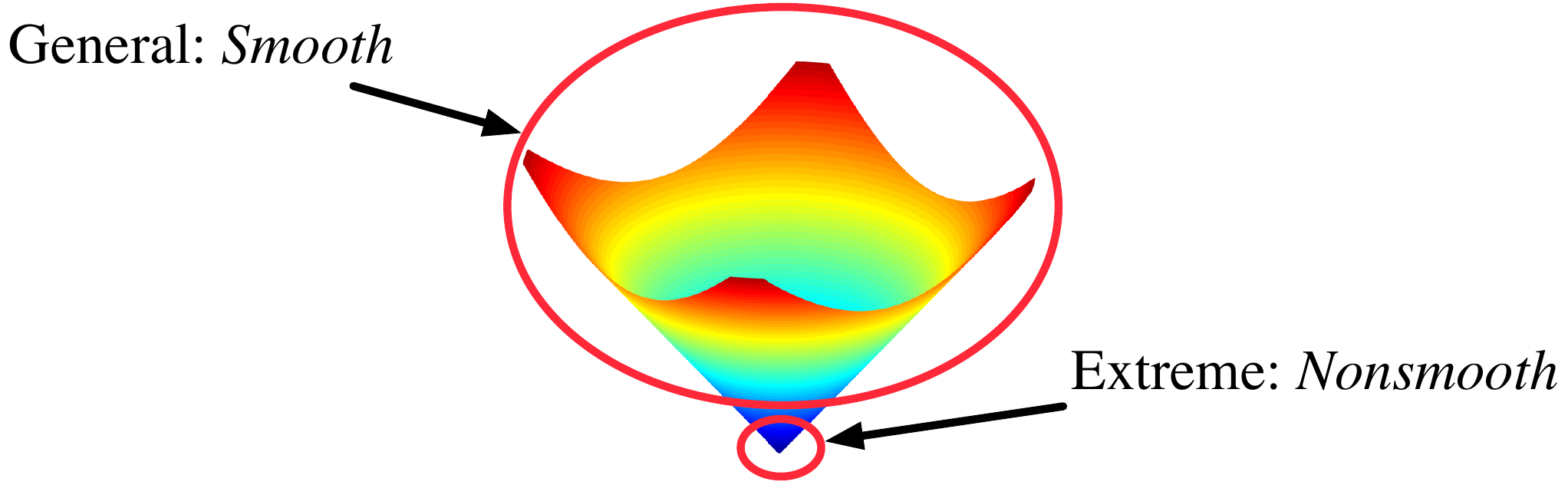}
	\caption{The extreme and general cases of the $\ell_2$ loss. The nonsmooth region $\cbr{\theta: y -  X \theta =  0}$ is out of our interest, since it corresponds to those overfitted regression models}\label{fig:graph}
\end{figure}

\noindent{\bf Our Motivations.} The major drawback of the aforementioned algorithms is that they do not explore the modeling structure of the problem. The $\ell_2$ loss function is not differentiable only when the model are overfitted, i.e., the residuals are zero values $ y- X\theta= 0$. Such an extreme scenario rarely happens in practice, especially when SQRT-Lasso is equipped with a sufficiently large regularization parameter $\lambda_{\sf SQRT}$ to yield a sparse solution and prevent overfitting. Thus, we can treat the $\ell_2$ loss as an ``almost" smooth function. Moreover, our theoretical investigation indicates that the $\ell_2$ loss function also enjoys the restricted strong convexity, smoothness, and Hessian smoothness. In other words, the $\ell_2$ loss function behaves as a strongly convex and smooth over a sparse domain. An illustration is provided in Figure~\ref{fig:graph}. 

\noindent{\bf Our Contributions.} Given these nice geometric properties of the $\ell_2$ loss function, we can directly solve \eqref{sqrt-lasso-formulation} by proximal gradient descent (Prox-GD), proximal Newton (Prox-Newton), and proximal Quasi-Newton (Prox-Quasi-Newton) algorithms \citep{nesterov2013gradient,lee2014proximal}. Existing literature only apply these algorithms to solve optimization problems in statistical machine learning when the loss function is smooth.
Our theoretical analysis shows that both algorithms enjoy fast convergence. Specifically, the Prox-GD algorithm achieves a local linear convergence and the Prox-Newton algorithm achieves a local quadratic convergence. The computational performance of these two algorithms can be further boosted in practice, when combined with the pathwise optimization scheme. Specifically, the pathwise optimization scheme solves \eqref{sqrt-lasso-formulation} with a decreasing sequence of regularization parameters, $\lambda_0\geq \ldots \geq \lambda_N$ with $\lambda_N=\lambda_{\sf SQRT}$. The pathwise optimization scheme helps yield sparse solutions and avoid overfitting throughout all iterations. Therefore, the nonsmooth loss function is differentiable. Besides sparse linear regression, we extend our algorithms and theory to sparse multitask regression and sparse precision matrix estimation. Extensive numerical results show our algorithms uniformly outperform the competing algorithms.


\noindent{\bf Hardness of Analysis.} We highlight that our local analysis with strong convergence guarantees are novel and highly nontrivial for solving the SQRT-Lasso problem using simple and efficient proximal algorithms. First of all, sophisticated analysis is required to demonstrate the restricted strong convexity/smoothness and Hessian smoothness of the $\ell_2$ loss function over a neighborhood of the underlying model parameter $\theta^*$ in high dimensions. These are key properties for establishing the strong convergence rates of proximal algorithms. Moreover, it is involved to guarantee that the output solution of the proximal algorithms do not fall in the nonsmooth region of the $\ell_2$ loss function. This is important in guaranteeing the favored computational and statistical properties. In addition, it is highly technical to show that the pathwise optimization does enter the strong convergence region at certain stage. We defer all detailed analysis to the appendix.




\noindent\textbf{Notations.} Given a vector $v \in \RR^{d}$, we define the subvector of $v$ with the $j$-th entry removed as $v_{\backslash j} \in \RR^{d-1}$. Given an index set $\cI \subseteq \{1,...,d\}$, let $\Ic$ be the complementary set to $\cI$ and $v_{\cI}$ be a subvector of $v$ by extracting all entries of $v$ with indices in $\cI$. Given a matrix $A\in\RR^{d \times d}$, we denote $A_{*j} $ ($A_{k*}$) the $j$-th column ($k$-th row), $A_{\backslash i\backslash j}$ as a submatrix of $A$ with the $i$-th row and the $j$-th column removed and $A_{\backslash i j}$ ($A_{i\backslash j}$) as the $j$-th column ($i$-th row) of $A$ with its $i$-th entry ($j$-th entry) removed. Let $\Lambda_{\max}(A)$ and $\Lambda_{\min}(A)$ be the largest and smallest eigenvalues of $A$ respectively. Given an index set $\cI \subseteq \{1,...,d\}$, we use $A_{\cI\cI}$ to denote a submatrix of $A$ by extracting all entries of $A$ with both row and column indices in $\cI$. We denote $A \succ 0$ if $A$ is a positive-definite matrix. 
Given two real sequences $\{A_n\},\{a_n\}$, we use conventional notations $A_n = \cO(a_n)$ (or $A_n =\Omega(a_n)$) denote the limiting behavior, ignoring constant, $\tilde{\cO}$ to denote  limiting behavior further ignoring logarithmic factors, and $\cO_P(\cdot)$ to denote the limiting behavior in probability. $A_n \asymp a_n$ if $A_n = \cO(a_n)$ and $A_n = \Omega(a_n)$ simultaneously. Given a vector $ x\in\RR^d$ and a real value $\lambda>0$, we denote the soft thresholding operator $S_{\lambda}( x) = [\sgn(x_{j})\max\{|x_j|-\lambda, 0 \}]_{j=1}^d$. We use ''w.h.p.'' to denote ''with high probability''.


\section{Algorithm}\label{sec:alg}

We present the Prox-GD and Prox-Newton algorithms. For convenience, we denote 
\begin{align*}
\cF_{\lambda}(\theta) = \cL(\theta) + \lambda\norm{\theta}_1,
\end{align*}
where $\cL(\theta) = \frac{1}{\sqrt{n}} \norm{ y- X\theta}_2$. Since SQRT-Lasso is equipped with a sufficiently large regularization parameter $\lambda$ to prevent overfitting, i.e., $ y -  X \theta \neq 0$, we treat $\cL(\theta)$ as a differentiable function in this section. Formal justifications will be provided in the next section.

\subsection{Proximal Gradient Desccent Algorithm}

Given $\theta^{(t)}$ at $t$-th iteration, we consider a quadratic approximation of $\cF_{\lambda}(\theta)$ at $\theta=\theta^{(t)}$ as
\begin{align}
\cQ_{\lambda}(\theta, \theta^{(t)}) = \cL (\theta^{(t)}) + \nabla \cL (\theta^{(t)})^\top (\theta - \theta^{(t)}) + \frac{L^{(t)}}{2} \norm{\theta - \theta^{(t)}}_2^2 + \lambda\norm{\theta}_1,\label{eqn:prox}
\end{align}
where $L^{(t)}$ is a step size parameter determined by the backtracking line search. We then take
\begin{align*}
\theta^{(t+1)} = \argmin_{\theta} \cQ_{\lambda}(\theta, \theta^{(t)}) \textstyle = \cS_{\frac{\lambda}{L^{(t)}}}\left(\theta^{(t)} - \frac{\nabla \cL(\theta^{(t)})}{L^{(t)}}\right).
\end{align*}
For simplicity, we denote $\theta^{(t+1)}=\cT_{L^{(t+1)},\lambda}(\theta^{(t)})$. Given a pre-specified precision $\varepsilon$, we terminate the iterations when the approximate KKT condition holds:
\begin{align}
\omega_{\lambda}(\theta^{(t)}) = \min_{ g \in \partial \norm{\theta^{(t)}}_1} \norm{\nabla \cL (\theta^{(t)}) + \lambda  g}_{\infty} \leq \varepsilon. \label{eqn:approx_kkt}
\end{align}

\subsection{Proximal Newton Algorithm}

Given $\theta^{(t)}$ at $t$-th iteration, we denote a quadratic term of $\theta$ as
\begin{align*}
\norm{\theta-\theta^{(t)}}_{\nabla^2\cL(\theta^{(t)})}^2 = (\theta-\theta^{(t)})^{\top} \nabla^2\cL(\theta^{(t)}) (\theta-\theta^{(t)}),
\end{align*}
and consider a quadratic approximation of $\cF_{\lambda}(\theta)$ at $\theta=\theta^{(t)}$ is
\begin{align}
\cQ_{\lambda}(\theta, \theta^{(t)}) = \cL (\theta^{(t)}) + \nabla \cL (\theta^{(t)})^\top (\theta - \theta^{(t)}) + \frac{1}{2} \norm{\theta - \theta^{(t)}}_{\nabla^2\cL(\theta^{(t)})}^2 + \lambda\norm{\theta}_1.\label{eqn:prox-newton}
\end{align}We then take
\begin{align}
\theta^{(t+0.5)} = \argmin_{\theta} \cQ_{\lambda}(\theta, \theta^{(t)}). \label{eqn:quadratic}
\end{align}
An additional backtracking line search procedure is required to obtain
\begin{align*}
\theta^{(t+1)} =\theta^{(t)} + \eta_t(\theta^{(t+0.5)}-\theta^{(t)}),
\end{align*}
which guarantees $\cF_{\lambda}(\theta^{(t+1)})\leq\cF_{\lambda}(\theta^{(t)})$. The termination criterion for Prox-Newton is same with \eqref{eqn:approx_kkt}.
\begin{remark}
The $\ell_1$ regularized quadratic problem in \eqref{eqn:quadratic} can be solved efficiently by the coordinate descent algorithm combined with the active set strategy. See more details in \cite{zhao2014pathwise}. The computational cost is $\tilde{\cO}(snd)$, where $s\ll d$ is the solution sparsity.
\end{remark}

\begin{algorithm}[htb!]
	\caption{Prox-GD algorithm for solving the SQRT-Lasso optimization \eqref{sqrt-lasso-formulation}. We treat $\cL(\theta)$ as a differentiable function.} 
	\begin{algorithmic}
		\STATE  \textbf{Input:} $ y$, $X$, $\lambda$, $\varepsilon$, $L_{\max} > 0$
		\STATE  \textbf{Initialize:} ${\theta}^{(0)}$, $t \leftarrow 0$, $L^{(0)} \leftarrow L_{\max}$, $\tilde{L}^{(0)} \leftarrow L^{(0)}$
		\STATE  \textbf{Repeat:} $t \leftarrow t+1$
		\STATE  \hspace{.1in} \textbf{Repeat:} (Line Search)
		\STATE  \hspace{.2in} $\theta^{(t)} \leftarrow \cT_{\tilde{L}^{(t)},\lambda}(\theta^{(t-1)})$
		\STATE  \hspace{.2in} \textbf{If} {$\cF_{\lambda}(\theta^{(t)}) < \cQ_{\lambda}(\theta^{(t)},\theta^{(t-1)})$}
		\STATE  \hspace{.2in} \textbf{Then} $\tilde{L}^{(t)} \leftarrow \frac{\tilde{L}^{(t)}}{2}$
		\STATE  \hspace{.1in} \textbf{Until:} {$\cF_{\lambda}(\theta^{(t)}) \geq \cQ_{\lambda}(\theta^{(t)},\theta^{(t-1)})$}
		\STATE  \hspace{.1in} $L^{(t)} \leftarrow \min\{ 2\tilde{L}^{(t)}, L_{\max}\}$, $\tilde{L}^{(t)} \leftarrow L^{(t)}$
		\STATE  \hspace{.1in} $\theta^{(t)} \leftarrow \cT_{L^{(t)},\lambda}(\theta^{(t-1)})$
		\STATE  \textbf{Until:} {$\omega_{\lambda}(\theta^{(t)}) \leq \varepsilon$}
		\STATE  \textbf{Return:} $\hat{\theta} \leftarrow \theta^{(t)}$
	\end{algorithmic} \label{alg:prox-GD}
\end{algorithm}

\begin{algorithm}[htb!]
	\caption{Prox-Newton algorithm for solving the SQRT-Lasso optimization \eqref{sqrt-lasso-formulation}. We treat $\cL(\theta)$ as a differentiable function.} 
	\begin{algorithmic}
		\STATE  \textbf{Input:} $ y$, $ X$, $\lambda$, $\varepsilon$
		\STATE  \textbf{Initialize:} ${\theta}^{(0)}$, $t\gets 0$, $\mu \gets 0.9$, $\alpha \gets \frac{1}{4}$
		\STATE  \textbf{Repeat:} $t\gets t+1$
		\STATE   \hspace{.1in} $\theta^{(t)}\gets\argmin_{\theta}\cQ_{\lambda}(\theta, \theta^{(t-1)})$
		\STATE  \hspace{.1in}  $\Delta \theta^{(t)} \gets \theta^{(t)} - \theta^{(t-1)}$
		\STATE  \hspace{.1in}  $\gamma_t \gets \nabla\cL\rbr{\theta^{(t-1)}}^{\top} \Delta \theta^{(t)} + \lambda \left( \norm{\theta^{(t)}}_1 - \norm{{\theta^{(t-1)}}}_1 \right)$
		\STATE  \hspace{.1in}  $\eta_t \gets 1$, $q \gets 0$
		\STATE  \hspace{.1in}  \textbf{Repeat:} $q \gets q+1$ (Line Search)
		\STATE  \hspace{.2in}  $\eta_t \gets \mu^q$
		\STATE  \hspace{.1in}  \textbf{Until} $\cF_{\lambda} \rbr{\theta^{(t-1)} + \eta_t\Delta \theta^{(t)}}  \le \cF_{\lambda} \rbr{\theta^{(t-1)}} + \alpha \eta_t \gamma_t$
		\STATE  \hspace{.1in}  $\theta^{(t)} \gets \theta^{(t)}  + \eta_t\Delta \theta^{(t-1)}$
		\STATE  \textbf{Until:} {$\omega_{\lambda}(\theta^{(t)}) \leq \varepsilon$}
		\STATE  \textbf{Return:} $\hat{\theta} \leftarrow \theta^{(t)}$
	\end{algorithmic} \label{alg:prox-Newton}
\end{algorithm}

Details of Prox-GD and Prox-Newton algorithms are summarized in Algorithms~\ref{alg:prox-GD} and \ref{alg:prox-Newton} respectively. To facilitate global fast convergence, we further combine the pathwise optimization \cite{friedman2007pathwise} with the proximal algorithms. See more details in Section \ref{sec:pathwise}.

\begin{remark} We can also apply proximal quasi-Newton method. Accordingly, at each iteration, the Hessian matrix in \eqref{eqn:prox-newton} is replaced with an approximation. See \cite{bertsekas1999nonlinear} for more details.
\end{remark}

	

\section{Theoretical Analysis}

We start with defining the locally restricted strong convexity/smoothness and Hessian smoothness.
\begin{definition}\label{def:rscrss}
	Denote $$\cB_{r} = \{ \theta \in \RR^d : \norm{\theta - \theta^*}_2^2 \leq r \}$$ for some constant $r \in \RR^+$. For any $ v,  w \in \cB_{r}$ satisfying $\norm{ v -  w}_0 \leq s$, $\cL$ is \textit{locally restricted strongly convex} (LRSHC), \textit{smooth} (LRSS), and \textit{Hessian smooth} (LRHS) respectively on $\cB_{r}$ at sparsity level $s$, if there exist universal constants $\rho^{-}_{s}, \rho^{+}_{s}, L_{s} \in (0,\infty)$ such that
	\begin{align}
	\text{LRSC:}& \cL( v) - \cL( w)  -  \nabla \cL( w)^\top  ( v -  w)  \geq  \frac{{\rho}^{-}_{s}}{2} \norm{ v -  w}_2^2, \nonumber \\
	\text{LRSS:}& \cL( v)  -  \cL( w)  -  \nabla \cL( w)^\top  ( v -  w)  \leq  \frac{{\rho}^{+}_{s}}{2} \norm{ v -  w}_2^2, \nonumber \\
	\text{LRHS:}& u^\top(\nabla^2\cL( v)-\nabla^2\cL( w)) u \leq L_{s}\norm{ v- w}_2^2, \label{eqn:rscrss}
	\end{align}
	for any $ u$ satisfying $\norm{ u}_0\leq s$ and $\norm{ u}_2=1$.
	We define the locally restricted condition number at sparsity level $s$ as $\kappa_{s} = \frac{\rho^{+}_{s}}{\rho^{-}_{s}}$. 
\end{definition}
LRSC and LRSS are locally constrained variants of restricted strong convexity and smoothness  \citep{agarwal2010fast,xiao2013proximal}, which are keys to establishing the strong convergence guarantees in high dimensions. The LRHS is parallel to the local Hessian smoothness for analyzing the proximal Newton algorithm in low dimensions \citep{lee2014proximal}. This is also closely related to the self-concordance \citep{nemirovski2004interior} in the analysis of Newton method \citep{boyd09}. Note that $r$ is associated with the radius of the neighborhood of $\theta^*$ excluding the nonsmooth (and overfitted) region of the problem to guarantee strong convergence, which will be quantified below.

Next, we prove that the $\ell_2$ loss of SQRT-Lasso enjoys the good geometric properties defined in Definition \ref{def:rscrss} under mild modeling assumptions. 
\begin{lemma}\label{lem:rsc}
	Suppose  $\bepsilon$ has i.i.d. sub-Gaussian entries with $\EE[\epsilon_i]=0$ and $\EE[\epsilon_i^2] = \sigma^2$, $\norm{\theta^*}_0=s^*$. Then for any $\lambda \geq {C_{1}\sqrt{\frac{\log d}{n}}}$, w.h.p. we have 
	\begin{align*}
	\lambda \geq \frac{C_1}{4} \norm{\nabla \cL(\theta^*)}_{\infty}.
	\end{align*}
	Moreover, given each row of the design matrix $ X$ independently sampled from a sub-Gaussian distribution with the positive definite covariance matrix $\Sigma_{ X} \in \RR^{d \times d}$ with bounded eigenvalues. Then for $$n \geq C_2 s^* \log d,$$ $\cL(\theta)$ satisfies LRSC, LRSS, and LRHS properties on $\cB_{r}$ at sparse level $s^*+2\tilde{s}$ with high probability. Specifically, \eqref{eqn:rscrss} holds with $$\rho^{+}_{s^*+2\tilde{s}} \leq \frac{C_3}{\sigma},~\rho^{-}_{s^*+2\tilde{s}} \geq \frac{C_4}{\sigma}~\textrm{and}~L_{s^*+2\tilde{s}} \leq \frac{C_5}{\sigma},$$ where  $C_1,\ldots,C_{5} \in \RR^+$ are generic constants, and $r$ and $\tilde{s}$ are sufficiently large constants, i.e., $\tilde{s} > (196{\kappa}_{s^*+2\tilde{s}}^2+144{\kappa}_{s^*+2\tilde{s}})s^*$. 
\end{lemma}


The proof is provided in Appendix~\ref{pf:lem:rsc}.
Lemma~\ref{lem:rsc} guarantees that with high probability:

\noindent {\bf (i)} $\lambda$ is sufficiently large to eliminate the irrelevant variables and yields sufficiently sparse solutions \citep{bickel2009simultaneous,negahban2012unified};

\noindent {\bf (ii)} LRSC, LRSS, and LRHS hold for the $\ell_2$ loss of SQRT-Lasso such that fast convergence of the proximal algorithms can be established in a sufficiently large neighborhood of $\theta^*$ associated with $r$;

\noindent {\bf (iii)} \eqref{eqn:rscrss} holds in $\cB_{r}$ at sparsity level $s^*+2\tilde{s}$. Such a property is another key to the fast convergence of the proximal algorithms, because the algorithms can not ensure that the nonzero entries exactly falling in the true support set of $\theta^*$.

\subsection{Local Linear Convergence of Prox-GD}

For notational simplicity, we denote
\begin{align*}
\cS^*=\{j~|~\theta_j^*\neq 0\},\quad\overline{\cS}^*=\{j~|~\theta_j^*= 0\},~~\textrm{and} \cB_{r}^{s^*+\tilde{s}} = \cB_{r} \cap \{ \theta \in \RR^d : \norm{\theta - \theta^*}_0 \leq s^*+\tilde{s} \}.
\end{align*}
To ease the analysis, we provide a local convergence analysis when $\theta \in \cB_{r}^{s^*+\tilde{s}}$ is sufficiently close to $\theta^*$. The convergence of Prox-GD is presented as follows. 
\begin{theorem}\label{thm:proxgd}
	Suppose $ X$ and $n$ satisfy conditions in Lemma~\ref{lem:rsc}. Given $\lambda$ and $\theta^{(0)}$ such that $\lambda \geq \frac{C_1}{4} \norm{\nabla \cL(\theta^*)}_{\infty}$, $\norm{\theta^{(0)} - \theta^*}^2_2 \leq s^*\left({8 \lambda}/{\rho^{-}_{s^*+\tilde{s}}}\right)^2$ and $\theta^{(0)}\in\cB_{r}^{s^*+\tilde{s}}$, we have sufficiently sparse solutions throughout all iterations, i.e., $$\norm{[\theta^{(t)}]_{\overline{\cS}^*}}_{0}\leq \tilde{s}.$$ Moreover, given $\varepsilon>0$, we need at most 
	\begin{align*}
	T=\cO \left( {\kappa}_{s^*+2\tilde{s}}\log \left({ \frac{\kappa_{s^*+2\tilde{s}}^3 {s^*} \lambda^2}{\varepsilon^2}} \right) \right)
	\end{align*}
	iterations to guarantee that the output solution $\hat{\theta}$ satisfies
	\begin{align*}
	\norm{\hat{\theta}-\overline{\theta}}_2^2 \textstyle = \cO \left(\left(1 - {\frac{1}{8{\kappa}_{s^*+2\tilde{s}}}}\right)^{T} \varepsilon \lambda s^* \right)~~~\textrm{and}~~~
	\cF_{\lambda} (\hat{\theta}) - \cF_{\lambda} (\overline{\theta}) \textstyle = \cO \left( {\left(1 - {\frac{1}{8{\kappa}_{s^*+2\tilde{s}}}}\right)^{T} \varepsilon \lambda {s^*}} \right),
	\end{align*}
	where $\overline{\theta}$ is the unique sparse global optimum to \eqref{sqrt-lasso-formulation} with $\norm{[\overline{\theta}]_{\overline{\cS}^*}}_{0}\leq \tilde{s}.$
\end{theorem}
The proof is provided in Appendix~\ref{pf:thm:proxgd}. Theorem~\ref{thm:proxgd} guarantees that when properly initialized, the Prox-GD algorithm iterates within the smooth region, maintains the solution sparsity, and achieves a local linear convergence to the unique sparse global optimum to \eqref{sqrt-lasso-formulation}. 

\subsection{Local Quadratic Convergence of Prox-Newton}

We then present the convergence analysis of the Prox-Newton algorithm as follows. 
\begin{theorem}\label{thm:proxnewton}
	Suppose $ X$ and $n$ satisfy conditions in Lemma~\ref{lem:rsc}. Given $\lambda$ and $\theta^{(0)}$ such that $\lambda \geq \frac{C_1}{4} \norm{\nabla \cL(\theta^*)}_{\infty}$, $\norm{\theta^{(0)} - \theta^*}^2_2 \leq s^*\left({8 \lambda}/{\rho^{-}_{s^*+\tilde{s}}}\right)^2$ and $\theta^{(0)}\in\cB_{r}^{s^*+\tilde{s}}$, we have sufficiently sparse solutions throughout all iterations, i.e., $$\norm{[\theta^{(t)}]_{\overline{\cS}^*}}_{0}\leq \tilde{s}.$$ Moreover, given $\varepsilon>0$, we need at most
	\begin{align*}
	T=\cO \left( \log {\log\rbr{{\frac{3\rho^+_{s^*+2\tilde{s}}}{\varepsilon}}}} \right)
	\end{align*}
	iterations to guarantee that the output solution $\hat{\theta}$ satisfies
	\begin{align*}
	\norm{\hat{\theta}-\overline{\theta}}_2^2 \textstyle = \cO \left(\left({ \frac{L_{s^*+2\tilde{s}}}{2\rho^-_{s^*+2\tilde{s}}}}\right)^{2^T} \varepsilon \lambda s^* \right)~~~\textrm{and}~~~
	\cF_{\lambda} (\hat{\theta}) - \cF_{\lambda} (\overline{\theta}) \textstyle = \cO \left( {\left({ \frac{L_{s^*+2\tilde{s}}}{2\rho^-_{s^*+2\tilde{s}}}}\right)^{2^T} \varepsilon \lambda {s^*}} \right),
	\end{align*}
	where $\overline{\theta}$ is the unique sparse global optimum to \eqref{sqrt-lasso-formulation}.
\end{theorem}
The proof is provided in Appendix~\ref{pf:thm:proxnewton}. Theorem~\ref{thm:proxnewton} guarantees that when properly initialized, the Prox-Newton algorithm also iterates within the smooth region, maintains the solution sparsity, and achieves a local quadratic convergence to the unique sparse global optimum to \eqref{sqrt-lasso-formulation}.

\begin{remark}
	Our analysis can be further extended to the proximal quasi-Newton algorithm. The only technical difference is controlling the error of the Hessian approximation under restricted spectral norm.
\end{remark}

\subsection{Statistical Properties}

Next, we characterize the statistical properties for the output solutions of the proximal algorithms. 
\begin{theorem}\label{thm:statrate}
	Suppose $ X$, and $n$ satisfy conditions in Lemma~\ref{lem:rsc}. Given $\lambda=C_1\sqrt{\log d/n}$, if the output solution $\hat{\theta}$ obtained from Algorithm~\ref{alg:prox-GD} and \ref{alg:prox-Newton} satisfies the approximate KKT condition, $$\omega_{\lambda}(\hat{\theta}) \leq \varepsilon = \cO\left(\frac{\sigma s^* \log d}{n}\right),$$ then we have:
	\begin{align*}
	\norm{ \hat{\theta} - \theta^*}_2 \textstyle = \cO_P \left( \sigma \sqrt{\frac{s^* \log d}{n}} \right)~~~\text{and}~~~
	\norm{ \hat{\theta} - \theta^*}_1 \textstyle = \cO_P \left( \sigma s^* \sqrt{\frac{\log d}{n}} \right). 
	\end{align*}
	Moreover, we have
	\begin{align*}
	\left| \hat{\sigma} - \sigma \right| = \cO_P \left(\frac{\sigma s^* \log d}{n} \right),~\textrm{where}~\hat{\sigma} = \frac{\norm{ y -  X \hat{\theta}}_2}{\sqrt{n}}.
	\end{align*}
\end{theorem}
The proof is provided in Appendix~\ref{pf:thm:statrate}. Recall that we use $\cO_P(\cdot)$ to denote the limiting behavior in probability. Theorem~\ref{thm:statrate} guarantees that the output solution $\hat{\theta}$ obtained from Algorithm~\ref{alg:prox-GD} and \ref{alg:prox-Newton}  achieves the minimax optimal rate of convergence in parameter estimation \citep{raskutti2011minimax,ye2010rate}. Note that in the stopping criteria $\omega_{\lambda}(\hat{\theta}) \leq \varepsilon$, $\varepsilon$ is not a tuning parameter, where $\cO\left(\frac{\sigma s^* \log d}{n}\right)$ only serves as an upper bound and we can choose a small $\varepsilon$ as desired. This is fundamentally different with the optimal $\lambda_{\sf Lasso}$ that tightly depends on $\sigma$.

\section{Boosting Performance via Pathwise Optimization Scheme}\label{sec:pathwise}

We then apply the pathwise optimization scheme to the proximal algorithms, which extends the local fast convergence established in Section 3 to the global setting\footnote{We only provide partial theoretical guarantees.}. The pathwise optimization is essentially a multistage optimization scheme for boosting the computational performance \cite{friedman2007pathwise,xiao2013proximal,zhao2014pathwise}. 

Specifically, we solve \eqref{sqrt-lasso-formulation} using a geometrically decreasing sequence of regularization parameters $$\lambda_{[0]} > \lambda_{[1]}> \ldots > \lambda_{[N]},$$ where $\lambda_{[N]}$ is the target regularization parameter of SQRT-Lasso. This yields a sequence of output solutions $$\hat{\theta}_{[0]},~\hat{\theta}_{[1]},\ldots,~\hat{\theta}_{[N]},$$ also known as the solution path. At the $K$-th optimization stage, we choose $\hat{\theta}_{[K-1]}$ (the output solution of the $(K-1)$-th stage) as the initial solution, and solve \eqref{sqrt-lasso-formulation} with $\lambda=\lambda_{[K]}$ using the proximal algorithms. This is also referred as the warm start initialization in existing literature \citep{friedman2007pathwise}. Details of the pathwise optimization is summarized in Algorithm~\ref{alg:homotopy}. In terms of $\epsilon_{[K]}$, because we only need high precision for the final stage, we set $\epsilon_{[K]}=\lambda_{[K]}/4 \gg \epsilon_{[N]}$ for $K<N$.

\begin{algorithm}[htb!]
	\caption{The pathwise optimization scheme for the proximal algorithms. We solve the optimization problem using a geometrically decreasing sequence of regularization parameters.} 
	\begin{algorithmic}
		\STATE  \textbf{Input:} $ y$, $ X$, $N$, $\lambda_{[N]}$, $\varepsilon_{[N]}$
		\STATE \textbf{Initialize:}  $\hat{\theta}_{[0]} \leftarrow  0$,  $\lambda_{[0]} \leftarrow \norm{\nabla{\cL}( 0)}_{\infty}$, $\eta_{\lambda}  \leftarrow \rbr{\frac{\lambda_{[N]}}{\lambda_{[0]}}}^{\frac{1}{N}}$
		\STATE \hspace{.1in} \textbf{For:} $K = 1, \ldots, N$
		\STATE \hspace{.1in} $\lambda_{[K]} \leftarrow \eta_{\lambda} \lambda_{[K-1]}$, $\theta_{[K]}^{(0)} \leftarrow \hat{\theta}_{[K-1]}$, $\varepsilon_{[K]} \leftarrow \varepsilon_{[N]}$
		\STATE  $\hat{\theta}_{[K]} \leftarrow \text{Prox-Alg}\left(  y,  X, \lambda_{[K]}, \theta_{[K]}^{(0)}, \varepsilon_{[K]} \right)$
		\STATE  \textbf{End For} \\
		\STATE  \textbf{Return:} $\hat{\theta}_{[N]}$
	\end{algorithmic} \label{alg:homotopy}
\end{algorithm}

As can be seen in Algorithm 3, the pathwise optimization scheme starts with $$\lambda_{[0]} = \norm{\nabla \cL( 0)}_{\infty} = \left\|\frac{ X^\top  y}{\sqrt{n}\norm{ y}_2}\right\|_{\infty},$$ which yields an all zero solution $\hat{\theta}_{[0]} =  0$ (null fit). We then gradually decrease the regularization parameter, and accordingly, the number of nonzero coordinates gradually increases.


The next theorem proves that there exists an $N_1<N$ such that the fast convergence of the proximal algorithms holds for all $\lambda_{[K]}$'s, where $K \in [N_1+1,..,N]$.
\begin{theorem}\label{thm:rsc2}
	 Suppose the design matrix $ X$ is sub-Gaussian, and $\lambda_{[N]}=C_1\sqrt{\log d/n}$. For $n \geq C_2 s^* \log d$ and $\eta_{\lambda} \in (\frac{5}{6},1)$, the following results hold:

\noindent {\bf (I)} There exists an $N_1< N$ such that $$r> s^*\left({8 \lambda_{N_1}}/{\rho^{-}_{s^*+\tilde{s}}}\right)^2;$$

\noindent {\bf (II)} For any $K \in [N_1+1,..,N],$ we have $\norm{\theta^{(0)}_{[K]} - \theta^*}^2_2 \leq s^*\left({8 \lambda_{[K]}}/{\rho^{-}_{s^*+\tilde{s}}}\right)^2$, $\theta^{(0)}_{[K]}\in \cB_{r}^{s^*+\tilde{s}}$ w.h.p.;

\noindent {\bf (III)} Theorems~\ref{thm:proxgd} and \ref{thm:proxnewton} hold for all $\lambda_K$'s, where $K \in [N_1+1,..,N]$ w.h.p..
\end{theorem}

\begin{figure}[t]
	\centering
	\includegraphics[width=0.9\linewidth]{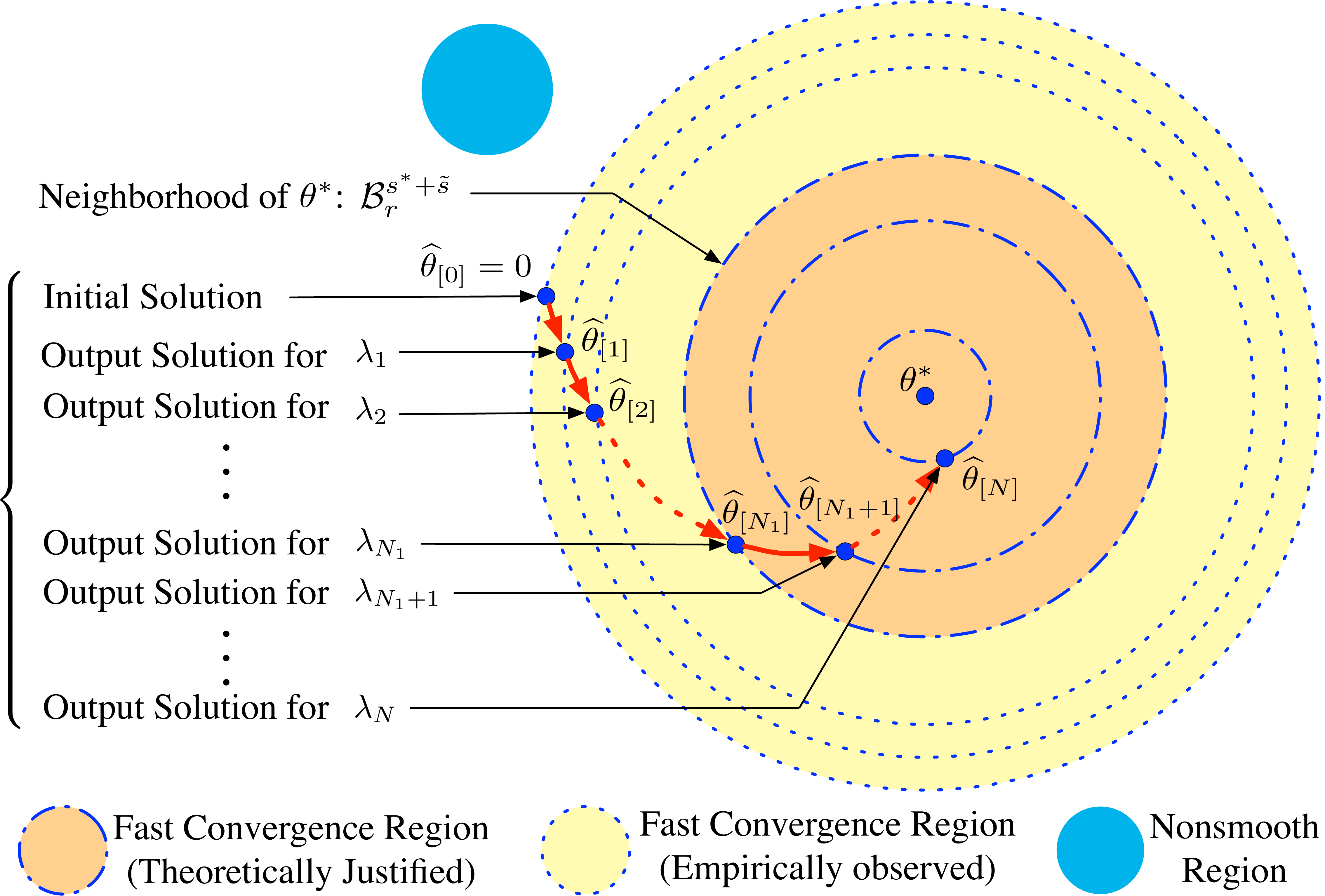}
	\caption{A geometric illustration for the fast convergence of the proximal algorithms. The proximal algorithms combined with the pathwise optimization scheme suppress the overfitting and yield sparse solutions along the solution path. Therefore, the nonsmooth region of the $\ell_2$ loss, i.e., the set $\cbr{\theta: y -  X \theta =  0}$, is avoided, and LRSC, LRSS, and LRHS enable the proximal algorithms to achieve fast convergence.}\label{fig:ball}
\end{figure}

The proof is provided in Appendix~\ref{pf:thm:rsc2}. Theorem~\ref{thm:rsc2} implies that for all $\lambda_{[K]}$'s, where $K \in [N_1,N_1+1,..,N]$, the regularization parameter is large enough for ensuring the solution sparsity and preventing overfitting. Therefore, the fast convergence of proximal algorithms can be guaranteed. For $\lambda_{[0]}$ to $\lambda_{[N_1]}$, we do not have theoretical justification for the fast convergence due to the limit of our proof technique. However, as $\lambda_{[0]}$,..., $\lambda_{[N_1]}$ are all larger than $\lambda_{[N_1+1]}$, we can expect that the obtained model is very unlikely to be overfitted. Accordingly, we can also expect that all intermediate solutions $\hat{\theta}_{[K]}$'s stay out of the nonsmooth region, and LRSC, LRSS, and LRHS properties should also hold along the solution path. Therefore, the proximal algorithms achieve fast convergence in practice. Note that when the design $ X$ is normalized, we have $\lambda_{[0]} = \cO(d)$, which implies that the total number $N$ of regularization parameter satisfies $$N = \cO(\log d).$$ A geometric illustration of the pathwise optimization is provided in Figure~\ref{fig:ball}. The supporting numerical experiments are provided in Section 6.

\section{Extension to CMR and SPME}\label{sec:extension}

We extend our algorithm and theory to calibrated multivariate regression (CMR, \cite{liu2015calibrated}) and sparse precision matrix estimation (SPME, \cite{liu2017tiger}). Due to space limit, we only provide a brief discussion and omit the detailed theoretical deviation.

\noindent\textbf{Extension to CMR.} Recall that CMR solves
\begin{align*}
\overline{\Theta}^{\sf CMR} = \argmin_{\theta\in\RR^{d\times m}} \frac{1}{\sqrt{n}} \sum_{k=1}^m\norm{ Y_{*k}- X\Theta_{*k}}_2 + \lambda_{\sf CMR}\norm{\Theta}_{1,2}.
\end{align*}
Similar to SQRT-Lasso, we choose a sufficiently large $\lambda_{\sf CMR}$ to prevent overfitting. Thus, we can expect $$\norm{ Y_{*k}- X\Theta_{*k}}_2 \neq 0~\textrm{for all}~k=1,...,m,$$ and treat the nonsmooth loss of CMR as a differentiable function. 
Accordingly, we can trim our algorithms and theory for the nonsmooth loss of CMR, and establish fast convergence guarantees, as we discussed in \S\ref{sec:pathwise}.
	


\noindent\textbf{Extension to SPME.} \cite{liu2017tiger} show that a $d\times d$ sparse precision matrix estimation problem is equivalent to a collection of $d$ sparse linear model estimation problems. For each linear model, we apply SQRT-Lasso to estimate the regression coefficient vector and the standard deviation of the random noise. Since SQRT-Lasso is adaptive to imhomogenous noise, we can use one singular regularization parameter to prevent overfitting for all SQRT-Lasso problems. Accordingly, we treat the nonsmooth loss function in every SQRT-Lasso problem as a differentiable function, and further establish fast convergence guarantees for the proximal algorithms combined with the pathwise optimization scheme.

	
	

\section{Numerical Experiments}

We compare the computational performance of the proximal algorithms with other competing algorithms using both synthetic and real data. All algorithms are implemented in \texttt{C++} with double precision using a PC with an Intel 2.4GHz Core i5 CPU and 8GB memory. All algorithms are combined with the pathwise optimization scheme to boost the computational performance. 
Due to space limit, we omit some less important details.

\begin{figure*}[htb!]
	\centering
	\includegraphics[width=0.45\linewidth]{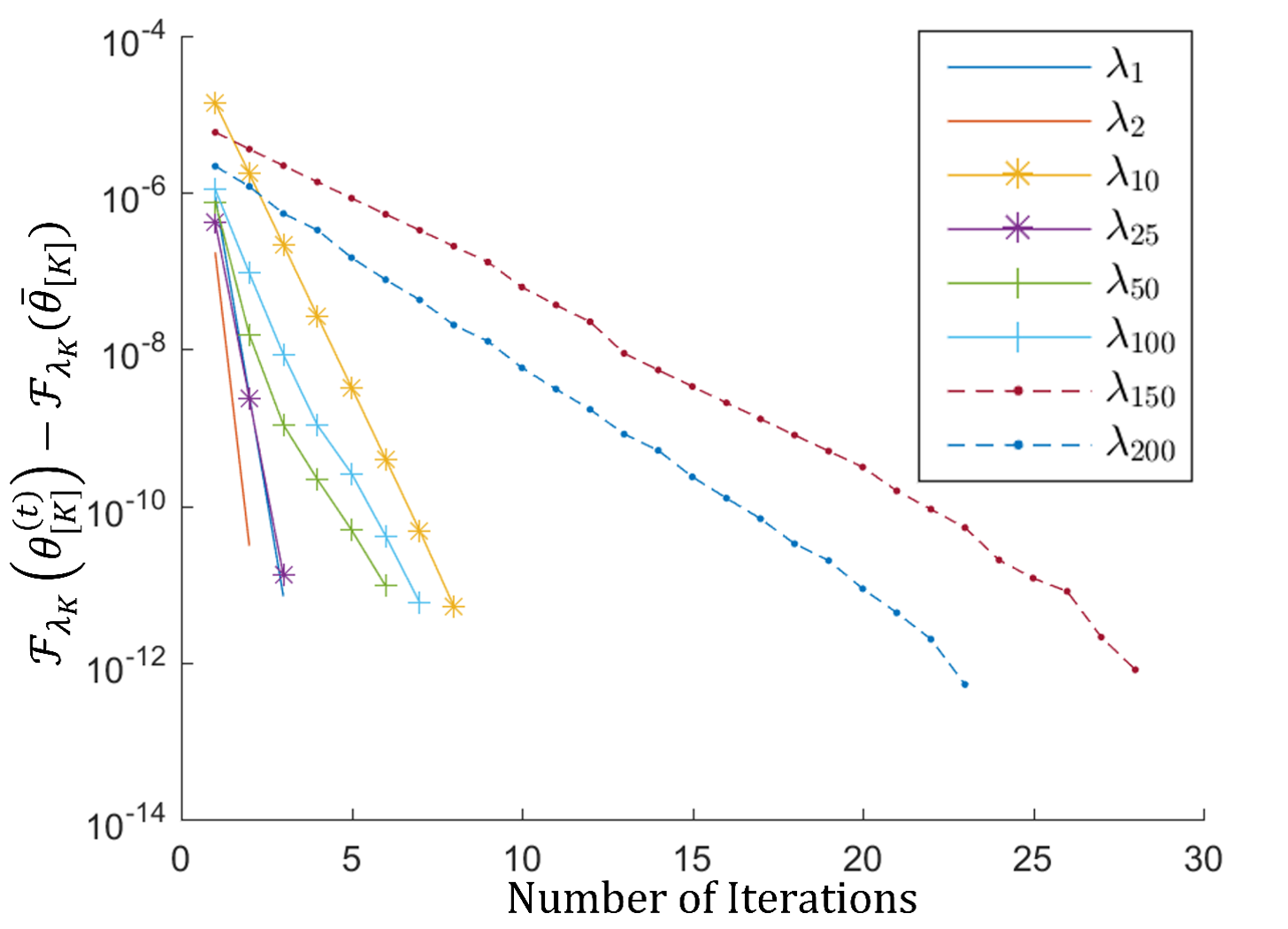}
	\includegraphics[width=0.45\linewidth]{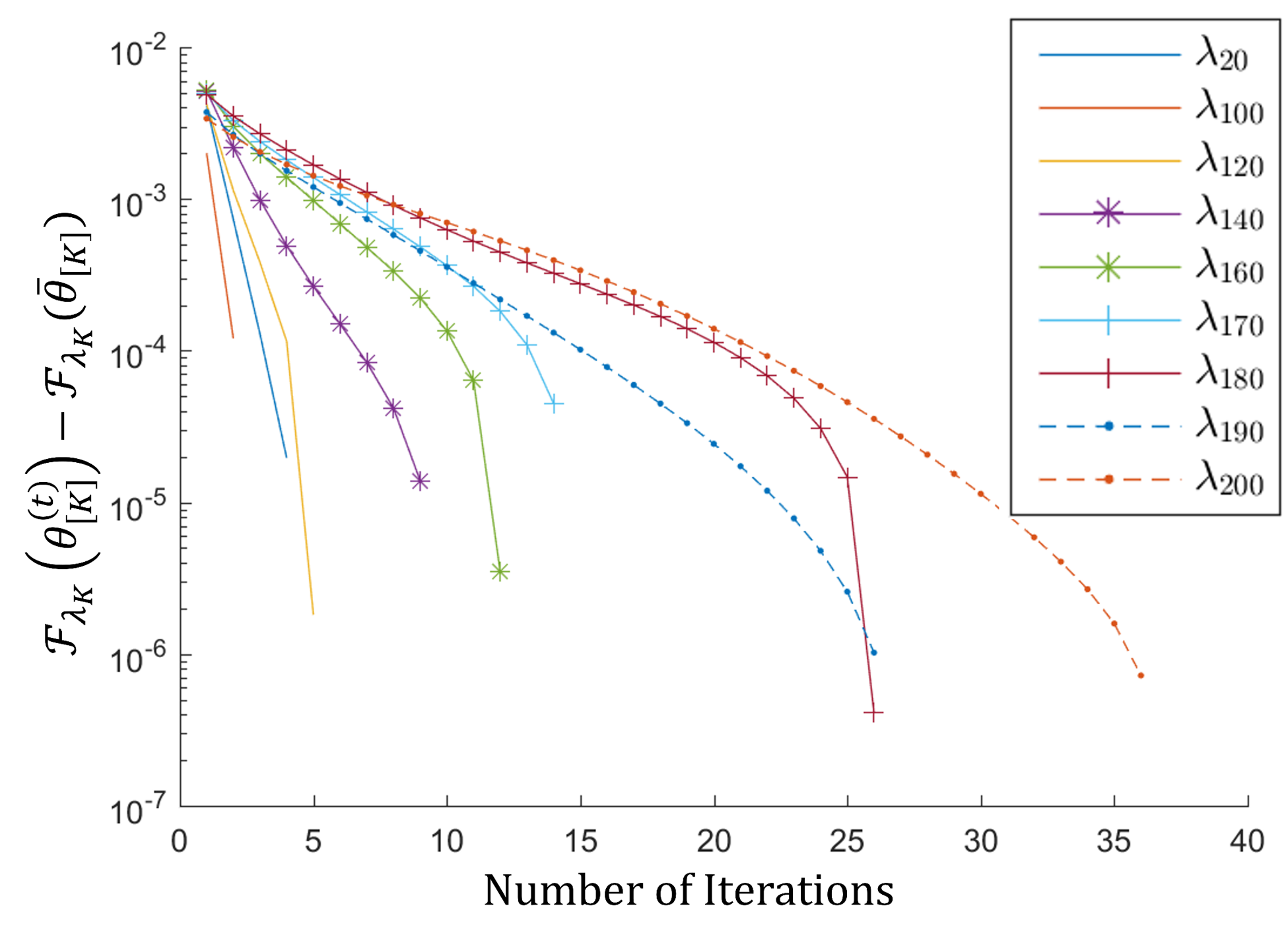}
	\caption{The objective gap v.s. the number of iterations. We can see that the Prox-GD (Left) and Prox-Newton (Right) algorithms achieve linear and quadratic convergence at every stage respectively.}\label{fig:fast_cvg}
\end{figure*}


\begin{table*}[!htb]

\caption{Computational performance of Prox-GD on synthetic data under different choices of variance $\sigma$, the number of stages $N$, and the stopping criterion $\varepsilon_{N}$. The training time is presented, where each entry is the mean execution time in seconds over 100 random trials. The minimal mean square error (MSE) is $\frac{1}{n} \norm{ y -  X \hat{\theta}_{[K]}}_2^2$, where $\hat{\theta}_{[K]}$ is the optimal solution that attains $\min \cF_{\lambda_K}(\theta)$ for all stages $K=1,\ldots,N$.}
\begin{center}
\renewcommand{\arraystretch}{1.0}
{
	\begin{tabular}{c|c|c|c|c|c|c|c|c|c|c}
		\Xhline{1 pt}
		\multirow{2}{*}{$\sigma$} & \multirow{2}{*}{$N$} &\multicolumn{3}{c|}{$\varepsilon_{N}$}& Minimal &
		\multirow{2}{*}{$\sigma$} &\multicolumn{3}{c|}{$\varepsilon_{N}$}& Minimal  \\
		\cline{3-5}
		\cline{8-10}
		& &$10^{-4}$ &$10^{-5}$ &$10^{-6}$&MSE &
		&$10^{-4}$ &$10^{-5}$ &$10^{-6}$&MSE\\
		\Xhline{1 pt}
		
		\multirow{3}{*}{$0.1$} 
		& 1 &0.3718 &0.3721 &0.3647  &\multirow{3}{*}{0.0132}&
		\multirow{3}{*}{$0.5$} 
		&0.2850 &0.2951 &0.2886 &\multirow{3}{*}{0.3054} \\
		\cline{2-5}
		\cline{8-10}
		
		& 10   &\textbf{0.2749 } &\textbf{0.2764 } &\textbf{0.2804 } & & 
		&\textbf{0.1646} &\textbf{0.1698} &\textbf{0.1753} & 
		\\
		\cline{2-5}
		\cline{8-10}
		
		& 30   &0.3364  &0.3452  &0.3506  & & 
		&0.2207 &0.2247 &0.2285 & \\
		\cline{2-5}
		\cline{8-10}
		
		\hline
		\multirow{3}{*}{$1$} 
		& 1 &0.2347 &0.2478 &0.2618 &\multirow{3}{*}{1.1833} &
		\multirow{3}{*}{$2$} 
		& 0.4317 &0.4697 &0.4791 &\multirow{3}{*}{4.2197} \\
		\cline{2-5}
		\cline{8-10}
		& 10   &\textbf{0.1042} &\textbf{0.1031} &\textbf{0.1091} & &
		&\textbf{0.1661} &\textbf{0.1909} &\textbf{0.2110} & \\
		\cline{2-5}
		\cline{8-10}
		& 30   &0.2172 &0.2221 &0.2199 & &
		&0.2701  &0.2955  &0.3134 &\\
		\cline{2-5}
		\cline{8-10}
		
		\Xhline{1 pt}
	\end{tabular}
}\label{table:pista_ista_cmp}
\end{center}

\caption{Timing comparison between multiple algorithms on real data. Each entry is the execution time in seconds. All experiments are conducted to achieve similar suboptimality.}
\begin{center}
\renewcommand{\arraystretch}{1.0}
{
	\begin{tabular}{c|c|c|c|c|c|c|c}
		\Xhline{1 pt}
		\multirow{2}{*}{Data Set} & \multicolumn{6}{c|}{SQRT-Lasso} & Lasso \\
		\cline{2-8}
		& Prox-GD & Newton &ADMM  &ScalReg &CD &Alt.Min & PISTA  \\
		\Xhline{1 pt}
		{\rm Greenhouse} &~5.812    &~{\bf 1.708}     &~1027.590    &~3180.747    &~14.311  &~99.814   &~5.113 \\
		{\rm DrivFace}   &~{\bf 0.421}    &~0.426     &~18.879      &~124.032    &~3.138    &~17.691  &~0.414 \\
		\Xhline{1 pt}
	\end{tabular}
}\label{table:realdata_exp}
\end{center}
\end{table*}

\begin{table*}[!htb]

\caption{Timing comparison between multiple algorithms for sparse precision matrix estimation on biology data under different levels of sparsity recovery. Each entry is the execution time in seconds. All experiments are conducted to achieve similar suboptimality. Here CD failed to converge and the program aborted before reaching the desired suboptimality. Scalreg failed to terminate in 1 hour for Estrogen. }
\begin{center}
	\renewcommand{\arraystretch}{1.0}
		{
		\begin{tabular}{c|c|c|c|c|c|c}
			\Xhline{1 pt}
			\multirow{2}{*}{\rm Sparsity} & \multicolumn{6}{c}{Arabidopsis} \\
			\cline{2-7}
			& Prox-GD & Newton & ADMM & ScalReg &CD & Alt.Min \\
			\Xhline{1 pt}
			1\% &5.099  &{\bf 1.264} &292.05 &411.74 &12.02 &183.63 \\
			3\% &6.201 &{\bf 2.088} &339.22 &426.08 &18.18 &217.72 
			\\
			5\% &7.122 &{\bf 2.258} &366.67 &435.50 &28.60 &256.97 \\
			\Xhline{1 pt}
			\multirow{1}{*}{\rm Sparsity} & \multicolumn{6}{c}{Estrogen}\\
			\Xhline{1 pt}
			1\% &108.24  &{\bf 3.099}   &1597.41  &$>$3600        &136.181  &634.128     \\
			3\% &130.93  &{\bf 7.101}   &1845.60  &$>$3600        &332.028  &662.232   \\
			5\% &143.54  &{\bf 10.120}  &2029.61  &$>$3600        &588.407  &739.464  
			     \\
			\Xhline{1 pt}
			\multirow{1}{*}{\rm Sparsity} & \multicolumn{6}{c}{Lymph}  \\
			\Xhline{1 pt}
			1\% 
			&3.709  &{\bf 0.625} &256.43 &354.93 &7.208 &120.25   \\
			3\% 
			&4.819  &{\bf 0.905} &289.08 &355.30 &10.51 &130.61   \\
			5\% 
			&4.891  &{\bf 1.123} &310.16 &358.70 &14.95 &148.92 \\
			\Xhline{1 pt}
			\multirow{1}{*}{\rm Sparsity} & \multicolumn{6}{c}{Leukemia}  \\
			\Xhline{1 pt}
			1\% 
			&8.542   &{\bf 2.715}   &331.28   &610.147  &173.319  &239.247 \\
			3\% 
			&10.562  &{\bf 3.935}   &384.74   &766.072  &174.295  &285.127   \\
			5\% 
			&10.768  &{\bf 4.712}   &442.54   &1274.38  &288.884  &333.611   \\
			\Xhline{1 pt}
	\end{tabular}}\label{table:pissta_admm_spme}
\end{center}


\caption{Timing comparison between multiple algorithms for calibrated multivariate regression on synthetic and real data with different values of $\lambda_N$. Each entry is the execution time in seconds. All experiments are conducted to achieve similar suboptimality. Here CD failed to converge and the program aborted before reaching the desired suboptimality.}
\begin{center}
	\renewcommand{\arraystretch}{1.18}
	{
		\begin{tabular}{c|c|c|c|c|c|c|c|c}
			\Xhline{1 pt}
			\multirow{2}{*}{$\lambda_N $} &
			\multicolumn{4}{c|}{Synthetic ($\sigma = 1$)}& \multicolumn{4}{c}{DrivFace}  \\
			\cline{2-9}
			& Prox-GD & Newton & ADMM & CD & Prox-GD & Newton & ADMM & CD \\
			\Xhline{1 pt}
			$\sqrt{\log{d}/n}  $ & 0.2964  &{\bf 0.0320} &14.8307 &2.4098  &9.5621 &{\bf 0.2186} &158.8559 &12.7693 \\
			\Xhline{1 pt}
			$2\sqrt{\log{d}/n} $ & 0.1725  &{\bf 0.0213} &2.2307 &2.2272  &8.6883 &{\bf 0.1603} &129.3729 &20.4183 \\
			\Xhline{1 pt}
			$4\sqrt{\log{d}/n} $ & 0.0478  &{\bf 0.0112} &1.8683 &1.3656  &1.8236 &{\bf 0.0924} &94.3733 &19.1710 \\
			\Xhline{1 pt}
	\end{tabular}}\label{table:pissta_admm_smr}
\end{center}
\end{table*}

\noindent{\bf Synthetic Data:} For synthetic data, we generate a training dataset of $200$ samples, where each row of the design matrix $ X_{i*}$is independently from a $2000$-dimensional normal distribution $N( 0,\Sigma)$ where $\Sigma_{jj} = 1$ and $\Sigma_{jk} = 0.5$ for all $k\neq j$. We set $s^*=3$ with $\theta_1^*=3$, $\theta_2^*=-2$, and $\theta_4^*=1.5$, and $\theta_j^*=0$ for all $j\neq 1,2,4$. The response vector is generated by $ y= X \theta^* + \epsilon$, where $\epsilon$ is sampled from $N( 0,\sigma^2 I)$.

We first show the {\bf fast convergence} of the proximal algorithms at {\bf every stage} of the pathwise optimization scheme. Here we set $\sigma = 0.5$, $N=200$,  $\lambda_N = \sqrt{\log{d}/n}$, $\varepsilon_K = 10^{-6}$ for all $K=1,\ldots,N$. Figure~\ref{fig:fast_cvg} presents the objective gap versus the number of iterations. We can see that the proximal algorithms achieves linear (prox-GD) and quadratic (prox-Newton) convergence at every stage. Since the solution sparsity levels are different at each stage, the slopes of these curves are also different.


Next, we show that the computational performance of the pathwise optimization scheme under different settings. Table~\ref{table:pista_ista_cmp} presents the timing performance of Prox-GD combined with the pathwise optimization scheme. We can see that $N=10$ actually leads to better timing performance than $N=1$. That is because when $N=1$, the solution path does not fall into the local fast convergence region as illustrated in Figure~\ref{fig:ball}. We can also see that the timing performance of Prox-GD is not sensitive to $\sigma$. Moreover, we see that the minimal residual sum of squares along the solution path is much larger than 0, thus the overfitting is prevented and the Prox-GD algorithm enjoys the smoothness of the $\ell_2$ loss.




\noindent{\bf Real Data}: We adopt two data sets. The first one is the Greenhouse Gas Observing Network Data Set \cite{gi-4-121-2015}, which contains $2921$ samples and $5232$ variables. The second one is the DrivFace data set, which contains 606 samples and $6400$ variables. We compare our proximal algorithms with ADMM in \cite{li2015flare}, Coordinate Descent (CD) in \cite{ndiaye2016efficient}, Prox-GD (solving Lasso) in \cite{xiao2013proximal} and Alternating Minimization (Alt.Min.) \cite{sun2012scaled} and ScalReg (a simple variant of Alt. Min) in \cite{sun2013package}. Table~\ref{table:realdata_exp} presents the timing performance of the different algorithms. We can see that Prox-GD for solving SQRT-Lasso significantly outperforms the competitors, and is almost as efficient as Prox-GD for solving Lasso. Prox-Newton is even more efficient than Prox-GD.

%
%
%


\noindent{\bf Sparse Precision Matrix Estimation}. We compare the proximal algorithms with ADMM and CD over real data sets for precision matrix estimation. Particularly,  we use four real world biology data sets preprocessed by \cite{li2010inexact}: Arabidopsis ($d=834$), Lymph ($d=587$), Estrogen ($d = 692$), Leukemia ($d = 1,225$). We set three different values for $\lambda_N$ such that the obtained estimators achieve different levels of sparse recovery. We set $N=10$, and $\varepsilon_{K} = 10^{-4}$ for all $K$'s. The timing performance is summarized in Table \ref{table:pissta_admm_spme}. Prox-GD for solving SQRT-Lasso significantly outperforms the competitors, and is almost as efficient as Prox-GD for solving Lasso. Prox-Newton is even more efficient than Prox-GD.


\noindent{\bf Calibrated Multivariate Regression}. We compare the proximal algorithms with ADMM and CD for CMR on both synthetic data and DrivFace data. For synthetic data, the data generating scheme is the same as \cite{liu2015calibrated}. Table \ref{table:pissta_admm_smr} presents the timing performance. Prox-GD for solving SQRT-Lasso significantly outperforms the competitors, and is almost as efficient as Prox-GD for solving Lasso. Prox-Newton is even more efficient than Prox-GD. CD failed to converge and the program aborted before reaching the desired suboptimality.

\section{Discussion and Conclusions}\label{sec:discuss}



This paper shows that although the loss function in the SQRT-Lasso optimization problem is nonsmooth, we can directly apply the proximal gradient and Newton algorithms with fast convergence. First, the fast convergence rate can be established locally in a neighborhood of $\theta^*$. Note that, due to the limited analytical tools, we are not able to directly extend the analysis to establish a global fast convergence rate. Instead, we resort to the pathwise optimization scheme, which helps establishing empirical global fast convergence for the proximal algorithms as illustrated in Figure~\ref{fig:ball}. Specifically, in the early stage of pathwise scheme, with a large regularization parameter $\lambda$, the solution quickly falls into the neighborhood of $\theta^*$, where the problem enjoys good properties. After that, the algorithm can quickly converges to $\theta^*$ thanks to the fast local convergence property. Our results corroborate that exploiting modeling structures of machine learning problems is of great importance from both computational and statistical perspectives.

Moreover, we remark that to establish the local fast convergence rate, we prove the restricted strong convexity, smoothness, and Hessian smoothness hold over a neighborhood of $\theta^*$. Rigorously establishing the global fast convergence, however, requires these conditions to hold along the solution path. We conjecture that these conditions do hold because our empirical results show the proximal algorithms indeed achieve fast convergence along the entire solution path of the pathwise optimization. We will look for more powerful analytic tools and defer a sharper characterization to the future effort.
	
	\newpage
	{
		\bibliographystyle{ims}
		\bibliography{../Library}
	}
	
	\newpage
	
	\onecolumn
	\appendix
	

\section{Proof of Lemma~\ref{lem:rsc}}\label{pf:lem:rsc}

\textbf{Part 1}. We first show the claim on $\lambda$.
By $ y =  X \theta^* + \epsilon$ and \eqref{eqn:grad-smooth}, we have
\begin{align}
\nabla \cL(\theta^*) = \frac{ X^\top ( X \theta^* -  y)}{\sqrt{n}\norm{ y -  X \theta^*}_2} = -\frac{ X^\top \epsilon}{\sqrt{n}\norm{\epsilon}_2}. \label{eqn:gradL}
\end{align}
Since $\epsilon$ has i.i.d. sub-Gaussian entries with $\EE[\epsilon_i]=0$ and $\EE[\epsilon_i^2] = \sigma^2$ for all $i = 1,\ldots,n$, then from \cite{wainwright2015high} we have 
\begin{align}
\PP \left[ \norm{\epsilon}_2^2 \leq \frac{1}{4}n \sigma^2 \right] \leq \exp \left(-\frac{n}{32} \right), \label{eqn:subgaus5}
\end{align}

By \cite{negahban2012unified}, we have the following result. 
\begin{lemma}\label{lem:infgdbd}
	Assume $ X$ satisfies $\norm{\xb_j}_2 \leq \sqrt{n}$ for all $j \in \{1,\ldots,d\}$ and $\epsilon$ has i.i.d. zero-mean sub-Gaussian entries with $\EE[w_i^2] = \sigma^2$ for all $i = 1,\ldots,n$, then we have  $\PP\left[ \frac{1}{n} \norm{ X^\top \epsilon}_{\infty} \geq 2 \sigma \sqrt{\frac{\log d}{n}} \right] \leq 2 d^{-1}$.
\end{lemma}

Combining \eqref{eqn:gradL}, \eqref{eqn:subgaus5} and Lemma~\ref{lem:infgdbd}, we have $\norm{\nabla \cL(\theta^*)}_{\infty} \leq 4\sqrt{{\log d}/{n}}$ with probability at least $1-2 d^{-1}-\exp \left(-\frac{n}{32} \right)$.

\textbf{Part 2}. Next, we show that LRSC, LRSS, and LRHS holds. First, for correlated sub-Gaussian random design with the covariance satisfying the bounded eigenvalues, we have from \citep{Rudelson2013}
that the design matrix $ X$ satisfies the RE condition with high probability given $n \geq c s^* \log d$, i.e., for any $ v \in \cB^{s^*+\tilde{s}}_r = \cB_{r} \cap \{ \theta \in \RR^d : \norm{\theta - \theta^*}_0 \leq s^*+\tilde{s} \}$,
\begin{align}
\psi_{\min} \|  v \|_2^2 - \varphi_{\min} \frac{\log d}{n} \|  v \|_1^2 \leq \frac{\|  X v \|_2^2}{n} \leq
\psi_{\max} \|  v \|_2^2 + \varphi_{\max} \frac{\log d}{n} \|  v \|_1^2 ,  \label{eqn:RE_X}
\end{align}
where $\psi_{\min}, \psi_{\max}, \varphi_{\min}, \varphi_{\max} \in (0,\infty)$ are generic constants. The RE condition has been extensively studied for sparse recovery \citep{candes2005decoding,bickel2009simultaneous,raskutti2010restricted}.

We divide the proof into three steps. 

\textbf{Step 1}. When $ X$ satisfies the RE condition, i.e.
\begin{align*}
\psi_{\min} \|  v \|_2^2 - \varphi_{\min} \frac{\log d}{n} \|  v \|_1^2 &\leq \frac{\|  X v \|_2^2}{n} \\  
\psi_{\max} \|  v \|_2^2 + \varphi_{\max} \frac{\log d}{n} \|  v \|_1^2 &\geq \frac{\|  X v \|_2^2}{n},
\end{align*}
Denote $s = s^* + 2\tilde{s}$. Since $\norm{ v}_0 \leq s$, which implies $\norm{ v}_1^2 \leq s \norm{ v}_2^2$, then we have
\begin{align*}
\left(\psi_{\min} - \varphi_{\min} \frac{s\log d}{n} \right) \|  v \|_2^2 &\leq \frac{\|  X v \|_2^2}{n}   \\ 
\left(\psi_{\max} + \varphi_{\max} \frac{s\log d}{n} \right) \|  v \|_2^2 &\geq \frac{\|  X v \|_2^2}{n},
\end{align*}
Then there exists a universal constant $c_1$ such that if $n \geq c_1 s^*\log d$, we have
\begin{align}
\frac{1}{2}\psi_{\min} \|  v \|_2^2 \leq \frac{\|  X v \|_2^2}{n} \leq 2 \psi_{\max} \|  v \|_2^2. \label{eqn:SE}
\end{align}

\textbf{Step 2}. Conditioning on \eqref{eqn:SE}, we show that $\cL$ satisfies LRSC and LRSS with high probability. The gradient of $\cL(\theta)$ is
\begin{align}
\nabla \cL(\theta) &= \frac{1}{\sqrt{n}}\left( \left(\frac{\partial\norm{ y- X\theta}_2}{\partial( y- X\theta)}\right)^\top \left(\frac{\partial( y- X\theta)}{\partial\theta}\right)^\top \right)^\top = \frac{ X^\top ( X \theta -  y)}{\sqrt{n}\norm{ y -  X \theta}_2}. \label{eqn:grad-smooth}
\end{align}
The Hessian of $\cL(\theta)$ is
\begin{align}
&\nabla^2 \cL(\theta) = \frac{1}{n}\frac{\partial(- X^\top \tilde{ z})}{\partial\theta} = \frac{1}{\sqrt{n}\norm{ y- X\theta}_2}  X^\top \left(  I - \frac{( y -  X \theta)( y -  X \theta)^\top}{\norm{ y- X\theta}_2^2} \right)  X. \label{eqn:hess-smooth}
\end{align}

For notational convenience, we define $\Delta =  v -  w$ for any $ v,  w \in \cB^{s^*+\tilde{s}}_r$. Also denote the residual of the first order Taylor expansion as $\delta \cL( w+\Delta,  w) = \cL ( w+\Delta) - \cL ( w) - \nabla \cL( w)^\top \Delta$. Using the first order Taylor expansion of $\cL (\theta)$ at $ w$ and the Hessian of $\cL(\theta)$ in \eqref{eqn:hess-smooth}, we have from mean value theorem that there exists some $\alpha \in [0,1]$ such that $\delta \cL( w+\Delta,  w) = \frac{1}{\sqrt{n}\norm{\xi}_2} \Delta^\top  X^\top \left(  I - \frac{\xi \xi^\top}{\norm{\xi}_2^2} \right)  X \Delta$, where $\xi =  y- X( w+\alpha\Delta)$. For notational simplicity, let`s denote $\dot{ z} =  X ( v-\theta^*)$ and $\ddot{ z} =  X ( w-\theta^*)$, which can be considered as two fixed vectors in $\RR^n$. Without loss of generality, assume $\norm{\dot{ z}}_2 \leq \norm{\ddot{ z}}_2$. Then we have
\begin{align*}
\norm{\dot{ z}}_2^2 \leq \norm{\ddot{ z}}_2^2 \leq 2 \psi_{\max} n \norm{ w-\theta^*}_2^2 \leq \frac{n \sigma^2}{4}.
\end{align*}
Further, we have
\begin{align*}
\xi &=  y- X( w+\alpha\Delta) = \epsilon -  X( w+\alpha\Delta-\theta^*) = \epsilon - \alpha \dot{ z} - (1-\alpha) \ddot{ z}, ~~\text{and}~~  X \Delta = \dot{ z} - \ddot{ z}.
\end{align*}

We have from \cite{wainwright2015high} that
\begin{align}
\PP \left[ \norm{\epsilon}_2^2 \leq n \sigma^2 (1-\delta) \right] &\leq \exp \left(-\frac{n \delta^2}{16} \right), \label{eqn:subgaus7}
\end{align}
Then by taking $\delta=1/3$ in \eqref{eqn:subgaus7}, we have with probability $1-\exp\left(-\frac{n}{144}\right)$,
\begin{align}
\norm{\xi}_2 &\geq \norm{\epsilon}_2 - \alpha\norm{\dot{ z}}_2 - (1-\alpha)\norm{\ddot{ z}}_2 {\geq} \norm{\epsilon}_2 - \norm{\ddot{ z}}_2 {\geq} \frac{4}{5}\sqrt{n} \sigma - \frac{1}{2}\sqrt{n} \sigma \geq \frac{1}{4}\sqrt{n} \sigma. \label{eqn:xi_lb}
\end{align}

We first discuss the RSS property. From \eqref{eqn:xi_lb}, we have
\begin{align*}
\delta \cL( w+\Delta,  w) = \frac{\Delta^\top  X^\top \left(  I - \frac{\xi \xi^\top}{\norm{\xi}_2^2} \right)  X \Delta}{\sqrt{n}\norm{\xi}_2} = \frac{\left( \|  X \Delta \|_2^2 - \frac{(\xi^\top  X \Delta)^2}{\norm{\xi}_2^2} \right)}{\sqrt{n}\norm{\xi}_2} \leq \frac{\|  X \Delta \|_2^2}{\sqrt{n}\norm{\xi}_2} \leq \frac{8 \psi_{\max}}{\sigma} \| \Delta \|_2^2
\end{align*}

Next, we verify the RSC property. We want to show that with high probability, for any constant $a \in (0,3/5]$
\begin{align}
\left| \frac{\xi^\top}{\norm{\xi}_2}  X \Delta \right| \leq \sqrt{1-a} \norm{ X \Delta}_2. \label{eqn:case2bd}
\end{align}
Consequently, we have
\begin{align*}
&\Delta^\top  X^\top \left(  I - \frac{\xi \xi^\top}{\norm{\xi}_2^2} \right)  X \Delta = \norm{ X \Delta}_2^2 - \left( \frac{\xi^\top}{\norm{\xi}_2}  X \Delta \right)^2 \geq a \norm{ X \Delta}_2^2.
\end{align*}
This further implies
\begin{align}
\delta \cL( w+\Delta,  w) = \frac{1}{\sqrt{n}\norm{\xi}_2} \Delta^\top  X^\top \left(  I - \frac{\xi \xi^\top}{\norm{\xi}_2^2} \right)  X \Delta \geq \frac{a \psi_{\min}}{2\norm{\xi}_2/\sqrt{n}} \| \Delta \|_2^2. \label{eqn:case2}
\end{align}

Since $\norm{\dot{ z}}_2 \leq \norm{\ddot{ z}}_2$, then for any real constant $a \in (0,1)$,
\begin{align}
&\PP \left[ \left| \frac{\xi^\top}{\norm{\xi}_2}  X \Delta \right| \leq \sqrt{1-a} \norm{ X \Delta}_2 \right] = \PP \left[ \left| \frac{(\epsilon - \alpha \dot{ z} - (1-\alpha) \ddot{ z})^\top}{\norm{\epsilon - \alpha \dot{ z} - (1-\alpha) \ddot{ z}}_2} (\dot{ z} - \ddot{ z}) \right| \leq \sqrt{1-a} \norm{\dot{ z} - \ddot{ z}}_2 \right] \nonumber\\
&\overset{(i)}{\geq} \PP \left[ \left| \frac{(\epsilon - \dot{ z})^\top(\dot{ z} - \ddot{ z})}{\norm{\epsilon - \dot{ z}}_2} \right| \leq \sqrt{1-a} \norm{\dot{ z} - \ddot{ z}}_2 \right] = \PP \left[ \left(\epsilon^\top \hspace{-0.02in}(\dot{ z} - \ddot{ z}) \hspace{-0.02in}-\hspace{-0.02in} \dot{ z}^\top (\dot{ z} - \ddot{ z}) \right)^2 \hspace{-0.06in}\leq\hspace{-0.03in} (1-a) \norm{\epsilon - \dot{ z}}_2^2 \norm{\dot{ z} - \ddot{ z}}_2^2 \right] \nonumber \\
& \overset{(ii)}{\geq} \PP \left[  \left| \left(\frac{\epsilon^\top (\dot{ z} - \ddot{ z})}{\norm{\dot{ z} - \ddot{ z}}_2} \right)^2 + \norm{\dot{ z}}_2^2 - \frac{2 \epsilon^\top (\dot{ z} - \ddot{ z})  z^\top (\dot{ z} - \ddot{ z}) }{\norm{\dot{ z} - \ddot{ z}}_2^2} \right| 
\leq (1-a)(\norm{\epsilon}_2^2 + \norm{\dot{ z}}_2^2 - 2\epsilon^\top \dot{ z}) \right], \label{eqn:probbd1}
\end{align}
where (i) is from a geometric inspection and the randomness of $\epsilon$, i.e., for any $\alpha \in [0,1]$ and $\norm{\dot{ z}}_2 \leq \norm{\ddot{ z}}_2$, we have $\left|\frac{-\dot{ z}^\top}{\norm{-\dot{ z}}_2}(\dot{ z} - \ddot{ z}) \right| \leq \left|\frac{(-\alpha \dot{ z} - (1-\alpha) \ddot{ z})^\top}{\norm{-\alpha \dot{ z} - (1-\alpha) \ddot{ z}}_2}(\dot{ z} - \ddot{ z}) \right|$, and (ii) is from dividing both sides by $\norm{\dot{ z} - \ddot{ z}}_2^2$. The random vector $\epsilon$ with i.i.d. entries does not affect the inequality above. Let`s first discuss one side of the probability in \eqref{eqn:probbd1},
\begin{align}
&\PP \left[  \left(\frac{\epsilon^\top (\dot{ z} - \ddot{ z})}{\norm{\dot{ z} - \ddot{ z}}_2} \right)^2 + \norm{\dot{ z}}_2^2 - \frac{2 \epsilon^\top (\dot{ z} - \ddot{ z})  z^\top (\dot{ z} - \ddot{ z}) }{\norm{\dot{ z} - \ddot{ z}}_2^2} 
\leq (1-a)(\norm{\epsilon}_2^2 + \norm{\dot{ z}}_2^2 - 2\epsilon^\top \dot{ z}) \right] \nonumber\\
&= \PP \left[(1-a) \norm{\epsilon}_2^2 \geq \left(\frac{\epsilon^\top (\dot{ z} - \ddot{ z})}{\norm{\dot{ z} - \ddot{ z}}_2} \right)^2 - 2(1-a)\epsilon^\top \dot{ z} + a\norm{\dot{ z}}_2^2 + \frac{2 \epsilon^\top (\dot{ z} - \ddot{ z})  z^\top (\dot{ z} - \ddot{ z}) }{\norm{\dot{ z} - \ddot{ z}}_2^2} \right]. \label{eqn:bd1}
\end{align}

Since $\epsilon$ has i.i.d. sub-Gaussian entries with $\EE[\epsilon_i]=0$ and $\EE[\epsilon_i^2] = \sigma^2$ for all $i = 1,\ldots,n$, then $\frac{\epsilon^\top (\dot{ z} - \ddot{ z})}{\norm{\dot{ z} - \ddot{ z}}_2}$, $\epsilon^\top \dot{ z}$, and $\frac{\epsilon^\top (\dot{ z} - \ddot{ z})  z^\top (\dot{ z} - \ddot{ z}) }{\norm{\dot{ z} - \ddot{ z}}_2^2}$ are also zero-mean sub-Gaussians with variances $\sigma^2$, $\sigma^2 \norm{\dot{ z}}_2^2$, and  $\sigma^2 \norm{\dot{ z}}_2^2$ respectively. We have from \cite{wainwright2015high} that\\ 
\begin{align}
&\PP \left[ \norm{\epsilon}_2^2 \leq n \sigma^2 (1-\delta) \right] \leq \exp \left(-\frac{n \delta^2}{16} \right), \PP\left[ \left(\frac{\epsilon^\top (\dot{ z} - \ddot{ z})}{\norm{\dot{ z} - \ddot{ z}}_2} \right)^2 \geq n \sigma^2 \delta \right] \leq \exp \left( -\frac{n \delta^2}{2} \right), \label{eqn:subgaus2} \\
&\PP\left[ \epsilon^\top \dot{ z} \leq -n\sigma^2 \delta \right] \leq \exp \left( -\frac{n^2 \sigma^2 \delta^2}{2 \norm{\dot{ z}}_2^2} \right), \PP\left[ \frac{\epsilon^\top (\dot{ z} - \ddot{ z})  z^\top (\dot{ z} - \ddot{ z}) }{\norm{\dot{ z} - \ddot{ z}}_2^2} \geq n\sigma^2 \delta \right] \leq \exp \left( -\frac{n^2 \sigma^2 \delta^2}{2 \norm{\dot{ z}}_2^2} \right). \nonumber
\end{align}

Combining \eqref{eqn:subgaus2} with $\norm{\dot{ z}}_2^2 \leq n \sigma^2/4$, we have from union bound that with probability at least $1-\exp\left( -\frac{n}{144} \right) - \exp\left( -\frac{n}{128} \right) - \exp\left( -\frac{n}{128} \right) \geq 1-3\exp\left( -\frac{n}{144} \right)$,
\begin{align*}
&\norm{\epsilon}_2^2 \geq \frac{2}{3}n \sigma^2,~\left(\frac{\epsilon^\top (\dot{ z} - \ddot{ z})}{\norm{\dot{ z} - \ddot{ z}}_2} \right)^2 \leq \frac{1}{64} n \sigma^2,~-\epsilon^\top \dot{ z} \leq \frac{1}{16} n\sigma^2,\frac{\epsilon^\top (\dot{ z} - \ddot{ z})  z^\top (\dot{ z} - \ddot{ z}) }{\norm{\dot{ z} - \ddot{ z}}_2^2} \leq \frac{1}{16} n\sigma^2.
\end{align*}
This implies for $a \leq 3/5$, we have $\frac{\xi^\top}{\norm{\xi}_2}  X \Delta \leq  \sqrt{1-a} \norm{ X \Delta}_2$. For the other side of \eqref{eqn:probbd1}, we have
\begin{align}
&\PP \left[ \begin{matrix}\left(\frac{\epsilon^\top (\dot{ z} - \ddot{ z})}{\norm{\dot{ z} - \ddot{ z}}_2} \right)^2 + \norm{\dot{ z}}_2^2 - 2\epsilon^\top \dot{ z} \geq \\
-(1-a)(\norm{\epsilon}_2^2 + \norm{\dot{ z}}_2^2 - 2\epsilon^\top \dot{ z}) \end{matrix}\right] \nonumber\\
&\overset{(i)}{\geq} \PP \left[ \begin{matrix}-\left(\frac{\epsilon^\top (\dot{ z} - \ddot{ z})}{\norm{\dot{ z} - \ddot{ z}}_2} \right)^2 -( \norm{\dot{ z}}_2^2 - 2\epsilon^\top \dot{ z}) \geq \\
-(1-a)(\norm{\epsilon}_2^2 + \norm{\dot{ z}}_2^2 - 2\epsilon^\top \dot{ z}) \end{matrix}\right] 
= \PP \left[ (1-a) \norm{\epsilon}_2^2 \geq \left(\frac{\epsilon^\top (\dot{ z} - \ddot{ z})}{\norm{\dot{ z} - \ddot{ z}}_2} \right)^2 + a(\norm{\dot{ z}}_2^2 - 2\epsilon^\top \dot{ z}) \right]. \label{eqn:bd2}
\end{align}
where (i) is from the fact that $\PP[c_1 \geq -c_2] \geq \PP[-c_1 \geq -c_2]$ for $c_1,c_2 \geq 0$.

Combining \eqref{eqn:probbd1}, \eqref{eqn:bd1} and \eqref{eqn:bd2}, we have \eqref{eqn:case2bd} holds with high probability, i.e., for any ${r} > 0$,
\begin{align*}
\PP \left[ \left| \frac{\xi^\top}{\norm{\xi}_2}  X \Delta \right| \leq \sqrt{1-a} \norm{ X \Delta}_2 \right] \geq 1-6\exp\left( -\frac{n}{144} \right).
\end{align*}

Now we bound $\norm{\xi}_2$ to obtain the desired result. From \cite{wainwright2015high}, we have
\begin{align}
\PP \left[ \norm{\epsilon}_2^2 \geq n \sigma^2(1+\delta) \right] \leq \exp \left(-\frac{n \delta^2}{18} \right) = \exp \left(-\frac{n}{72} \right), \label{eqn:subgaus4} 
\end{align}
where we take $\delta = 1/2$. From  $\xi = \epsilon - \alpha \dot{ z} - (1-\alpha) \ddot{ z}$, we have
\begin{align}
\norm{\xi}_2 &\leq \norm{\epsilon}_2 + \alpha\norm{\dot{ z}}_2 + (1-\alpha)\norm{\ddot{ z}}_2 \overset{(i)}{\leq} \norm{\epsilon}_2 + \norm{\ddot{ z}}_2 \overset{(ii)}{\leq}  \sqrt{\frac{3n}{2}} \sigma + \frac{1}{2}\sqrt{n} \sigma.
\end{align}
where (i) is from $\norm{\dot{ z}}_2 \leq \norm{\ddot{ z}}_2$ and (ii) is from \eqref{eqn:subgaus4} and $\norm{\dot{ z}}_2^2 \leq n \sigma^2/4$. Then by the union bound setting $a=1/2$, with probability at least $1-7\exp\left(-\frac{n}{144}\right)$, we have $\delta \cL( w+\Delta,  w) \geq  \frac{\psi_{\min}}{8 \sigma} \| \Delta \|_2^2$. Moreover, we also have $r = \frac{\sigma^2}{8\psi_{\max}} $ for large enough  $n \geq c_1 s^*\log d$. 

\textbf{Step 3}. Given the proposed conditions, we have that $\cL$ satisfies the LRHS property by combining the analysis in \cite{ning2014likelihood}.

\section{Intermediate Results of Theorem~\ref{thm:proxgd}}\label{int:thm:linearmodel}

We introduce some important implications of the proposed assumptions. Recall that $\cS^* = \{ j : \theta^*_{j} \neq 0 \}$ be the index set of non-zero entries of $\theta^*$ with $s^* = |\cS^*|$ and $\overline{\cS}^* = \{ j : \theta^*_{j} = 0 \}$ be the complement set. Lemma~\ref{lem:rsc} implies RSC and RSS hold with parameter $\rho^{-}_{s^*+2\tilde{s}}$ and $\rho^{+}_{s^*+2\tilde{s}}$ respectively. By \cite{nesterov2004introductory},  the following conditions are equivalent to RSC and RSS, i.e., for any $ v,  w \in \RR^d$ satisfying $\norm{ v -  w}_0 \leq s^* + 2\tilde{s}$,
\begin{align}
\rho^{-}_{s^*+2\tilde{s}} \norm{ v -  w}_2^2 \leq ( v -  w)^\top \nabla \cL( w) &~~\text{and}~~ \rho^{+}_{s^*+2\tilde{s}} \norm{ v -  w}_2^2 \geq ( v -  w)^\top \nabla \cL( w), \label{eqn:rsc2} \\
\frac{1}{\rho^{+}_{s^*+2\tilde{s}}} \norm{\nabla \cL( v) - \nabla \cL( w)}_2^2 \leq ( v -  w)^\top \nabla \cL( w) &~~\text{and}~~ \frac{1}{\rho^{-}_{s^*+2\tilde{s}}} \norm{\nabla \cL( v) - \nabla \cL( w)}_2^2 \geq ( v -  w)^\top \nabla \cL( w) . \label{eqn:rsc3}
\end{align}
From the convexity of $\ell_1$ norm, we have
\begin{align}
\norm{ v}_1 - \norm{ w}_1 \geq ( v -  w)^\top  g, \label{eqn:cvxl1}
\end{align}
where $ g \in \partial \norm{ w}_1$. Combining and \eqref{eqn:rsc2} and \eqref{eqn:cvxl1}, we have for any $ v,  w \in \RR^d$ satisfying $\norm{ v -  w}_0 \leq s^* + 2\tilde{s}$,
\begin{align}
&\cF_{\lambda}( v) - \cF_{\lambda}( w) - ( v -  w)^\top \nabla \cF_{\lambda}( w) \geq \rho^{-}_{s^*+2\tilde{s}} \norm{ v -  w}_2^2, \label{eqn:rsc9}
\end{align}

\begin{remark}\label{remark:stepsize}
	For any $t$ and $k$, the line search satisfies
	\begin{align}
	&\tilde{L}^{(t)} \leq L^{(t)} \leq L_{\max} ,~ L_{} \leq \tilde{L}^{(t)} \leq L^{(t)} \leq 2 L_{}~~\text{and}~~\rho^{+}_{s^*+2\tilde{s}} \leq \tilde{L}^{(t)} \leq L^{(t)} \leq 2 \rho^{+}_{s^*+2\tilde{s}},\label{eqn:Lbd2}
	\end{align}
	where $L_{} = \min \{ L: \norm{\nabla \cL( v) - \nabla \cL( w)}_2 \leq L \norm{\bx -  y}_2, \forall  v, w \in \RR^d \}$.
\end{remark}

We first show that when $\theta$ is sparse and the approximate KKT condition is satisfied, then both estimation error and objective error, w.r.t. the true model parameter, are bounded. This is formalized in Lemma~\ref{lem:step1}, and its proof is deferred to Appendix~\ref{pf:lem:step1}.
\begin{lemma}\label{lem:step1}
	Suppose conditions in Lemma~\ref{lem:rsc} hold and $s^*\left({8 \lambda}/{\rho^{-}_{s^*+\tilde{s}}}\right)^2 < r$. If $\theta$ satisfies $\norm{\theta_{\overline{\cS}^*}}_0 \leq \tilde{s}$ and the approximate KKT condition $\min_{ g \in \partial \norm{\theta}_1}  \norm{\nabla \cL (\theta) + \lambda  g}_{\infty} \leq \lambda/2$, then we have\\ 
	\begin{align}
	\norm{(\theta - \theta^*)_{\overline{\cS}^*}}_1 &\leq 5 \norm{(\theta - \theta^*)_{\cS^*}}_1, \label{eqn:l1Scbd}\\
	\norm{\theta - \theta^*}_2 &\leq \frac{2 \lambda \sqrt{s^*} }{\rho^{-}_{s^*+\tilde{s}}} \leq \frac{2 \lambda \sqrt{s^*} }{\rho^{-}_{s^*+2\tilde{s}}}, \label{eqn:l2bd} \\
	\norm{\theta - \theta^*}_1&\leq \frac{12 \lambda s^*}{\rho^{-}_{s^*+\tilde{s}}} \leq \frac{12 \lambda s^*}{\rho^{-}_{s^*+2\tilde{s}}}, \label{eqn:l1bd}\\
	\cF_{\lambda} (\theta) - \cF_{\lambda} (\theta^*) &\leq \frac{6 \lambda^2 s^*}{\rho^{-}_{s^*+\tilde{s}}}\leq \frac{6 \lambda^2 s^*}{\rho^{-}_{s^*+2\tilde{s}}}. \label{eqn:objbd}
	\end{align}
\end{lemma}

Next, we show that if $\theta$ is sparse and the objective error is bounded, then the estimation error is also bounded. This is formalized in Lemma~\ref{lem:step2}, and its proof is deferred to Appendix~\ref{pf:lem:step2}.
\begin{lemma}\label{lem:step2}
	Suppose conditions in Lemma~\ref{lem:rsc} hold and $s^*\left({8 \lambda}/{\rho^{-}_{s^*+\tilde{s}}}\right)^2 < r$. If $\theta$ satisfies $\norm{\theta_{\overline{\cS}^*}}_0 \leq \tilde{s}$ and the objective satisfies $\cF_{\lambda} (\theta) - \cF_{\lambda} (\theta^*) \leq \frac{6 \lambda^2 s^*}{\rho^{-}}$, where $\rho^{-}$ can be either $\rho^{-}_{s^*+\tilde{s}}$ or $\rho^{-}_{s^*+2\tilde{s}}$, then we have\\
	\begin{align}
	\norm{\theta - \theta^*}_2 &\leq \frac{4\lambda \sqrt{3s^*}}{\rho^{-}}, \label{eqn:l2bd2}\\
	\norm{\theta - \theta^*}_1 &\leq \frac{24\lambda s^*}{\rho^{-}}. \label{eqn:l1bd2}
	\end{align}
\end{lemma}

We then show that if $\theta$ is sparse and the objective error is bounded, then each proximal-gradient update preserves solution to be sparse. This is formalized in Lemma~\ref{lem:step3}, and its proof is deferred to Appendix~\ref{pf:lem:step3}.
\begin{lemma}\label{lem:step3}
	Suppose conditions in Lemma~\ref{lem:rsc} hold and $s^*\left({8 \lambda}/{\rho^{-}_{s^*+\tilde{s}}}\right)^2 < r$. If $\theta$ satisfies $\norm{\theta_{\overline{\cS}^*}}_0 \leq \tilde{s}$, ${L}$ satisfies ${L}<2\rho^{+}_{s^*+2\tilde{s}}$, and the objective satisfies $\cF_{\lambda} (\theta) - \cF_{\lambda} (\theta^*) \leq \frac{6 \lambda^2 s^*}{\rho^{-}_{s^*+2\tilde{s}}}$, then we have $\norm{\left(\cT_{L,\lambda}(\theta)\right)_{\overline{\cS}^*}}_0 \leq \tilde{s}$.
\end{lemma}

Moreover, we show that if $\theta$ satisfies the approximate KKT condition, then the objective has a bounded error w.r.t. the regularizaSuppose conditions in Lemma~\ref{lem:rsc} with parameter $\lambda$. This characterizes the geometric decrease of the objective error when we choose a geometrically decreasing sequence of regularization parameters. This is formalized in Lemma~\ref{lem:objbd}, and its proof is deferred to Appendix~\ref{pf:lem:objbd}.

\begin{lemma}\label{lem:objbd}
	. If $\theta$ satisfies $\omega_{\lambda} (\theta) \leq \lambda/2$, then for $\overline{\theta} = \argmin_{\theta} \cF_{{\lambda}} (\theta)$, we have $\cF_{{\lambda}} (\theta) - \cF_{{\lambda}} (\overline{\theta}) \leq \frac{24 \lambda \omega_{\lambda}(\theta) {s^*} }{\rho^{-}_{s^*+2\tilde{s}}}$.
\end{lemma}

Furthermore, we show a local linear convergence rate if the initial value $\theta^{(0)}$ is sparse and satisfies the approximate KKT condition with adequate precision. Besides, the estimation after each proximal gradient update is also sparse. This is the key result in demonstrating the overall geometric convergence rate of the algorithm. This is formalized in Lemma~\ref{lem:controlsp}, and its proof is deferred to Appendix~\ref{pf:lem:controlsp}.

\begin{lemma}\label{lem:controlsp}
	Suppose conditions in Lemma~\ref{lem:rsc} hold and $s^*\left({8 \lambda}/{\rho^{-}_{s^*+\tilde{s}}}\right)^2 < r$s. If the initialization $\theta^{(0)}$ satisfies $\norm{\theta^{(0)}}_0 \leq \tilde{s}$. Then with  $\overline{\theta} = \argmin_{\theta} \cF_{\lambda}(\theta)$, for any $t=1,2,\ldots$, we have $\norm{\theta^{(t)}}_0 \leq \tilde{s}$ and $\cF_{\lambda}(\theta^{(t)}) - \cF_{\lambda} (\overline{\theta}) \leq \left( 1 - \frac{1}{8{\kappa}_{s^*+2\tilde{s}}} \right)^t \left( \cF_{\lambda}(\theta^{(0)}) - \cF_{\lambda} (\overline{\theta}) \right)$.
\end{lemma}

Finally, we introduce two results characterizing the proximal gradient mapping operation, adapted from \cite{nesterov2013gradient} and \cite{xiao2013proximal} without proof. The first lemma describes sufficient descent of the objective by proximal gradient method.
\begin{lemma}[Adapted from Theorem 2 in \cite{nesterov2013gradient}]\label{lem:suf_des_obj}
	For any $L>0$, 
	\begin{align*}
	\cQ_{\lambda} \left(\cT_{L,\lambda}(\theta),\theta \right) \leq \cF_{\lambda} \left( \theta \right) - \frac{L}{2} \norm{\cT_{L,\lambda}(\theta) - \theta}_2^2. 
	\end{align*}
	Besides, if $\cL(\theta)$ is convex, we have
	\begin{align}
	\cQ_{\lambda} \left(\cT_{L,\lambda}(\theta),\theta \right) \leq \min_{\xb} \cF_{\lambda} \left( \xb \right) + \frac{L}{2} \norm{\xb - \theta}_2^2. \label{eqn:proxobjbd2}
	\end{align}
	Further, we have for any $L \geq L_{}$,
	\begin{align}
	\cF_{\lambda} \left( \cT_{L,\lambda}(\theta) \right) \leq \cQ_{\lambda} \left(\cT_{L,\lambda}(\theta),\theta \right) \leq \cF_{\lambda} \left( \theta \right) - \frac{L}{2} \norm{\cT_{L,\lambda}(\theta) - \theta}_2^2. \label{eqn:proxobjbd1}
	\end{align}
\end{lemma}

The next lemma provides an upper bound of the optimal residue $\omega(\cdot)$. 
\begin{lemma}[Adapted from Lemma 2 in \cite{xiao2013proximal}]\label{lem:upbd_omg}
	For any $L>0$, if $L_{}$ is the Lipschitz constant of $\nabla\cL$, then
	\begin{align*}
	\omega_{\lambda} \left( \cT_{L,\lambda}(\theta) \right) 
	&\leq \left( L + S_L(\theta) \right) \norm{\cT_{L,\lambda}(\theta) - \theta}_2 \leq 2L \norm{\cT_{L,\lambda}(\theta) - \theta}_2,
	\end{align*}
	where $S_L(\theta)  =\frac{\norm{\nabla \cL(\cT_{L,\lambda}(\theta)) - \nabla \cL(\theta)}_2}{\norm{\cT_{L,\lambda}(\theta) - \theta}_2}$ is a local Lipschitz constant, which satisfies $S_L(\theta) \leq L_{}$. 
\end{lemma}

\section{Proof of Theorem~\ref{thm:proxgd}}\label{pf:thm:proxgd}
We demonstrate the linear rate when the initial value $\theta^{(0)}$ satisfies $\omega_{\lambda}(\theta^{(0)}) \leq \frac{\lambda}{2}$ with $\norm{(\theta^{(0)})_{\overline{\cS}^*}}_0 \leq \tilde{s}$. The proof is provided in Appendix~\ref{pf:thm:proxrate}. 
\begin{theorem}\label{thm:proxrate}
	Suppose conditions in Lemma~\ref{lem:rsc} hold and $s^*\left({8 \lambda}/{\rho^{-}_{s^*+\tilde{s}}}\right)^2 < r$. Let $\overline{\theta} = \argmin_{\theta} \cF_{\lambda} (\theta)$ be the optimal solution with regularization parameter $\lambda$. If the initial value $\theta^{(0)}$ satisfies $\omega_{\lambda}(\theta^{(0)}) \leq \frac{\lambda}{2}$ with $\norm{(\theta^{(0)})_{\overline{\cS}^*}}_0 \leq \tilde{s}$, then for any $t = 1,2,\ldots$, we have $\norm{(\theta^{(t)})_{\overline{\cS}^*}}_0 \leq \tilde{s}$,
	\begin{align}
	&\norm{\theta^{(t)} - \overline{\theta}}_2^2 \leq \left( 1 - \frac{1}{8{\kappa}_{s^*+2\tilde{s}}} \right)^t \frac{24 \lambda {s^*} \omega_{\lambda}(\theta^{(t)}) }{(\rho^{-}_{s^*+2\tilde{s}})^2} ~~\text{and}~~ \\ &\cF_{\lambda} (\theta^{(t)}) - \cF_{\lambda} (\overline{\theta}) \leq \left( 1 - \frac{1}{8{\kappa}_{s^*+2\tilde{s}}} \right)^t \frac{24 \lambda {s^*} \omega_{\lambda}(\theta^{(t)}) }{\rho^{-}_{s^*+2\tilde{s}}}, \label{eqn:objerr}
	\end{align}
	In addition, to achieve the approximate KKT condition $\omega_{\lambda}(\theta^{(t)}) \leq \varepsilon$, the number of proximal gradient steps is no more than
	\begin{align}
	\frac{\log \left( {96 \left( 1+{\kappa}_{s^*+2\tilde{s}} \right)^2 \lambda^2 {s^*} {\kappa}_{s^*+2\tilde{s}} }/{\varepsilon^2} \right)}{\log \left( {8{\kappa}_{s^*+2\tilde{s}}}/({8{\kappa}_{s^*+2\tilde{s}}}-1) \right)}. \label{eqn:itebd2}
	\end{align}
\end{theorem}
From basic inequalities, since ${\kappa}_{s^*+2\tilde{s}} \geq 1$, we have $\log \left( \frac{8{\kappa}_{s^*+2\tilde{s}}}{{8{\kappa}_{s^*+2\tilde{s}}}-1} \right) \geq \log \left( 1+ \frac{1}{8{\kappa}_{s^*+2\tilde{s}}-1} \right) \geq \frac{1}{8{\kappa}_{s^*+2\tilde{s}}}$. Then \eqref{eqn:itebd2} can be simplified as $\cO \left( {\kappa}_{s^*+2\tilde{s}} \left( \log \left( {{\kappa}_{s^*+2\tilde{s}}^3 \lambda^2 {s^*} }/{\varepsilon^2} \right) \right) \right)$.

As can be seen from Theorem~\ref{thm:proxrate}, when the initial value $\theta^{(0)}$ satisfies $\omega_{\lambda}(\theta^{(0)}) \leq \frac{\lambda}{2}$ with $\norm{(\theta^{(0)})_{\overline{\cS}^*}}_0 \leq \tilde{s}$, then we can guarantee the geometric convergence rate of the estimated objective value towards the minimal objective. 

Next, we need to show that when ${\theta}_{(0)} \in \cB_{r}$, the approximate KKT holds for ${\theta}_{(1)}$, which is also sparse. We demonstrate this result in Lemma~\ref{lem:sublin_to_lin} and provide its proof in Appendix~\ref{pf:lem:sublin_to_lin}.
\begin{lemma}\label{lem:sublin_to_lin}
	Suppose conditions in Lemma~\ref{lem:rsc} hold and $s^*\left({8 \lambda}/{\rho^{-}_{s^*+\tilde{s}}}\right)^2 < r$.s. If $\frac{\rho^{-}_{s^*+\tilde{s}}}{8}\sqrt{\frac{r}{s^*}} > \lambda$ and $\norm{\theta - \theta^*}_2^2 \leq r$ holds, then we have $\omega_{\lambda} (\theta) \leq 4\sqrt{r}$ and $\norm{\theta_{\overline{\cS}^*}}_0 \leq \tilde{s}$.
\end{lemma}

Combining the results above, we finish the proof.

\section{Proof of Theorem~\ref{thm:proxnewton}} \label{pf:thm:proxnewton}

We present a few important intermediate results that are key components of our main proof. The first result shows that in a neighborhood of the true model parameter $\theta^*$, the sparsity of the solution is preserved when we use a sparse initialization. The proof is provided in Appendix~\ref{pf:lem:sparse_preserve}.
\begin{lemma}[Sparsity Preserving Lemma]\label{lem:sparse_preserve}
	Suppose conditions in Lemma~\ref{lem:rsc} hold and $s^*\left({8 \lambda}/{\rho^{-}_{s^*+\tilde{s}}}\right)^2 < r$ with $\varepsilon \leq \frac{\lambda}{8}$. Given $\theta^{(t)} \in \cB \left({\theta^*}, R\right)$ and $\norm{\theta^{(t)}_{\Sc}}_0\leq \tilde{s}$, there exists a generic constant $C_1$ such that
	\begin{align*}
	&\norm{\theta^{(t+1)}_{\Sc}}_0\leq\tilde{s},~~\norm{\theta^{(t+1)} - \theta^{*}}_2 \leq \frac{C_1 \lambda \sqrt{s^*}}{\rho^-_{s^*+2\tilde{s}}}\quad\text{and}\quad\cF_{\lambda}({\theta}^{(t)})\leq\cF_{\lambda}(\theta^*) + \frac{15 \lambda^2 s^*}{4\rho^{-}_{s^*+2\tilde{s}}}..
	\end{align*}
\end{lemma}

Denote ${\cB(\theta, r){=}\cbr{\bphi \in \RR^d ~|~ \norm{\bphi-\theta}_2 \leq r}}$. We then show that every step of proximal Newton updates within each stage has a quadratic convergence rate to a local minimizer, if we start with a sparse solution in the refined region. The proof is provided in Appendix~\ref{pf:lem:quadratic_rate}.

\begin{lemma} \label{lem:quadratic_rate}
	Suppose conditions in Lemma~\ref{lem:rsc} hold and $s^*\left({8 \lambda}/{\rho^{-}_{s^*+\tilde{s}}}\right)^2 < r$. If $\theta^{(t)}\in \cB\rbr{{\theta^*}, r}$ and $\nbr{{\theta^{(t)}}_{\Sc}}_0\le \tilde{s}$, then for each stage $K \geq 2$, we have
	\begin{align*}
	\norm{\theta^{(t+1)} - \overline{\theta}}_2 \le \frac{L_{s^*+2\tilde{s}}}{2\rho^-_{s^*+2\tilde{s}}} \norm{\theta^{(t)} -  \overline{\theta}}_2^2.
	\end{align*}
	
\end{lemma}

In the following, we need to use the property that the iterates $\theta^{(t)} \in \cB(\overline{\theta}, 2r)$ instead of $\theta^{(t)}\in \cB\rbr{{\theta^*}, r}$ for convergence analysis of the proximal Newton method. This property holds since we have $\theta^{(t)}\in \cB\rbr{{\theta^*}, r}$ and $\overline{\theta} \in \cB\rbr{{\theta^*}, r}$ simultaneously. Thus $\theta^{(t)}\in \cB\rbr{\overline{\theta}, 2r}$, where $2r = \frac{\rho^-_{s^*+2\tilde{s}}}{L_{s^*+2\tilde{s}}}$ is the radius for quadratic convergence region of the proximal Newton algorithm.

The following lemma demonstrates that the step size parameter is simply 1 if the the sparse solution is in the refined region. The proof is provided in Appendix~\ref{pf:lem:unit_step}.

\begin{lemma} \label{lem:unit_step}
	Suppose conditions in Lemma~\ref{lem:rsc} hold and $s^*\left({8 \lambda}/{\rho^{-}_{s^*+\tilde{s}}}\right)^2 < r$. If $\theta^{(t)}\in \cB(\overline{\theta}, 2r)$ and $\norm{{\theta^{(t)}}_{\Sc}}_0\le \tilde{s}$ at each stage $K \geq 2$ with $\frac{1}{4}\le \alpha< \frac{1}{2}$, then $\eta_{t}=1$. Further, we have
	\begin{align*}
	\cF_{\lambda}({\theta^{(t+1)}})\leq\cF_{\lambda}(\theta^{(t)}) + \frac{1}{4}\gamma_t.
	\end{align*}
\end{lemma}

Moreover, we present a critical property of $\gamma_t$. The proof is provided in Appendix~\ref{pf:lem:prop_lamb}.

\begin{lemma}\label{lem:prop_lamb}
	Denote $\Delta \theta^{(t)} = \theta^{(t)} - \theta^{(t+1)}$ and
	\begin{align*}
	\gamma_t = \nabla\cL(\theta^{(t)})^{\top} \Delta \theta^{(t)}+ \nbr{ \lambda (\theta^{(t)} + \Delta \theta^{(t)}) }_1 - \nbr{ \lambda (\theta^{(t)}) }_1.
	\end{align*}
	Then we have $\gamma_t \le -\norm{\Delta \theta^{(t)}}^2_{\nabla^2 \cL(\theta^{(t)})}$.
\end{lemma}

In addition, we present the sufficient number of iterations for each convex relaxation stage to achieve the approximate KKT condition. The proof is provided in Appendix~\ref{pf:lem:approx_kkt}.
\begin{lemma}\label{lem:approx_kkt}
	Suppose conditions in Lemma~\ref{lem:rsc} hold and $s^*\left({8 \lambda}/{\rho^{-}_{s^*+\tilde{s}}}\right)^2 < r$. To achieve the approximate KKT condition $\omega_{\lambda} \rbr{\theta^{(t)}} \le \varepsilon$ for any $\varepsilon >0$ at each stage $K \geq 2$, the number of iteration for proximal Newton updates is at most
	\begin{align*}
	\log {\log\rbr{\frac{3\rho^+_{s^*+2\tilde{s}}}{\varepsilon}}}.
	\end{align*}
\end{lemma}

Combining the results above, we have desired results in Theorem~\ref{thm:proxnewton}.

\section{Proof of Theorem~\ref{thm:statrate}}\label{pf:thm:statrate}

\textbf{Part 1}. We first show that estimation errors are as claimed. We have that $\omega_{\lambda}(\hat{\theta}^{(0)}) \leq \lambda/2$. By Theorem~\ref{thm:proxrate}, we have for any $t=1,2,\ldots$, $\norm{(\theta_{[K+1]}^{(t)})_{\overline{\cS}^*}}_0 \leq \tilde{s}$. Applying Lemma~\ref{lem:step1} recursively, we have
\begin{align*}
\norm{ \hat{\theta} - \theta^*}_2 \leq \frac{2 \lambda \sqrt{s^*}}{\rho^{-}_{s^*+2\tilde{s}}}~~~\text{and}~~~\norm{ \hat{\theta} - \theta^*}_1 \leq \frac{12 \lambda s^*}{\rho^{-}_{s^*+2\tilde{s}}}.
\end{align*}
Applying Lemma~\ref{lem:rsc} with $\lambda \leq { 24\sqrt{{\log d}/{n}}}$ and $\rho^{-}_{s^*+2\tilde{s}} = \frac{\psi_{\min}}{8\sigma}$, then by union bound, with probability at least $1-8\exp\left( -\frac{n}{144} \right) - 2d^{-1}$, we have
\begin{align*}
&\norm{ \hat{\theta} - \theta^*}_2 \leq \frac{384 \sigma \sqrt{s^* \log d/n}}{\psi_{\min}} ~~\text{and}~~ \norm{ \hat{\theta} - \theta^*}_1 \leq \frac{2304 \sigma s^* \sqrt{\log d/n}}{\psi_{\min}}. 
\end{align*}

\textbf{Part 2}. Next, we demonstrate the result of the estimation of variance. Let $\overline{\theta} = \argmin_{\theta} \cF_{\lambda} (\theta)$ be the optimal solution. Apply the argument in Part 1 recursively, we have
\begin{align}
\norm{ \overline{\theta} - \theta^*}_1 &\leq \frac{2304 \sigma s^* \sqrt{\log d/n}}{\psi_{\min}}. \label{eqn:l1errbd}
\end{align} 
Denote $c_1,c_2,\ldots$ as positive universal constants. Then we have
\begin{align}
\cL(\overline{\theta}) - \cL(\theta^*) &\leq \lambda(\norm{\theta^*}_1 - \norm{\overline{\theta}}_1) \leq \lambda(\norm{\theta^*_{\cS^*}}_1 - \norm{(\overline{\theta})_{\cS^*}}_1 - \norm{(\overline{\theta})_{\overline{\cS}^*}}_1) \nonumber\\
&\leq \lambda \norm{(\overline{\theta} - \theta^*)_{\cS^*}}_1 \leq \lambda \norm{\overline{\theta} - \theta^*}_1 \overset{(i)}{\leq} c_1 \frac{\sigma s^* \log d}{n}, \label{eqn:var_est_ub}
\end{align}
where (i) is from the value of $\lambda$ and $\ell_1$ error bound in \eqref{eqn:l1errbd}. 

On the other hand, from the convexity of $\cL(\theta)$, we have
\begin{align}
\cL(\overline{\theta}) - \cL(\theta^*) &\geq (\overline{\theta} - \theta^*)^\top \nabla \cL(\theta^*) \geq - \norm{\nabla \cL(\theta^*)}_{\infty} \norm{\hat{\theta} - \theta}_1 \nonumber\\
& \overset{(i)}{\geq} - c_2 \lambda \norm{\overline{\theta} - \theta}_1 \overset{(ii)}{\geq} -c_3 \frac{\sigma s^* \log d}{n}, \label{eqn:var_est_lb}
\end{align}
where (i) is from Lemma~\ref{lem:rsc} and (ii) value of $\lambda$ and $\ell_1$ error bound in \eqref{eqn:l1errbd}. By definition, we have
\begin{align}
\cL(\overline{\theta}) - \cL(\theta^*) = \frac{\norm{ y -  X \overline{\theta}}_2}{\sqrt{n}} - \frac{\norm{\epsilon}_2}{\sqrt{n}}.\label{eqn:Lmu}
\end{align}
From \cite{wainwright2015high}, we have for any $\delta>0$,
\begin{align}
\PP \left[ \left| \frac{\norm{\epsilon}_2^2}{n} - \sigma^2 \right| \geq \sigma^2\delta \right] \leq 2\exp \left(-\frac{n \delta^2}{18} \right). \label{eqn:tailbd}
\end{align}
Combining \eqref{eqn:var_est_ub}, \eqref{eqn:var_est_lb}, \eqref{eqn:Lmu} and \eqref{eqn:tailbd} with $\delta^2 = \frac{c_3 s^* \log d}{n}$, we have with high probability,
\begin{align}
\left| \frac{\norm{ y -  X \overline{\theta}}_2}{\sqrt{n}} - \sigma \right| = \cO \left(\frac{\sigma s^* \log d}{n}\right). \label{eqn:sigma_bar}
\end{align}

From Part 1, for $n \geq c_4 {s^* \log d}$, with high probability, we have  $\norm{\overline{\theta} - \theta^*}_2 \leq \frac{384\sigma \sqrt{s^* \log d/n}}{\psi_{\min}} \leq \frac{\sigma}{2\sqrt{2\psi_{\max}}}$, then $\overline{\theta} \in \cB_{r}^{s^*+\tilde{s}}$ and $\norm{\hat{\theta}  - \overline{\theta}}_0 \leq s^* + 2\tilde{s}$. Then from the analysis of Theorem~\ref{thm:proxrate}, we have
\begin{align*}
\omega_{\lambda} (\theta^{(t+1)} ) \leq \left( 1+{\kappa}_{s^*+2\tilde{s}} \right) \sqrt{4 \rho^{+}_{s^*+2\tilde{s}} \left(\cF_{\lambda}(\theta^{(t)}) - \cF_{\lambda}(\overline{\theta}) \right)} \leq \varepsilon. 
\end{align*}
This implies
\begin{align}
\cF_{\lambda}(\theta^{(t)}) - \cF_{\lambda}(\overline{\theta}) \leq \frac{\epsilon^2}{4 \rho^{+}_{s^*+2\tilde{s}} \left( 1+{\kappa}_{s^*+2\tilde{s}} \right)^2}. \label{eqn:upbdF}
\end{align}
On the other hand, from the LRSC property of $\cL$, convexity of $\ell_1$ norm and optimality of $\overline{\theta}$, we have
\begin{align}
\cF_{\lambda}(\theta^{(t)}) - \cF_{\lambda}(\overline{\theta}) \geq \rho^{-}_{s^*+2\tilde{s}} \norm{\hat{\theta}  - \overline{\theta}}_2^2. \label{eqn:lwbdF}
\end{align}
Combining \eqref{eqn:upbdF}, \eqref{eqn:lwbdF} and Lemma~\ref{lem:rsc}, we have
\begin{align}
\frac{\norm{ X (\hat{\theta} - \overline{\theta})}_2}{\sqrt{n}} \leq \sqrt{\frac{8 \rho^{+}_{s^*+2\tilde{s}}}{\sigma}} \norm{\hat{\theta} - \theta^*}_2 \leq \sqrt{\frac{2}{\sigma \rho^{-}_{s^*+2\tilde{s}}}}\frac{\epsilon}{\left( 1+{\kappa}_{s^*+2\tilde{s}} \right)} \leq \frac{4\epsilon}{\left( 1+{\kappa}_{s^*+2\tilde{s}} \right) \sqrt{\psi_{\min}}}. \label{eqn:bd3}
\end{align}
Combining \eqref{eqn:sigma_bar} and \eqref{eqn:bd3}, we have
\begin{align*}
\left| \frac{\norm{ y -  X \hat{\theta}}_2}{\sqrt{n}} - \frac{\norm{ y -  X \overline{\theta}}_2}{\sqrt{n}} \right| &\leq \frac{\norm{ X (\hat{\theta} - \overline{\theta})}_2}{\sqrt{n}} \leq \frac{4\epsilon}{\left( 1+{\kappa}_{s^*+2\tilde{s}} \right) \sqrt{\psi_{\min}}}.
\end{align*}
If $\epsilon \leq c_5 \frac{\sigma s^* \log d}{n}$ for some constant $c_5$, then we have the desired result. 

\section{Intermediate Results of Theoremm~\ref{thm:rsc2}}\label{int:thm:intrsc2}

We first characterize the sparsity of $\hat{\theta}$ and its distance to $\theta^*$ when approximate KKT condition holds in Lemma~\ref{lem:sublin_to_linv2} and provide its proof in Appendix~\ref{pf:lem:sublin_to_linv2}.
\begin{lemma}\label{lem:sublin_to_linv2}
	Suppose conditions in Lemma~\ref{lem:rsc} hold and $s^*\left({8 \lambda}/{\rho^{-}_{s^*+\tilde{s}}}\right)^2 < r$, and the approximate KKT satisfies $\omega_{\lambda} (\theta) \leq \lambda/4$. If  $\frac{\rho^{-}_{s^*+\tilde{s}}}{8}\sqrt{\frac{r}{s^*}} > \lambda > \lambda_{[N]}$, then we have
	\begin{align*}
	\norm{\theta - \theta^*}_2^2 \leq r~~~\text{and}~~~\norm{\theta_{\overline{\cS}^*}}_0 \leq \tilde{s}.
	\end{align*}
\end{lemma}
Next, we show that if the optimal solution $\hat{\theta}_{[K-1]}$ from $K-1$-th path following stage satisfies the approximate KKT condition and the regularization parameter $\lambda_{[K]}$ in the $K$-th path following stage is chosen properly, then $\hat{\theta}_{[K-1]}$ satisfies the approximate KKT condition for $\lambda_{[K]}$ with a slightly larger bound. This characterizes that good computational properties are preserved by using the warm start $\theta_{[K]}^{(0)} = \hat{\theta}_{[K-1]}$ and geometric sequence of regularization parameters $\lambda_{[K]}$. We formalize this notion in Lemma~\ref{lem:warmst}, and its proof is provided in Appendix~\ref{pf:lem:warmst}.
\begin{lemma}\label{lem:warmst}
	Let $\hat{\theta}_{[K-1]}$ be the approximate solution of $K-1$-th path following state, which satisfies the approximate KKT condition $\omega_{\lambda_{[K-1]}}(\hat{\theta}_{[K-1]}) \leq \lambda_{[K-1]}/4$. Then we have
	\begin{align*}
	\omega_{\lambda_{[K]}}(\hat{\theta}_{[K-1]}) \leq \lambda_{[K]}/2,
	\end{align*}
	where $\lambda_{[K]} = \eta_{\lambda} \lambda_{[K-1]}$ with $\eta_{\lambda} \in (5/6,1)$. 
\end{lemma}

\section{Proof of Theorem~\ref{thm:rsc2}}\label{pf:thm:rsc2}

\textbf{Part 1}. We first show the existence of $N_1$. Following the notation in Appendix~\ref{pf:lem:rsc}, $r = \frac{\sigma^2}{8\psi_{\max}}$ is a constant independent of $n$. As a result for a large enough $n>C_2 s^* \log d$,  we have 	
\begin{align*}
r = \frac{\sigma^2}{8\psi_{\max}} \overset{(i)}{>} s^*\left({64 \sigma \lambda_{[N_1]}}/{\psi_{\min}}\right)^2 \overset{(ii)}{\geq} s^*\left({8 \lambda_{[N_1]}}/{\rho^{-}_{s^*+\tilde{s}}}\right)^2,
\end{align*}
where $(i)$ is from $n>C_2 s^* \log d$ with a sufficiently large constant $C_2$ and $\lambda_{[N_1]} = \frac{1}{\eta_\lambda}^{N-N_1} \lambda_{[N]} = \frac{1}{\eta_\lambda}^{N-N_1} C_1\sqrt{\frac{\log d}{n}}$, and $(ii)$ is from $\rho^{-}_{s^*+2\tilde{s}} \geq \frac{\psi_{\min}}{8\sigma}$. 

\textbf{Part 2}. We next show that for $K \in [N_1,...,N-1]$, $\lambda_{[K]}$, $\hat{\theta}_{[K]}$ is a good initial for $\theta^{(0)}_{[K+1]}$, i.e., satisfies $$\norm{\hat{\theta}_{[K]}-\theta^*}^2_2 \leq s^*\left({8 \lambda_{[K+1]}}/{\rho^{-}_{s^*+\tilde{s}}}\right)^2~~\textrm{and}~~\norm{[\hat{\theta}_{[K]}]_{\overline{\cS}^*}}_{0}\leq \tilde{s}.$$ 

Lemma~\ref{lem:sublin_to_linv2} directly implies  $\norm{[\hat{\theta}_{[K]}]_{\overline{\cS}^*}}_{0}\leq \tilde{s}$. Applying Lemma~\ref{lem:warmst}, we have $\omega_{\lambda_{[K+1]}}(\hat{\theta}_{[K]}) \leq \lambda_{[K+1]}/2$. Then we can apply Lemma~\ref{lem:step1}, we have $\norm{\hat{\theta}_{[K]} - \theta^*}_2^2 \leq ( 2 \lambda_{[K+1]} \sqrt{s^*} /\rho^{-}_{s^*+\tilde{s}})^2 \leq s^*\left({8 \lambda_{[K+1]}}/{\rho^{-}_{s^*+\tilde{s}}}\right)^2$

\textbf{Part 3}.  So far we prove that for $K \in [N_1+1,...,N-1]$, $$\norm{\theta^{(0)}_{[K]}-\theta^*}^2_2 \leq s^*\left({8 \lambda_{[K]}}/{\rho^{-}_{s^*+\tilde{s}}}\right)^2 \leq s^*\left({8 \lambda_{[N_1]}}/{\rho^{-}_{s^*+\tilde{s}}}\right)^2 <r~~\textrm{and}~~\norm{[\theta^{(0)}_{[K]}]_{\overline{\cS}^*}}_{0} \leq \tilde{s}.$$ So the fast convergence rate in Theorems~\ref{thm:proxgd} and \ref{thm:proxnewton} hold for $\lambda_K$.

\section{Proof of Theorem~\ref{thm:proxrate}}\label{pf:thm:proxrate}

Note that the RSS property implies that line search terminate when $\tilde{L}^{(t)}$ satisfies
\begin{align}
\rho^{+}_{s^*+2\tilde{s}} \leq \tilde{L}^{(t)} \leq 2\rho^{+}_{s^*+2\tilde{s}}. \label{eqn:line_rss}
\end{align}

Since the initialization $\theta^{(0)}$ satisfies $\omega_{\lambda}(\theta^{(0)}) \leq \frac{\lambda}{2}$ with $\norm{(\theta^{(0)})_{\overline{\cS}^*}}_0 \leq \tilde{s}$, then by Lemma~\ref{lem:step1}, we have $\cF_{\lambda} (\theta^{(0)}) - \cF_{\lambda} (\theta^*) \leq \frac{6 \lambda^2 s^*}{\rho^{-}_{s^*+2\tilde{s}}}$. Then by Lemma~\ref{lem:step3}, we have $\norm{(\theta^{(1)})_{\overline{\cS}^*}}_0 \leq \tilde{s}$.

By monotone decrease of $\cF_{\lambda}(\theta^{(t)})$ from \eqref{eqn:proxobjbd1} in Lemma~\ref{lem:suf_des_obj} and recursively applying Lemma~\ref{lem:step3}, $\norm{(\theta^{(t)})_{\overline{\cS}^*}}_0 \leq \tilde{s}$ holds in \eqref{eqn:objerr} for all $t=1,2,\ldots$.

For the objective error, we have
\begin{align}
&\cF_{\lambda}(\theta^{(t)}) - \cF_{\lambda}(\overline{\theta})  \overset{(i)}{\leq} \left( 1 - \frac{1}{8{\kappa}_{s^*+2\tilde{s}}} \right)^t \left(\cF_{\lambda}(\theta^{(0)}) - \cF_{\lambda}(\overline{\theta}) \right) \overset{(ii)}{\leq} \left( 1 - \frac{1}{8{\kappa}_{s^*+2\tilde{s}}} \right)^t \frac{24 \lambda {s^*} \omega_{\lambda}(\theta^{(t)} }{\rho^{-}_{s^*+2\tilde{s}}}, \label{eqn:objerr2}
\end{align}
where (i) is from Lemma~\ref{lem:controlsp}, and (ii) is from Lemma~\ref{lem:objbd} and $\omega_{\lambda}(\theta^{(t+1)} ) \leq \lambda/2 \leq \lambda$, which results in \eqref{eqn:objerr}. 

Combining \eqref{eqn:objerr2},  \eqref{eqn:rsc9} with $\nabla \cF_{\lambda}(\overline{\theta})= 0$, we have
\begin{align*}
\norm{\theta^{(t)} - \overline{\theta}}_2^2 &\leq \frac{1}{\rho^{-}_{s^*+2\tilde{s}}} \left( \cF_{\lambda}(\theta^{(t)}) - \cF_{\lambda}(\overline{\theta}) - \nabla \cF_{\lambda}(\overline{\theta}) \right) \leq \left( 1 - \frac{1}{8{\kappa}_{s^*+2\tilde{s}}} \right)^t \frac{24 \lambda {s^*} \omega_{\lambda}(\theta^{(t)}) }{(\rho^{-}_{s^*+2\tilde{s}})^2}
\end{align*}

For $\omega_{\lambda} (\theta^{(t+1)} )$ of $(t+1)$-th iteration, we have
\begin{align}
&\omega_{\lambda} (\theta^{(t+1)} ) \nonumber\\&
\overset{(i)}{\leq} \left( \tilde{L}^{(t)}+S_{\tilde{L}^{(t)}}(\theta^{(t)}) \right) \norm{\theta^{(t+1)} - \theta^{(t)}}_2 \overset{(ii)}{\leq} \left( \tilde{L}^{(t)}+\rho^{+}_{s^*+2\tilde{s}} \right) \norm{\theta^{(t+1)} - \theta^{(t)}}_2 \nonumber \\
&\overset{(iii)}{\leq} \tilde{L}^{(t)} \left( 1+\frac{\rho^{+}_{s^*+2\tilde{s}}}{\rho^{-}_{s^*+2\tilde{s}}} \right) \norm{\theta^{(t+1)} - \theta^{(t)}}_2 \overset{(iv)}{\leq} \tilde{L}^{(t)} \left( 1+\frac{\rho^{+}_{s^*+2\tilde{s}}}{\rho^{-}_{s^*+2\tilde{s}}} \right) \sqrt{\frac{2 \left(\cF_{\lambda}(\theta^{(t)}) - \cF_{\lambda}(\theta^{(t+1)}) \right)}{\tilde{L}^{(t)}}} \nonumber\\
&\overset{(v)}{\leq} \left( 1+{\kappa}_{s^*+2\tilde{s}} \right) \sqrt{4 \rho^{+}_{s^*+2\tilde{s}} \left(\cF_{\lambda}(\theta^{(t)}) - \cF_{\lambda}(\overline{\theta}) \right)} \overset{(vi)}{\leq} \left( 1+{\kappa}_{s^*+2\tilde{s}} \right) \sqrt{96 \lambda^2 {s^*} {\kappa}_{s^*+2\tilde{s}} \left( 1 - \frac{1}{8{\kappa}_{s^*+2\tilde{s}}} \right)^t }, \label{eqn:omegabd}
\end{align}
where (i) is from Lemma~\ref{lem:upbd_omg}, (ii) is from $S_{\tilde{L}^{(t)}}(\theta^{(t)}) \leq \rho^{+}_{s^*+2\tilde{s}}$, (iii) is from $\rho^{-}_{s^*+2\tilde{s}} \leq \tilde{L}^{(t)}$ in \eqref{eqn:line_rss}, (iv) is from \eqref{eqn:proxobjbd1} in Lemma~\ref{lem:suf_des_obj}, (v) is from $\tilde{L}^{(t)} \leq 2 \rho^{+}_{s^*+2\tilde{s}}$ in \eqref{eqn:line_rss} and monotone decrease of $\cF_{\lambda}(\theta^{(t)})$ from \eqref{eqn:proxobjbd1} in Lemma~\ref{lem:suf_des_obj}, and (vi) is from \eqref{eqn:objerr2} and ${\kappa}_{s^*+2\tilde{s}} = \frac{\rho^{+}_{s^*+2\tilde{s}}}{\rho^{-}_{s^*+2\tilde{s}}}$.

Then we need $\omega_{\lambda} (\hat{\theta} ) \leq \varepsilon \leq \lambda/4$. Set the R.H.S. of \eqref{eqn:omegabd} to be no greater than $\varepsilon$, which is equivalent to require the number of iterations $k$ to be an upper bound of \eqref{eqn:itebd2}.

\section{Proof of Lemma~\ref{lem:sublin_to_lin}}\label{pf:lem:sublin_to_lin}

\textbf{Part 1}. We first show that given $\norm{\theta^{(0)} - \theta^*}_2^2 \leq r$, $\omega_{\lambda} (\theta^{(1)}) \leq 4\sqrt{r}$ holds. From Lemma~\ref{lem:upbd_omg}, we have
\begin{align*}
\omega_{\lambda} (\theta^{(1)}) \leq  2L \norm{\theta^{(1)} - \theta^{(0)}}_2 \leq 4 \norm{\theta^{(1)} - \theta^*}_2 \leq 4\sqrt{r}.
\end{align*}

\textbf{Part 2}. We next demonstrate the sparsity of $\theta$. From $\lambda \geq  6 \norm{\nabla \cL(\theta^*)}_{\infty}$, then we have
\begin{align}
\left| \left\{ i \in \overline{\cS}^* : |\nabla_i \cL(\theta^*) | \geq \frac{\lambda}{6} \right\} \right| = 0. \label{eqn:card_set1}
\end{align}
Denote $\check{\cS}_1 = \left\{ i \in \overline{\cS}^* : | \nabla_i \cL(\theta) - \nabla_i \cL(\theta^*) | \geq \frac{2\lambda}{3} \right\}$ and $\check{s}_1 = |\check{\cS}_1|$. Then there exists some $\bb \in \RR^d$ such that $\norm{\bb}_{\infty} = 1$, $\norm{\bb}_0 \leq \check{s}_1$ and $\bb^\top (\nabla \cL(\theta) - \nabla \cL(\theta^*)) \geq \frac{2 \lambda \check{s}_1}{3}$. Then by the mean value theorem, we have for some $\check{\theta} = (1-\alpha) \theta + \alpha \theta^*$ with $\alpha \in [0,1]$, $\nabla \cL(\theta) - \nabla \cL(\theta^*) = \nabla^2 \cL (\check{\theta}) \Delta$, where $\Delta = \theta - \theta^*$. Then we have
\begin{align}
\frac{2\lambda \check{s}_1}{3} \leq \bb^\top \nabla^2 \cL (\check{\theta}) \Delta \overset{(i)}{\leq} \sqrt{\bb^\top \nabla^2 \cL (\check{\theta}) \bb} \sqrt{\Delta^\top \nabla^2 \cL (\check{\theta}) \Delta} \overset{(ii)}{\leq} \sqrt{\check{s}_1 \rho^{+}_{\check{s}_1} } \sqrt{\Delta^\top (\nabla \cL(\theta) - \nabla \cL(\theta^*))}, \label{eqn:sp_bd1}
\end{align}
where (i) is from the generalized Cauchy-Schwarz inequality, (ii) is from the definition of RSS and the fact that $\norm{\bb}_2 \leq \sqrt{\check{s}_1} \norm{\bb}_{\infty} = \sqrt{\check{s}_1}$. Let $ g$ achieve $\min_{ g \in \partial \norm{\theta}_1} \cF_{\lambda}(\theta)$. Further, we have
\begin{align}
&\Delta^\top (\nabla \cL(\theta) - \nabla \cL(\theta^*)) \leq \norm{\Delta}_1 \norm{\nabla \cL(\theta) - \nabla \cL(\theta^*)}_{\infty} \leq \norm{\Delta}_1 (\norm{\nabla \cL(\theta^*)}_{\infty} + \norm{\nabla \cL(\theta)}_{\infty}) \nonumber\\
&\leq \norm{\Delta}_1 (\norm{\nabla \cL(\theta^*)}_{\infty} + \norm{\nabla \cL(\theta) + \lambda  g}_{\infty} + \lambda\norm{ g}_{\infty}) \overset{(i)}{\leq} \frac{28 \lambda s^*}{3 \rho^{-}_{s^*+\tilde{s}}} (\frac{\lambda}{6}+\frac{\lambda}{4} + \lambda) \leq \frac{14 \lambda^2 s^*}{\rho^{-}_{s^*+\tilde{s}}}, \label{eqn:sp_bd2}
\end{align}
where (i) is from $\norm{\tilde{\Delta}_{\overline{\cS}^*}}_1 \leq \frac{5}{2} \norm{\tilde{\Delta}_{\cS^*}}_1$ and $\norm{\tilde{\Delta}_{\cS^*}}_1 \leq \frac{8\lambda s^*}{3 \rho^{-}_{s^*+\tilde{s}}}$, condition on $\lambda$, approximate KKT condition and $\norm{ g}_{\infty} \leq 1$. Combining \eqref{eqn:sp_bd1} and \eqref{eqn:sp_bd2}, we have $\frac{2\sqrt{\check{s}_1}}{3} \leq \sqrt{\frac{14 \rho^{+}_{\check{s}_1} s^*}{\rho^{-}_{s^*+\tilde{s}}}}$, which further implies
\begin{align}
\check{s}_1 \leq \frac{32 \rho^{+}_{\check{s}_1} s^*}{\rho^{-}_{s^*+\tilde{s}}} \leq 32 {\kappa}_{s^*+2\tilde{s}} s^* \leq \tilde{s}. 
\end{align}
For any $ v \in \RR^d$ that satisfies $\norm{ v}_0 \leq 1$, we have
\begin{align*}
\check{\cS}_2 = \left\{ i \in \overline{\cS}^* : \left| \nabla_i \cL (\theta) + \frac{\lambda}{4} v_i \right| \geq \frac{5 \lambda}{6} \right\} 
\subseteq \left\{ i \in \overline{\cS}^* : |\nabla_i \cL(\theta^*) | \geq \frac{\lambda}{6} \right\} \bigcup \check{\cS}_1.
\end{align*}
Then we have $|\check{\cS}_2| \leq |\check{\cS}_1| \leq \tilde{s}$. Since for any $i \in \overline{\cS}^*$ and $\left| \nabla_i \cL (\theta) + \frac{\lambda}{4} v_i \right| < \frac{5 \lambda}{6}$, we can find $g_i$ that satisfies $|g_i| \leq 1$ such that $\nabla_i \cL (\theta) + \frac{\lambda}{4} v_i + \lambda g_i = 0$ which implies $\theta_i = 0$, then we have
\begin{align*}
\left| \left\{ i \in \overline{\cS}^* : \left| \nabla_i \cL (\theta) + \frac{\lambda}{4} v_i \right| < \frac{5 \lambda}{6} \right\} \right| = 0.
\end{align*}
This implies $\norm{\theta_{\overline{\cS}^*}}_0 \leq |\check{\cS}_2| \leq \tilde{s}$. 

\section{Proofs of Intermediate Lemmas in Appendix~\ref{int:thm:linearmodel}}\label{pf:int:thm:linearmodel}

\subsection{Proof of Lemma~\ref{lem:step1}}\label{pf:lem:step1}

We first bound the estimation error. From Lemma~\ref{lem:rsc}, we have the RSC property, which indicates
\begin{align}
\cL(\theta) &\geq \cL(\theta^*) + (\theta - \theta^*)^\top \nabla \cL(\theta^*) + \frac{\rho^{-}_{s^*+\tilde{s}}}{2} \norm{\theta - \theta^*}_2^2, \label{eqn:rsc8} \\
\cL(\theta^*) &\geq \cL(\theta) + (\theta^* - \theta)^\top \nabla \cL(\theta) + \frac{\rho^{-}_{s^*+\tilde{s}}}{2} \norm{\theta - \theta^*}_2^2, \label{eqn:rsc6}
\end{align}
Adding \eqref{eqn:rsc6} and \eqref{eqn:rsc8}, we have
\begin{align}
(\theta - \theta^*)^\top \nabla \cL(\theta) \geq (\theta - \theta^*)^\top \nabla \cL(\theta^*) + \rho^{-}_{s^*+\tilde{s}} \norm{\theta - \theta^*}_2^2. \label{eqn:rsc5}
\end{align}

Let $ g \in \partial \norm{\theta}_1$ be the subgradient that achieves the approximate KKT condition, then
\begin{align}
(\theta - \theta^*)^\top \left( \nabla \cL (\theta) + \lambda  g \right) &\leq \norm{\theta - \theta^*}_1 \left\Vert \nabla \cL (\theta) + \lambda  g \right\Vert_{\infty} \leq \frac{1}{2} \lambda \norm{\theta - \theta^*}_1. \label{eqn:upbd}
\end{align}
On the other hand, we have from \eqref{eqn:rsc5}
\begin{align}
&(\theta - \theta^*)^\top \left( \nabla \cL (\theta) + \lambda  g \right) {\geq} (\theta - \theta^*)^\top \nabla \cL (\theta^*) + \rho^{-}_{s^*+\tilde{s}} \norm{\theta - \theta^*}_2^2  + \lambda  g^\top (\theta - \theta^*), \label{eqn:lbd}
\end{align}

Since $\norm{\theta - \theta^*}_1 = \norm{(\theta - \theta^*)_{\cS^*}}_1 + \norm{(\theta - \theta^*)_{\overline{\cS}^*}}_1$, then 
\begin{align}
&(\theta - \theta^*)^\top \nabla \cL (\theta^*) \geq -\norm{(\theta - \theta^*)_{\cS^*}}_1 \norm{\cL (\theta^*)}_{\infty} - \norm{(\theta - \theta^*)_{\overline{\cS}^*}}_1 \norm{\cL (\theta^*)}_{\infty}. \label{eqn:lbd2}
\end{align}
Besides, we have
\begin{align}
(\theta - \theta^*)^\top  g &=  g_{\cS^*}^\top(\theta - \theta^*)_{\cS^*}  +  g_{\overline{\cS}^*}^\top (\theta - \theta^*)_{\overline{\cS}^*} \overset{(i)}{\geq} -\norm{ g_{\cS^*}}_{\infty} \norm{(\theta - \theta^*)_{\cS^*}}_1 +  g_{\overline{\cS}^*}^\top \theta_{\overline{\cS}^*} \nonumber\\
&\overset{(ii)}{\geq} - \norm{(\theta - \theta^*)_{\cS^*}}_1 + \norm{ g_{\overline{\cS}^*}}_1 \overset{(iii)}{=} - \norm{(\theta - \theta^*)_{\cS^*}}_1 + \norm{(\theta - \theta^*)_{\overline{\cS}^*}}_1, \label{eqn:lbd3}
\end{align}
where (i) and (iii) is from $\theta^*_{\overline{\cS}^*} =  0$, (ii) is from $\norm{ g_{\cS^*}}_{\infty} \leq 1$ and $ g \in \partial \norm{\theta}_1$. 

Combining \eqref{eqn:upbd}, \eqref{eqn:lbd}, \eqref{eqn:lbd2} and \eqref{eqn:lbd3}, we have
\begin{align*}
&\frac{1}{2} \lambda \norm{\theta - \theta^*}_1 \geq \rho^{-}_{s^*+\tilde{s}} \norm{\theta - \theta^*}_2^2 - (\lambda + \norm{\cL (\theta^*)}_{\infty}) \norm{(\theta - \theta^*)_{\cS^*}}_1 + (\lambda-\norm{\cL (\theta^*)}_{\infty} ) \norm{(\theta - \theta^*)_{\overline{\cS}^*}}_1.
\end{align*}
This implies
\begin{align}
&\rho^{-}_{s^*+\tilde{s}} \norm{\theta - \theta^*}_2^2 + (\frac{1}{2}\lambda-\norm{\cL (\theta^*)}_{\infty} ) \norm{(\theta - \theta^*)_{\overline{\cS}^*}}_1 \leq (\frac{3}{2}\lambda + \norm{\cL (\theta^*)}_{\infty}) \norm{(\theta - \theta^*)_{\cS^*}}_1, \label{eqn:l2bd1}
\end{align}
which results in \eqref{eqn:l1Scbd} from $\rho^{-}_{s^*+2\tilde{s}}>0$ and Lemma~\ref{lem:rsc} as
\begin{align*}
\norm{(\theta - \theta^*)_{\overline{\cS}^*}}_1 \leq \frac{\frac{3}{2}\lambda + \norm{\cL (\theta^*)}_{\infty}}{\frac{1}{2}\lambda - \norm{\cL (\theta^*)}_{\infty}} \norm{(\theta - \theta^*)_{\cS^*}}_1.
\end{align*}
Combining  $\frac{1}{2}\lambda-\norm{\cL (\theta^*)}_{\infty} \geq 0$, $\frac{3}{2}\lambda + \norm{\cL (\theta^*)}_{\infty} \leq 2\lambda$ and \eqref{eqn:l2bd1}, we have estimation errors in \eqref{eqn:l2bd} and \eqref{eqn:l1bd} as
\begin{align*}
&\rho^{-}_{s^*+2\tilde{s}} \norm{\theta - \theta^*}_2^2 \leq 2\lambda \norm{(\theta - \theta^*)_{\cS^*}}_1 \leq 2\lambda \sqrt{s^*} \norm{\theta - \theta^*}_2~~\text{and}~~\norm{\theta - \theta^*}_1 \leq 6 \norm{(\theta - \theta^*)_{\cS^*}}_1 \leq 6 \sqrt{s^*} \norm{\theta - \theta^*}_2.
\end{align*}

Next, we bound the objective error in \eqref{eqn:objbd}. We have
\begin{align*}
&\cF_{\lambda} (\theta) - \cF_{\lambda} (\theta^*) \overset{(i)}{\leq} -(\nabla \cL (\theta) + \lambda  g)^\top (\theta^* - \theta) \leq \norm{\nabla \cL (\theta) + \lambda  g}_{\infty} \norm{\theta^* - \theta}_1 \leq \frac{1}{2} \lambda \norm{\theta^* - \theta}_1  \\
&=\frac{1}{2} \lambda (\norm{(\theta^* - \theta)_{\cS^*}}_1 + \norm{(\theta^* - \theta)_{\overline{\cS}^*}}_1) \overset{(ii)}{\leq} 3 \lambda \norm{(\theta^* - \theta)_{\cS^*}}_1 \leq 3 \lambda \sqrt{s^*}\norm{(\theta^* - \theta)_{\cS^*}}_2 \overset{(iii)}{\leq} \frac{6 \lambda^2 s^*}{\rho^{-}_{s^*+2\tilde{s}}},
\end{align*}
where (i) is from the convexity of $\cF_{\lambda} (\theta)$ with subgradient $\nabla \cL (\theta) + \lambda  g$, (ii) is from \eqref{eqn:l1Scbd}, and (iii) is from \eqref{eqn:l2bd}.

\subsection{Proof of Lemma~\ref{lem:step2}}\label{pf:lem:step2}

Recall that $\rho^{-}$ can be either $\rho^{-}_{s^*+\tilde{s}}$ or $\rho^{-}_{s^*+2\tilde{s}}$. Assumption $\cF_{\lambda} (\theta) - \cF_{\lambda} (\theta^*) \leq {6 \lambda^2 s^*}/{\rho^{-}}$ implies
\begin{align}
\cL (\theta) - \cL (\theta^*) + \lambda(\norm{\theta}_1 - \norm{\theta^*}_1) \leq \frac{6 \lambda^2 s^*}{\rho^{-}}. \label{eqn:objbd1}
\end{align}
We have from the RSC property that 
\begin{align}
\cL(\theta) &\geq \cL(\theta^*) + (\theta - \theta^*)^\top \nabla \cL(\theta^*) + \frac{\rho^{-}}{2} \norm{\theta - \theta^*}_2^2, \label{eqn:rsc4}
\end{align}
Then we have \eqref{eqn:objbd1} and \eqref{eqn:rsc4},
\begin{align}
&\frac{\rho^{-}}{2} \norm{\theta - \theta^*}_2^2 {\leq} \frac{6 \lambda^2 s^*}{\rho^{-}} - (\theta - \theta^*)^\top \nabla \cL(\theta^*) + \lambda(\norm{\theta^*}_1 - \norm{\theta}_1). \label{eqn:l2bd3}
\end{align}
Besides, we have
\begin{align}
&(\theta - \theta^*)^\top \nabla \cL (\theta^*) \geq -\norm{(\theta - \theta^*)_{\cS^*}}_1 \norm{\cL (\theta^*)}_{\infty} - \norm{(\theta - \theta^*)_{\overline{\cS}^*}}_1 \norm{\cL (\theta^*)}_{\infty},~~\text{and} \label{eqn:l2bd4}\\
&\norm{\theta^*}_1 - \norm{\theta}_1 = \norm{\theta_{\cS^*}^*}_1 - \norm{\theta_{\cS^*}}_1 - \norm{(\theta - \theta^*)_{\overline{\cS}^*}}_1 \leq \norm{(\theta - \theta^*)_{\cS^*}}_1 - \norm{(\theta - \theta^*)_{\overline{\cS}^*}}_1. \label{eqn:l2bd5}
\end{align}
Combining \eqref{eqn:l2bd3}, \eqref{eqn:l2bd4} and \eqref{eqn:l2bd5}, we have
\begin{align}
&\frac{\rho^{-}}{2} \norm{\theta - \theta^*}_2^2 \leq \frac{6 \lambda^2 s^*}{\rho^{-}} + (\norm{\nabla \cL(\theta^*)}_{\infty} +\lambda)\norm{(\theta - \theta^*)_{\cS^*}}_1 + (\norm{\nabla \cL(\theta^*)}_{\infty} - \lambda) \norm{(\theta - \theta^*)_{\overline{\cS}^*}}_1. \label{eqn:l2bd6}
\end{align}
We discuss two cases as following:

\noindent\textbf{Case 1}. We first assume $\norm{\theta - \theta^*}_1 \leq \frac{12 \lambda s^*}{\rho^{-}}$. Then \eqref{eqn:l2bd6} implies 
\begin{align*}
&\frac{\rho^{-}}{2} \norm{\theta - \theta^*}_2^2 \overset{(i)}{\leq} \frac{6 \lambda^2 s^*}{\rho^{-}} + (\norm{\nabla \cL(\theta^*)}_{\infty} +\lambda)\norm{(\theta - \theta^*)_{\cS^*}}_1 \overset{(ii)}{\leq} \frac{6 \lambda^2 s^*}{\rho^{-}} + \frac{3}{2}\lambda \norm{(\theta - \theta^*)_{\cS^*}}_1 \leq  \frac{24 \lambda^2 s^*}{\rho^{-}}. 
\end{align*}
where (i) is from $\norm{\nabla \cL(\theta^*)}_{\infty} - \lambda \leq 0$ and (ii) is from $\norm{\nabla \cL(\theta^*)}_{\infty} + \lambda \leq \frac{3}{2}\lambda$. This indicates
\begin{align}
\norm{\theta - \theta^*}_2 \leq \frac{4\sqrt{3s^*}\lambda}{\rho^{-}}. \label{eqn:l2bdcase1}
\end{align}

\noindent\textbf{Case 2}. Next, we assume $\norm{\theta - \theta^*}_1 > \frac{12 \lambda s^*}{\rho^{-}}$. Then \eqref{eqn:l2bd6} implies 
\begin{align}
&\frac{\rho^{-}}{2} \norm{\theta - \theta^*}_2^2 \nonumber\\
&\leq (\norm{\nabla \cL(\theta^*)}_{\infty} +\lambda)\norm{(\theta - \theta^*)_{\cS^*}}_1 + (\norm{\nabla \cL(\theta^*)}_{\infty} - \lambda) \norm{(\theta - \theta^*)_{\overline{\cS}^*}}_1 + \frac{1}{2} \lambda \norm{\theta - \theta^*}_1 \nonumber\\
&= (\norm{\nabla \cL(\theta^*)}_{\infty} +\frac{3}{2}\lambda)\norm{(\theta - \theta^*)_{\cS^*}}_1 + (\norm{\nabla \cL(\theta^*)}_{\infty} - \frac{1}{2}\lambda) \norm{(\theta - \theta^*)_{\overline{\cS}^*}}_1 \nonumber \\
&\overset{(i)}{\leq}  2\lambda \norm{(\theta - \theta^*)_{\cS^*}}_1 \leq 2\sqrt{s^*}\lambda \norm{(\theta - \theta^*)_{\cS^*}}_2, \label{eqn:l2bd7}
\end{align}
where (i) is from $\norm{\nabla \cL(\theta^*)}_{\infty} +\frac{3}{2}\lambda \leq 2\lambda$ and $\norm{\nabla \cL(\theta^*)}_{\infty} - \frac{1}{2}\lambda \leq 0$. This indicates
\begin{align}
\norm{\theta - \theta^*}_2 \leq \frac{4\sqrt{s^*}\lambda}{\rho^{-}}. \label{eqn:l2bdcase2}
\end{align}
Besides, we have
\begin{align}
&\norm{\theta - \theta^*}_1 \overset{(i)}{\leq} 6 \norm{(\theta - \theta^*)_{\cS^*}}_1 \leq 6 \sqrt{s^*}\norm{(\theta - \theta^*)_{\cS^*}}_2 \leq \frac{24 \lambda s^*}{\rho^{-}}, \label{l1bd2}
\end{align}
where (i) is from $\norm{\nabla \cL(\theta^*)}_{\infty} +\frac{3}{2}\lambda \leq 2\lambda$ and \eqref{eqn:l2bd7}. 

Combining \eqref{eqn:l2bdcase1} and \eqref{eqn:l2bdcase2}, we have desired result \eqref{eqn:l2bd2}. Combining the assumption in Case 1 and \eqref{l1bd2}, we have desired result \eqref{eqn:l1bd2}.

\subsection{Proof of Lemma~\ref{lem:step3}}\label{pf:lem:step3}

Recall that the proximal-gradient update can be computed by the soft-thresholding operation,
\begin{align*}
\left(\cT_{L,\lambda}(\theta)\right)_i = \sgn(\check{\theta}_{i})\max \left\{ |\check{\theta}_{i}|-\lambda/{L}, 0 \right\} ~\forall i=1,\ldots,d,
\end{align*}
where $\check{\theta} = \theta - \nabla \cL(\theta)/{L}$. To bound $\norm{\left(\cT_{L,\lambda}(\theta)\right)_{\overline{\cS}^*}}_0$, we consider
\begin{align*}
\check{\theta} = \theta - \frac{1}{L} \nabla \cL(\theta) = \theta - \frac{1}{L} \nabla \cL(\theta^*) + \frac{1}{L} \left(\nabla \cL(\theta^*) - \nabla \cL(\theta)\right).
\end{align*}
We then consider the following three events:\\ 
\begin{align}
A_1 &= \left\{ i \in \overline{\cS}^* : |\theta_{i}| \geq {\lambda}/({3L}) \right\}, \label{eqn:a1}\\
A_2 &= \left\{ i \in \overline{\cS}^* : |(\nabla \cL(\theta^*)/L)_i| > \lambda/(6L) \right\}, \label{eqn:a2}\\ 
A_3 &= \left\{ i \in \overline{\cS}^* : \left| \left(\nabla \cL(\theta^*)/L - \nabla \cL(\theta)/L\right)_i \right| \geq \lambda/(2L) \right\}, \label{eqn:a3}
\end{align}

\noindent\textbf{Event} $A_1$. Note that for any $i \in \overline{\cS}^*$, $|\theta_{i}| = |\theta_{i} - \theta_{i}^*|$, then we have
\begin{align}
|A_1| &\leq \sum_{i \in \overline{\cS}^*} \frac{3L}{\lambda} |\theta_{i} - \theta_{i}^*| \cdot \mathds{1}(|\theta_{i} - \theta_{i}^*|\geq {\lambda}/({3L})) \leq \frac{3L}{\lambda} \sum_{i \in \overline{\cS}^*} |\theta_{i} - \theta_{i}^*| \leq \frac{3L}{\lambda} \norm{\theta - \theta^*}_1 \overset{(i)}{\leq} \frac{72 L s^* }{\rho^{-}_{s^*+2\tilde{s}}},\label{eqn:a1bd}
\end{align}
where (i) is from \eqref{eqn:l1bd2} in Lemma~\ref{lem:step2}. 

\noindent\textbf{Event} $A_2$. By  Lemma~\ref{lem:rsc}, we have
\begin{align}
0 &\leq |A_2| \leq \sum_{i \in \overline{\cS}^*} \frac{6L}{\lambda} |(\nabla \cL(\theta^*)/L)_i| \cdot \mathds{1}(|(\nabla \cL(\theta^*)/L)_i| > \lambda/(6L)) = \sum_{i \in \overline{\cS}^*} \frac{6L}{\lambda} |(\nabla \cL(\theta^*)/L)_i| \cdot 0 = 0,\label{eqn:a2bd}
\end{align}
which indicates that $|A_2| = 0$. 

\noindent\textbf{Event} $A_3$. Consider the event $\tilde{A} = \left\{ i : \left| \left(\nabla \cL(\theta^*) - \nabla \cL(\theta)\right)_i \right| \geq \lambda/2 \right\}$, which satisfies $A_3 \subseteq \tilde{A}$. We will provide an upper bound of $|\tilde{A}|$, which is also an upper bound of $|A_3|$. Let $ v \in \RR^d$ be chosen such that, $v_i = \sgn\left\{ \left(\nabla \cL(\theta^*)/L - \nabla \cL(\theta)/L\right)_i \right\}$ for any $i \in \tilde{A}$, and $v_i = 0$ for any $i \notin \tilde{A}$. Then we have
\begin{align}
 v^\top \left(\nabla \cL(\theta^*) - \nabla \cL(\theta)\right) &= \sum_{i \in \tilde{A}} v_i \left(\nabla \cL(\theta^*)/L - \nabla \cL(\theta)/L\right)_i = \sum_{i \in \tilde{A}} \left| \left(\nabla \cL(\theta^*) - \nabla \cL(\theta)\right)_i \right| \geq \lambda |\tilde{A}| /2. \label{eqn:a3bd2}
\end{align}
On the other hand, we have
\begin{align}
& v^\top \left(\nabla \cL(\theta^*) - \nabla \cL(\theta)\right) \leq \norm{ v}_2 \norm{\nabla \cL(\theta^*) - \nabla \cL(\theta)}_2 \notag\\
&\overset{(i)}{\leq} \sqrt{|\tilde{A}|} \cdot \norm{\nabla \cL(\theta^*) - \nabla \cL(\theta)}_2 \overset{(ii)}{\leq} \rho^{+}_{s^*+2\tilde{s}} \sqrt{|\tilde{A}|} \cdot \norm{\theta - \theta^*}_2, \label{eqn:a3bd3}
\end{align}
where (i) is from $\norm{ v}_2 \leq \sqrt{|\tilde{A}|} \max\{i:|A_i|\} \leq \sqrt{|\tilde{A}|}$, and (ii) is from \eqref{eqn:rsc2} and \eqref{eqn:rsc3}. 

Combining \eqref{eqn:a3bd2} and \eqref{eqn:a3bd3}, we have
\begin{align*}
\lambda |\tilde{A}| \leq 2\rho^{+}_{s^*+2\tilde{s}} \sqrt{|\tilde{A}|} \cdot \norm{\theta - \theta^*}_2 \overset{(i)}{\leq} {8\lambda {\kappa}_{s^*+2\tilde{s}}\sqrt{3 s^*|\tilde{A}|}}
\end{align*}
where (i) is from \eqref{eqn:l2bd2} in Lemma~\ref{lem:step2} and definition of ${\kappa}_{s^*+2\tilde{s}} = \frac{\rho^{+}_{s^*+2\tilde{s}}}{\rho^{-}_{s^*+2\tilde{s}}}$. Considering $A_3 \subseteq \tilde{A}$, this implies
\begin{align}
|A_3| \leq |\tilde{A}| \leq 196 {\kappa}_{s^*+2\tilde{s}}^2 s^*.  \label{eqn:a3bd}
\end{align}

Now combining Even $A_1$, $A_2$, $A_3$ and $L \leq 2\rho^{+}_{s^*+2\tilde{s}}$ in assumption, we close the proof as
\begin{align*}
\norm{\left(\cT_{L,\lambda}(\theta)\right)_{\overline{\cS}^*}}_0 &\leq |A_1| + |A_2| + |A_3| \leq \frac{72 L s^* }{\rho^{-}_{s^*+2\tilde{s}}} + 196 {\kappa}_{s^*+2\tilde{s}}^2 s^* \leq (144{\kappa}_{s^*+2\tilde{s}} + 196 {\kappa}_{s^*+2\tilde{s}}^2)s^* \leq \tilde{s}.
\end{align*}

\subsection{Proof of Lemma~\ref{lem:objbd}}\label{pf:lem:objbd}

Let $ g = \argmin_{ g \in \partial \norm{\theta}_1} \cL + \lambda \norm{\theta}_1$, then $\omega_{\lambda}  = \norm{\nabla \cL + \lambda  g}_{\infty}$. By the optimality of $\overline{\theta}$ and convexity of $\cF_{\lambda}$, we have
\begin{align}
\cF_{\lambda} (\theta) - \cF_{\lambda} (\overline{\theta}) &\leq \left( \nabla \cL + \lambda  g \right)^\top (\theta - \overline{\theta}) \leq \norm{\nabla \cL + \lambda  g}_{\infty} \norm{\theta - \overline{\theta}}_1 \leq \left( \omega_{\lambda}(\theta) \right) \norm{\theta - \overline{\theta}}_1. \label{eqn:objbd2}
\end{align}
Besides, we have
\begin{align}
\norm{\theta - \overline{\theta}}_1 & \leq \norm{\theta - \theta^*}_1 + \norm{\overline{\theta} - \theta^*}_1 \overset{(i)}{\leq} 6 \left( \norm{(\theta - \theta^*)_{\cS^*}}_1 + \norm{(\overline{\theta} - \theta^*)_{\cS^*}}_1 \right) \nonumber \\
& \leq 6 \sqrt{s^*} \left( \norm{(\theta - \theta^*)_{\cS^*}}_2 + \norm{(\overline{\theta} - \theta^*)_{\cS^*}}_2 \right) \overset{(ii)}{\leq} \frac{24 \lambda {s^*} }{\rho^{-}_{s^*+2\tilde{s}}}. \label{eqn:l1bd3}
\end{align}
where (i) and (ii) are from \eqref{eqn:l1Scbd} and \eqref{eqn:l2bd} in Lemma~\ref{lem:step1} respectively. Combining \eqref{eqn:objbd2} and \eqref{eqn:l1bd3}, we have desired result.

\subsection{Proof of Lemma~\ref{lem:controlsp}}\label{pf:lem:controlsp}

Our analysis has two steps. In the first step, we show that $\{ \theta^{(t)} \}_{t=0}^\infty$ converges to the unique limit point $\overline{\theta}$. In the second step, we show that the proximal gradient method has linear convergence rate. 

\noindent\textbf{Step 1}. Note that $\theta^{(t+1)} = \cT_{L_{},\lambda}(\theta^{(t)})$. Since $\cF_{\lambda}(\theta)$ is convex in $\theta$ (but not strongly convex), the sub-level set $\{ \theta : \cF_{\lambda}(\theta) \leq \cF_{\lambda}(\theta^{(0)}) \}$ is bounded. By the monotone decrease of $\cF_{\lambda}(\theta^{(t)})$ from \eqref{eqn:proxobjbd1} in Lemma~\ref{lem:suf_des_obj}, $\{ \theta^{(t)} \}_{t=0}^\infty$ is also bounded. By Bolzano$–$Weierstrass theorem, it has a convergent subsequence and we will show that $\overline{\theta}$ is the unique accumulation point. 

Since $\cF_{\lambda(\theta)}$ is bounded below, 
\begin{align*}
&\lim_{k \rightarrow \infty} \norm{\theta^{(t+1)} - \theta^{(t)}}_2 \leq \frac{2}{L_{}^{(t)}} \cdot \lim_{k \rightarrow \infty} \left[ \cF_{\lambda} \left( \theta^{(t+1)} \right) - \cF_{\lambda} \left( \theta^{(t)} \right) \right] = 0.
\end{align*}
By Lemma~\ref{lem:upbd_omg}, we have $\lim_{k \rightarrow \infty} \omega_{\lambda}(\theta^{(t)}) = 0$. This implies $\lim_{k \rightarrow \infty} \theta^{(t)}$ satisfies the KKT condition, hence is an optimal solution. 

Let $\overline{\theta}$ be an accumulation point. Since $\overline{\theta} = \argmin_{\theta} \cF_{\lambda}(\theta)$, then there exists some $ g \in \partial \norm{\overline{\theta}}_1$ such that
\begin{align}
\nabla \cF_{\lambda}(\overline{\theta}) = \cL_{ \lambda}(\overline{\theta}) + \lambda  g =  0. \label{eqn:optloss}
\end{align}
By Lemma~\ref{lem:step3}, every proximal update is sparse, hence $\norm{\overline{\theta}_{\overline{\cS}^*}}_0 \leq \tilde{s}$. By RSC property in \eqref{eqn:rscrss}, if $\norm{\theta_{\overline{\cS}^*}}_0 \leq \tilde{s}$, i.e.,$\norm{(\theta-\overline{\theta})_{\overline{\cS}^*}}_0 \leq \tilde{s}$ , then we have
\begin{align}
\cL(\theta) - \cL(\overline{\theta}) \geq(\theta - \overline{\theta})^\top \nabla \cL(\overline{\theta}) + \frac{\rho^{-}_{s^*+2\tilde{s}}}{2} \norm{\theta - \overline{\theta}}_2^2, \label{eqn:rsc7}
\end{align}
From the convexity of $\norm{\theta}_1$ and $ g \in \partial \norm{\overline{\theta}}_1$, we have
\begin{align}
\norm{\theta}_1 - \norm{\overline{\theta}}_1 \geq (\theta - \overline{\theta})^\top  g. \label{eqn:optreg}
\end{align}
Combining \eqref{eqn:rsc7} and \eqref{eqn:optreg}, we have for any $\norm{\theta_{\overline{\cS}^*}}_0 \leq \tilde{s}$,
\begin{align}
\cF_{\lambda}(\theta) - \cF_{\lambda}(\overline{\theta}) &= \cL(\theta) + \lambda \norm{\theta}_1 - \left(\cL(\overline{\theta}) - \lambda \norm{\overline{\theta}}_1 \right) \geq (\theta-\overline{\theta})^\top \left( \cL(\overline{\theta}) + \lambda  g \right) + \frac{\rho^{-}_{s^*+2\tilde{s}}}{2} \norm{\theta - \overline{\theta}}_2^2 \nonumber\\
&\overset{(i)}{=} \frac{\rho^{-}_{s^*+2\tilde{s}}}{2} \norm{\theta - \overline{\theta}}_2^2 \geq 0, \label{eqn:optaccu}
\end{align}
where (i) is from \eqref{eqn:optloss}. Therefore, $\overline{\theta}$ is the unique accumulation point, i.e. $\lim_{k \rightarrow \infty} \theta^{(t)} = \overline{\theta}$. 

\noindent\textbf{Step 2}. The objective $\cF_{\lambda}(\theta^{(t+1)})$ satisfies
\begin{align}
&\cF_{\lambda}(\theta^{(t+1)}) \overset{(i)}{\leq} \cQ_{\lambda} \left( \theta^{(t+1)},\theta^{(t)} \right) \overset{(ii)}{=} \min_{\theta} \cL (\theta^{(t)}) + \nabla \cL (\theta^{(t)})^\top (\theta - \theta^{(t)}) + \frac{\tilde{L}_{\lambda}^{(t)}}{2} \norm{\theta - \theta^{(t)}}_2^2 + \lambda \norm{\theta}_1. \label{eqn:proxbd1}
\end{align}
where (i) is from \eqref{eqn:proxobjbd1} in Lemma~\ref{lem:suf_des_obj}, (ii) is from the definition of $\cO_{\lambda}$ in \eqref{eqn:prox}. To further bound R.H.S. of \eqref{eqn:proxbd1}, we consider the line segment $S(\overline{\theta}, \theta^{(t)}) = \{ \theta: \theta = \alpha \overline{\theta} + (1-\alpha) \theta^{(t)}, \alpha \in [0,1] \}$. Then we restrict the minimization over the line segment $S(\overline{\theta}, \theta^{(t)})$,
\begin{align}
&\cF_{\lambda}(\theta^{(t+1)}) - \cL (\theta^{(t)}) \leq \min_{\theta \in S(\overline{\theta}, \theta^{(t)})} \nabla \cL (\theta^{(t)})^\top (\theta - \theta^{(t)}) + \frac{\tilde{L}_{\lambda}^{(t)}}{2} \norm{\theta - \theta^{(t)}}_2^2 + \lambda \norm{\theta}_1. \label{eqn:proxbd2}
\end{align}
Since $\norm{\overline{\theta}_{\overline{\cS}^*}}_0 \leq \tilde{s}$ and $\norm{\theta^{(t)}_{\overline{\cS}^*}}_0 \leq \tilde{s}$, then for any $\theta \in S(\overline{\theta}, \theta^{(t)})$, we have $\norm{\theta_{\overline{\cS}^*}}_0 \leq \tilde{s}$ and $\norm{(\theta-\theta^{(t)})_{\overline{\cS}^*}}_0 \leq 2\tilde{s}$. By RSC property, we have
\begin{align}
\cL(\theta) &\geq \cL (\theta^{(t)}) + \nabla \cL (\theta^{(t)})^\top (\theta - \theta^{(t)}) + \frac{\rho^{-}_{s^*+2\tilde{s}}}{2} \norm{\theta - \theta^{(t)}}_2^2 \geq \cL (\theta^{(t)}) + \nabla \cL (\theta^{(t)})^\top (\theta - \theta^{(t)}). \label{eqn:proxbd3}
\end{align}
Combining \eqref{eqn:proxbd2} and \eqref{eqn:proxbd3}, we have
\begin{align}
&\cF_{\lambda}(\theta^{(t+1)}) \nonumber \\
&\leq \min_{\theta \in S(\overline{\theta}, \theta^{(t)})} \cL(\theta) + \frac{\tilde{L}_{\lambda}^{(t)}}{2} \norm{\theta - \theta^{(t)}}_2^2 + \lambda \norm{\theta}_1 \nonumber \\
&= \min_{\alpha \in [0,1]} \cF_{\lambda}(\alpha \overline{\theta} + (1-\alpha) \theta^{(t)}) + \frac{\alpha^2 \tilde{L}_{\lambda}^{(t)}}{2} \norm{\overline{\theta} - \theta^{(t)}}_2^2 \nonumber \\
&\overset{(i)}{\leq} \min_{\alpha \in [0,1]} \alpha\cF_{\lambda}(\overline{\theta}) + (1-\alpha) \cF_{\lambda}(\theta^{(t)}) + \frac{\alpha^2 \tilde{L}_{\lambda}^{(t)}}{2} \norm{\overline{\theta} - \theta^{(t)}}_2^2 \nonumber \\
&\overset{(ii)}{\leq} \min_{\alpha \in [0,1]} \cF_{\lambda}(\theta^{(t)}) - \alpha \left( \cF_{\lambda}(\theta^{(t)}) - \cF_{\lambda}(\overline{\theta}) \right) + \frac{\alpha^2 \tilde{L}_{\lambda}^{(t)}}{\rho^{-}_{s^*+2\tilde{s}}} \left( \cF_{\lambda}(\theta^{(t)}) - \cF_{\lambda}(\overline{\theta}) \right) \nonumber \\
&= \min_{\alpha \in [0,1]} \cF_{\lambda}(\theta^{(t)}) - \alpha \left( 1 - \frac{\alpha \tilde{L}_{\lambda}^{(t)}}{\rho^{-}_{s^*+2\tilde{s}}} \right) \left( \cF_{\lambda}(\theta^{(t)}) - \cF_{\lambda}(\overline{\theta}) \right), \label{eqn:proxbd4}
\end{align}
where (i) is from the convexity of $\cF_{\lambda}$ and (ii) is from \eqref{eqn:optaccu}. 

Minimize the R.H.S. of \eqref{eqn:proxbd4} w.r.t. $\alpha$, the optimal value $\alpha = \frac{\rho^{-}_{s^*+2\tilde{s}}}{2\tilde{L}_{\lambda}^{(t)}}$ results in 
\begin{align}
\cF_{\lambda}(\theta^{(t+1)}) \leq \cF_{\lambda}(\theta^{(t)}) - \frac{\rho^{-}_{s^*+2\tilde{s}}}{4\tilde{L}_{\lambda}^{(t)}} \left( \cF_{\lambda}(\theta^{(t)}) - \cF_{\lambda}(\overline{\theta}) \right). \label{eqn:proxbd5}
\end{align}
Subtracting both sides of \eqref{eqn:proxbd5} by $\cF_{\lambda}(\overline{\theta})$, we have
\begin{align}
&\cF_{\lambda}(\theta^{(t+1)}) - \cF_{\lambda}(\overline{\theta}) \leq \left( 1 - \frac{\rho^{-}_{s^*+2\tilde{s}}}{4\tilde{L}_{\lambda}^{(t)}} \right) \left( \cF_{\lambda}(\theta^{(t)}) - \cF_{\lambda}(\overline{\theta}) \right) \overset{(i)}{\leq} \left( 1 - \frac{\rho^{-}_{s^*+2\tilde{s}}}{8\rho^{+}_{s^*+2\tilde{s}}} \right) \left( \cF_{\lambda}(\theta^{(t)}) - \cF_{\lambda}(\overline{\theta}) \right), \label{eqn:proxbd6}
\end{align}
where (i) is from Remark~\ref{remark:stepsize}. Apply \eqref{eqn:proxbd6} recursively, we have the desired result.

\section{Proof of Intermediate Results for Theorem~\ref{thm:proxnewton}}

We also introduce an important notion as follows, which is closely related with the SE properties.
\begin{definition}\label{def:LRE}
	We denote the local $\ell_1$ cone as
	\begin{align*}
	&\cC(s, vartheta,r) {=} \big\{  v,\theta : \cS \subseteq \cM, |\cM| \leq s, \|  v_{\Mc} \|_1 \leq  vartheta \|  v{\cM} \|_1, \\
	&\hspace{.8in} \| \theta - \theta^* \|_2 \leq r \big\}.
	\end{align*}
	Then we define the largest and smallest \textbf{localized restricted eigenvalues} (LRE) as
	\begin{align*}
	\psi^+_{s, vartheta,r} &= \sup_{u,\theta} \left\{ \frac{ v^\top \nabla^2 \cL(\theta)  v}{ v^\top  v} : ( v,\theta) \in \cC(s, vartheta,r) \right\},\\
	\psi^-_{s, vartheta,r} &= \inf_{u,\theta} \left\{ \frac{ v^\top \nabla^2 \cL(\theta)  v}{ v^\top  v} : ( v,\theta) \in \cC(s, vartheta,r) \right\}.
	\end{align*}
\end{definition}

The following proposition demonstrates the relationships between SE and LRE. The proof can be found in \cite{buhlmann2011statistics}, thus is omitted here.
\begin{proposition}\label{prop:se_lre}
	Given any $\theta,\theta' \in \cC(s, vartheta,r) \cap \cB(\theta^*,r)$, we have
	\begin{align*}
	c_1 \psi^-_{s, vartheta,r} \leq \rho^-_{s} \leq c_2 \psi^-_{s, vartheta,r} ,~~\text{and}~~c_3 \psi^+_{s, vartheta,r} \leq \rho^+_{s} \leq c_4 \psi^+_{s, vartheta,r}.
	\end{align*}
	where $c_1$, $c_2$, $c_3$, and $c_4$ are constants.
\end{proposition}

\subsection{Proof of Lemma~\ref{lem:sparse_preserve}}\label{pf:lem:sparse_preserve}

We first demonstrate the sparsity of the update. Since $\theta^{(t+1)}$ is the minimizer to the proximal Newton problem, we have
\begin{align*}
\nabla^2{\cL}(\theta^{(t)})(\theta^{(t+1)}-\theta^{(t)}) + \nabla\cL(\theta^{(t)}) + \lambda \xi^{(t+1)}=0,
\end{align*}
where $\xi^{(t+1)}\in\partial\norm{\theta^{(t+1)}}_1$.

It follows from \cite{fan2015tac} that if conditions in Lemma~\ref{lem:rsc} holds, then we have $\min_{j \in \Sc'} \{\lambda_j\} \geq \lambda/2$ for some set $\cS' \supset \cS$ with $|\cS'| \leq 2s^*$. Then the analysis of sparsity of can be performed through $\lambda$ directly. 

We then consider the following decomposition
\begin{align*}
&\nabla^2{\cL}(\theta^{(t)})(\theta^{(t+1)}-\theta^{(t)}) + \nabla\cL(\theta^{(t)})\\
&=\underbrace{\nabla^2{\cL}(\theta^{(t)})(\theta^{(t+1)}-\theta^*)}_{V_1} +\underbrace{\nabla^2{\cL}(\theta^{(t)})(\theta^*-\theta^{(t)})}_{V_2} + \underbrace{\nabla\cL(\theta^{(t)})-\nabla\cL(\theta^*)}_{V_3} + \underbrace{\nabla\cL(\theta^*)}_{V_4}.
\end{align*}

Consider the following sets: $\cA_i = \left\{ j \in \Sc' ~:~ |(V_i)_j| \geq \lambda/4 \right\}$, for all $i \in \{1,2,3,4\}$.

\noindent\textbf{Set $\cA_2$.}  Suppose we choose a vector $v \in \RR^d$ such that $v_j = \sign\left\{ (\nabla^2{\cL}(\theta^{(t)})(\theta^*-\theta^{(t)}))_j \right\}$ for all $j \in \cA_2$ and $v_j=0$ for $j \notin \cA_2$. Then we have
\begin{align}
v^\top \nabla^2{\cL}(\theta^{(t)})(\theta^*-\theta^{(t)}) &= \sum_{j \in \cA_2} v_j (\nabla^2{\cL}(\theta^{(t)})(\theta^*-\theta^{(t)}))_j = \sum_{j \in \cA_2} |(\nabla^2{\cL}(\theta^{(t)})(\theta^*-\theta^{(t)}))_j|  \geq \lambda |\cA_2| /4. \label{eqn:lb_V2}
\end{align}
On the other hand, we have
\begin{align}
&v^\top \nabla^2{\cL}(\theta^{(t)})(\theta^*-\theta^{(t)}) \leq \|v(\nabla^2{\cL}(\theta^{(t)}))^{1/2}\|_2 \|(\nabla^2{\cL}(\theta^{(t)}))^{1/2}(\theta^*-\theta^{(t)}) \|_2 \nonumber \\
&\overset{(i)}{\leq} \rho^+_{s^*+2\tilde{s}} \|v\|_2 \| \theta^*-\theta^{(t)} \|_2  \overset{(ii)}{\leq} \sqrt{|\cA_2|} \rho^+_{s^*+2\tilde{s}} \| \theta^*-\theta^{(t)} \|_2 \overset{(iii)}{\leq} C' \sqrt{|\cA_2|} \kappa_{s^*+2\tilde{s}} \lambda \sqrt{s^*}, \label{eqn:ub_V2}
\end{align}
where $(i)$ is from the SE properties, $(ii)$ is from the definition of $v$, and $(iii)$ is from $\| \theta^{(t)} - \theta^* \|_2 \leq C' \lambda \sqrt{s^*}/\rho^-_{s^*+2\tilde{s}}$. Combining \eqref{eqn:lb_V2} and \eqref{eqn:ub_V2}, we have $|\cA_2| \leq C_2 \kappa^2_{s^*+2\tilde{s}} s^*$.

\noindent\textbf{Set $\cA_3$.} Consider the event $\tilde{A} = \left\{ i : \left| \left(\nabla \cL(\theta^{(t)}) - \nabla \cL(\theta^*)\right)_i \right| \geq \lambda/4 \right\}$, which satisfies $\cA_3 \subseteq \tilde{A}$. We will provide an upper bound of $|\tilde{A}|$, which is also an upper bound of $|\cA_3|$. Let $v \in \RR^d$ be chosen such that $v_i = \sgn\left\{ \left(\nabla \cL(\theta^{(t)}) - \nabla \cL(\theta^*)\right)_i \right\}$ for any $i \in \tilde{A}$, and $v_i = 0$ for any $i \notin \tilde{A}$. Then we have
\begin{align}
&v^\top \left(\nabla \cL(\theta^{(t)}) - \nabla \cL(\theta^*)\right) = \sum_{i \in \tilde{A}} v_i \left(\nabla \cL(\theta^{(t)}) - \nabla \cL(\theta^*)\right)_i = \sum_{i \in \tilde{A}} \left| \left(\nabla \cL(\theta^{(t)}) - \nabla \cL(\theta^*)\right)_i \right| \geq \lambda |\tilde{A}| /4. \label{eqn:a3bd22}
\end{align}
On the other hand, we have
\begin{align}
&v^\top \left(\nabla \cL(\theta^{(t)}) - \nabla \cL(\theta^*)\right) \leq \norm{v}_2 \norm{\nabla \cL(\theta^{(t)}) - \nabla \cL(\theta^*)}_2\nonumber\\
& \overset{(i)}{\leq} \sqrt{|\tilde{A}|} \cdot \norm{\nabla \cL(\theta^{(t)}) - \nabla \cL(\theta^*)}_2 \overset{(ii)}{\leq} \rho^{+}_{s^*+2\tilde{s}} \sqrt{|\tilde{A}|} \cdot \norm{\theta^{(t)} - \theta^*}_2, \label{eqn:a3bd33}
\end{align}
where $(i)$ is from $\norm{v}_2 \leq \sqrt{|\tilde{A}|} \max\{i:|\cA_i|\} \leq \sqrt{|\tilde{A}|}$, and $(ii)$ is from the mean value theorem and the SE properties. 

Combining \eqref{eqn:a3bd22} and \eqref{eqn:a3bd33}, we have
\begin{align*}
\lambda |\tilde{A}| \leq 4\rho^{+}_{s^*+2\tilde{s}} \sqrt{|\tilde{A}|} \cdot \norm{\theta - \theta^*}_2 \overset{(i)}{\leq} {8\lambda {\kappa}_{s^*+2\tilde{s}}\sqrt{3 s^*|\tilde{A}|}}
\end{align*}
where $(i)$ is from $\| \theta^{(t)} - \theta^* \|_2 \leq C' \lambda \sqrt{s^*}/\rho^-_{s^*+2\tilde{s}}$ and definition of ${\kappa}_{s^*+2\tilde{s}} = \rho^{+}_{s^*+2\tilde{s}}/\rho^{-}_{s^*+2\tilde{s}}$. Considering $\cA_3 \subseteq \tilde{A}$, this implies $|\cA_3| \leq |\tilde{A}| \leq C_3 {\kappa}_{s^*+2\tilde{s}}^2 s^*$.

\noindent\textbf{Set $\cA_4$.} By  conditions in Lemma~\ref{lem:rsc}  and $\lambda \geq 4\norm{\nabla \cL(\theta^*)}_{\infty}$, we have
\begin{align*}
&0 \leq |V_4| \leq \sum_{i \in \Sc^*} \frac{4}{\lambda} |(\nabla \cL(\theta^*))_i| \cdot \mathds{1}(|(\nabla \cL(\theta^*))_i| > \lambda/(4)) = \sum_{i \in \Sc^*} \frac{4}{\lambda} |(\nabla \cL(\theta^*))_i| \cdot 0 = 0,
\end{align*}

\noindent\textbf{Set $\cA_1$.} 
From Lemma~\ref{lem:obj_bd}, we have $\cF_{\lambda}(\theta^{(t+1)}) \leq \cF_{\lambda}(\theta^*) + \frac{\lambda}{4} \norm{\theta^{(t+1)} - \theta^*}_1$. This implies
\begin{align}
&\cL(\theta^{(t+1)}) - \cL(\theta^*) \leq \lambda(\norm{\theta^*}_1 - \norm{\theta^{(t+1)}}_1)  + \frac{\lambda}{4} \norm{\theta^{(t+1)} - \theta^*}_1 \nonumber \\
&= \lambda(\norm{\theta^*_{\cS'}}_1 - \norm{\theta^{(t+1)}_{\cS'}}_1 - \norm{\theta^{(t+1)}_{\cS'_{\perp}}}_1) + \frac{\lambda}{4} \norm{\theta^{(t+1)} - \theta^*}_1 \nonumber \\
&\le \frac{5\lambda}{4} \norm{\theta^{(t+1)}_{\cS'} - \theta^*_{\cS'}}_1 - \frac{3\lambda}{4} \norm{\theta^{(t+1)}_{\cS'_{\perp}}  - \theta^*_{\cS'_{\perp}}}_1. \label{eqn:L_ub}
\end{align}
where the equality holds since $\theta^*_{\cS'_{\perp}}=0$. On the other hand, we have
\begin{align}
&\cL(\theta^{(t+1)}) - \cL(\theta^*) \overset{(i)}{\ge} \nabla \cL(\theta^*) (\theta^{(t+1)} - \theta^*) \ge -\norm{cL(\theta^*)}_{\infty} \norm{\theta^{(t+1)} - \theta^*}_1 \overset{(ii)}{\ge} -\frac{\lambda}{4}\norm{\theta^{(t+1)} - \theta^*}_1 \nonumber\\
&= -\frac{\lambda}{4} \norm{\theta^{(t+1)}_{\cS'} - \theta^*_{\cS'}}_1 - \frac{\lambda}{4} \norm{\theta^{(t+1)}_{\cS'_{\perp}}  - \theta^*_{\cS'_{\perp}}}_1,\label{eqn:L_lb}
\end{align}
where $(i)$ is from the convexity of $\cL$ and $(ii)$ is from conditions of Lemma~\ref{lem:rsc}. Combining \eqref{eqn:L_ub} and \eqref{eqn:L_lb}, we have
\begin{align*}
\norm{\theta^{(t+1)}_{\cS'_{\perp}}  - \theta^*_{\cS'_{\perp}}}_1 \leq 3 \norm{\theta^{(t+1)}_{\cS'} - \theta^*_{\cS'}}_1,
\end{align*}
which implies that $(\theta^{(t+1)}  - \theta^*, \theta^{(t+1)}) \in \cC(s^*,3,r)$ with respect to the set $\cS'$. Then we choose a vector $v \in \RR^d$ such that $v_j = \sign\left\{ (\nabla^2{\cL}(\theta^{(t)})(\theta^{(t+1)} - \theta^*))_j \right\}$ for all $j \in \cA_1$ and $v_j=0$ for $j \notin \cA_1$. Then we have
\begin{align}
&v^\top \nabla^2{\cL}(\theta^{(t)})(\theta^{(t+1)} - \theta^*) = \sum_{j \in \cA_2} v_j (\nabla^2{\cL}(\theta^{(t)})(\theta^{(t+1)} - \theta^*))_j \notag\\
&= \sum_{j \in \cA_2} |(\nabla^2{\cL}(\theta^{(t)})(\theta^{(t+1)} - \theta^*))_j| \geq \lambda |\cA_1| /4. \label{eqn:lb_V1}
\end{align}
On the other hand, we have
\begin{align}
&v^\top \nabla^2{\cL}(\theta^{(t)})(\theta^{(t+1)} - \theta^*) \leq \|v(\nabla^2{\cL}(\theta^{(t)}))^{1/2}\|_2 \|(\nabla^2{\cL}(\theta^{(t)}))^{1/2}(\theta^{(t+1)} - \theta^*) \|_2 \nonumber \\
&\overset{(i)}{\leq} c_1 \rho^+_{s^*+2\tilde{s}} \|v\|_2 \| \theta^{(t+1)} - \theta^* \|_2 \overset{(ii)}{\leq} c_1 \sqrt{|\cA_2|} \rho^+_{s^*+2\tilde{s}} \| \theta^{(t+1)} - \theta^* \|_2 \nonumber \\
&\overset{(iii)}{\leq} c_2 \sqrt{|\cA_1|} \kappa_{s^*+2\tilde{s}} \lambda \sqrt{s^*}, \label{eqn:ub_V1}
\end{align}
where $(i)$ is from the SE properties and Proposition~\ref{prop:se_lre}, $(ii)$ is from the definition of $v$, and $(iii)$ is from $\| \theta^{(t+1)} - \theta^* \|_2 \leq C' \lambda \sqrt{s^*}/\rho^-_{s^*+2\tilde{s}}$. Combining \eqref{eqn:lb_V1} and \eqref{eqn:ub_V1}, we have $|\cA_1| \leq C_1 \kappa^2_{s^*+2\tilde{s}} s^*$.



Combining the results for Set $\cA_1 \sim \cA_4$, we have that there exists some constant $C_0$ such that
\begin{align*}
\norm{\theta^{(t+1)}_{\Sc}}_0 \leq C_0 \kappa^2_{s^*+2\tilde{s}} s^* \leq \tilde{s}.
\end{align*}
This finishes the first part. The estimation error follows directly from Lemma~\ref{lem:est_bd}.

\subsection{Proof of Lemma~\ref{lem:quadratic_rate}} \label{pf:lem:quadratic_rate}

For notational simplicity, we introduce the following proximal operator,
\begin{align*}
\prox_{r}^{H, g}(\theta) = \textrm{argmin}_{\theta'} r(\theta') + g^{\top}(\theta' - \theta)+ \frac{1}{2}\norm{\theta'-\theta}_H^2.
\end{align*}
Then we have
\begin{align*}
\theta^{(t+1)} = \prox_{\cR^{\ell_1}_{\lambda}(\theta^{(t)})}^{\nabla^2 \cL\rbr{\theta^{(t)}}, \nabla \cL\rbr{\theta^{(t)}}}\rbr{\theta^{(t)}}.
\end{align*}
By Lemma~\ref{lem:sparse_preserve}, we have 
\begin{align*}
\norm{\theta^{(t+1)}_{\Sc}}_0\le \tilde{s}.
\end{align*}
By the KKT condition of function $\min \cF_{\lambda}$, i.e., $-\nabla \cL(\overline{\theta}) \in \partial \cR^{\ell_1}_{\lambda} (\overline{\theta})$, we also have
\begin{align*}
\overline{\theta} = \prox_{\cR^{\ell_1}_{\lambda}(\overline{\theta}) }^{\nabla^2\cL\rbr{\theta^{(t)}}, \nabla\cL(\overline{\theta})}\rbr{\overline{\theta}}.
\end{align*}

By monotonicity of sub-gradient of a convex function, we have the \emph{strictly non-expansive} property: for any $\theta, \theta'\in \RR$, let $u =  \prox_{r}^{H, g}(\theta)$ and 
$v= \prox_{r}^{H, g'}(\theta')$, then
\begin{align*}
\rbr{u-v}^{\top} H(\theta-\theta') - \rbr{u-v}^{\top}\rbr{g-g'}\ge \nbr{u-v}_H^2.
\end{align*}
Thus by the strictly non-expansive property of the proximal operator, we obtain
{
	\begin{align}
	&\norm{\theta^{(t+1)} - \overline{\theta}}_{\nabla^2\cL(\overline{\theta})}^2 \le \rbr{\theta^{(t+1)} - \overline{\theta}}^{\top}\sbr{\nabla^2\cL(\theta^{(t)})\rbr{\theta^{(t)} - \overline{\theta}} + \rbr{\nabla\cL(\overline{\theta}) -\nabla\cL(\theta^{(t)})}} \nonumber\\
	&\le \norm{\theta^{(t+1)} - \overline{\theta}}_2 \Big\| \nabla^2\cL(\theta^{(t)})\rbr{\theta^{(t)} - \overline{\theta}} + \rbr{\nabla\cL(\overline{\theta}) -\nabla\cL(\theta^{(t)})} \Big\|_2.\label{eqn:Hnorm_ub}
	\end{align}}
Note that both $\norm{\theta^{(t+1)}}_0 \leq \tilde{s}$ and $\norm{\overline{\theta}}_0 \leq \tilde{s}$. On the other hand, from the SE properties, we have
\begin{align}
\norm{\theta^{(t+1)} - \overline{\theta}}_{\nabla^2\cL(\overline{\theta})}^2 = (\theta^{(t+1)} - \overline{\theta})^\top \nabla^2\cL(\overline{\theta}) (\theta^{(t+1)} - \overline{\theta}) \geq \rho^-_{s^*+2\tilde{s}} \norm{\theta^{(t+1)} - \overline{\theta}}_2^2. \label{eqn:Hnorm_lb}
\end{align}
Combining \eqref{eqn:Hnorm_ub} and \eqref{eqn:Hnorm_lb}, we have
{
	\begin{align*}
	&\rho^-_{s^*+2\tilde{s}}\nbr{\theta^{(t+1)} - \overline{\theta}}_2 \le \nbr{\nabla^2\cL(\theta^{(t)})\rbr{\theta^{(t)} - \overline{\theta}} + \rbr{\nabla\cL(\overline{\theta}) -\nabla\cL(\theta^{(t)})}}_2 \\
	&=
	\nbr{
		\int_0^1 
		\sbr{
			\nabla^2\cL\rbr{\theta^{(t)} + \tau \rbr{\overline{\theta} - \theta^{(t)}}} - 
			\nabla^2\cL\rbr{\theta^{(t)}}}
		\cdot\rbr{\overline{\theta}-\theta^{(t)}}
		d \tau
	}_2 \\
	&\le \int_0^1 
	\nbr{
		\sbr{
			\nabla^2\cL\rbr{\theta^{(t)} + \tau \rbr{\overline{\theta} - \theta^{(t)}}} - 
			\nabla^2\cL\rbr{\theta^{(t)}}}
		\cdot\rbr{\overline{\theta}-\theta^{(t)}}
	}_2 d \tau \\
	&\le L_{s^*+2\tilde{s}} \nbr{\theta^{(t)} - \overline{\theta}}_2^2,
	\end{align*}}
where the last inequality is from the local restricted Hessian smoothness of $\cL$. Then we finish the proof by the definition of $r$.

\subsection{Proof of Lemma~\ref{lem:unit_step}} \label{pf:lem:unit_step}

Suppose the step size $\eta_{t}<1$. Note that we do not need the step size to be $\eta_{t}=1$ in Lemma~\ref{lem:sparse_preserve} and Lemma~\ref{lem:quadratic_rate}. We denote $\Delta \theta^{(t)} = \theta^{(t+1/2)} - \theta^{(t)}$. Then we have
\begin{align}
\nbr{\Delta \theta^{(t)}}_2 &\overset{(i)}{\le} \nbr{\theta^{(t)} - \overline{\theta}}_2  + \nbr{\theta^{(t+1/2)} - \overline{\theta}}_2 \overset{(ii)}{\le}  \nbr{\theta^{(t)} - \overline{\theta}}_2  +  \frac{L_{s^*+2\tilde{s}}}{2\rho^-_{s^*+2\tilde{s}}}\nbr{\theta^{(t)} - \overline{\theta}}_2^2 \overset{(iii)}{\le} \frac{3}{2}\nbr{\theta^{(t)} - \overline{\theta}}_2, \label{eqn:deltatheta}
\end{align}
where $(i)$ is from triangle inequality, $(ii)$ is from Lemma~\ref{lem:quadratic_rate}, and $(iii)$ is from $\nbr{\theta^{(t)} - \overline{\theta}}_2 \le r \le \frac{\rho^-_{s^*+2\tilde{s}}}{L_{s^*+2\tilde{s}}}$.

By  Lemma~\ref{lem:sparse_preserve}, we have $\nbr{{\Delta\theta^{(t)}}_{\Sc}}_0\le 2\tilde{s}$. To show $\eta_{t}=1$, it is now suffice to demonstrate that
\begin{align*}
\cF_{\lambda}(\theta^{(t+1/2)}) - \cF_{\lambda}(\theta^{(t)}) \le \frac{1}{4} \gamma_t.
\end{align*}
By expanding $\cF_{\lambda}$, 
we have
\begin{align*}
&\cF_{\lambda}(\theta^{(t)} + \Delta \theta^{(t)}) - \cF_{\lambda}(\theta^{(t)}) =\cL(\theta^{(t)}+\Delta \theta^{(t)}) -\cL(\theta^{(t)}) + \cR^{\ell_1}_{\lambda}(\theta^{(t)}+\Delta \theta^{(t)}) - \cR^{\ell_1}_{\lambda}(\theta^{(t)}) \\
& \overset{(i)}{\le} \nabla\cL(\theta^{(t)})^{\top} \Delta \theta^{(t)} + \frac{1}{2}\Delta (\theta^{(t)})^{\top} \nabla^2\cL(\theta) \Delta \theta^{(t)} + \frac{L_{s^*+2\tilde{s}}}{6}\nbr{\Delta \theta^{(t)}}_2^3 + \cR^{\ell_1}_{\lambda}(\theta^{(t)}+\Delta \theta^{(t)}) - \cR^{\ell_1}_{\lambda}(\theta^{(t)}) \\
& \overset{(ii)}{\le} \gamma_t - \frac{1}{2}\gamma_t + \frac{L_{s^*+2\tilde{s}}}{6}\nbr{\Delta \theta^{(t)}}_2^3  \overset{(iii)}{\le} \frac{1}{2}\gamma_t + \frac{L_{s^*+2\tilde{s}}}{6 \rho^-_{s^*+2\tilde{s}}}\nbr{\Delta \theta^{(t)}}_{\nabla^2\cL(\theta)}\nbr{\Delta \theta^{(t)}}_2 \overset{(iv)}{\le} \rbr{\frac{1}{2} - \frac{L_{s^*+2\tilde{s}}}{6 \rho^-_{s^*+2\tilde{s}}}\nbr{\Delta \theta^{(t)}}_2}\gamma_t  \\
&\overset{(v)}{\le} \frac{1}{4}\gamma_t,
\end{align*}
where $(i)$ is from the restricted Hessian smooth condition, $(ii)$ and $(iv)$ are from Lemma~\ref{lem:prop_lamb}, $(iii)$ is from the same argument of \eqref{eqn:Hnorm_lb}, and $(v)$ is from \eqref{eqn:deltatheta}, $\gamma_t < 0$, and $\nbr{\theta^{(t)} - \overline{\theta}}_2 \le r \le \frac{\rho^-_{s^*+2\tilde{s}}}{L_{s^*+2\tilde{s}}}$. This implies $\theta^{(t+1)}=\theta^{(t+1/2)}$.

\subsection{Proof of Lemma~\ref{lem:prop_lamb}} \label{pf:lem:prop_lamb}

We denote $H=\nabla^2 \cL(\theta^{(t)})$.  Since $\Delta \theta^{(t)}$ is the solution for 
\begin{align*}
\min_{\Delta \theta^{(t)}}  \nabla\cL\rbr{\theta^{(t)}}^{\top}\cdot \Delta \theta^{(t)} + \frac{1}{2} \nbr{\Delta \theta^{(t)}}_{H}^2+\cR^{\ell_1}_{\lambda}\rbr{\theta^{(t)} + \Delta \theta^{(t)}}
\end{align*}
then for any $\eta_t\in(0, 1]$, we have
\begin{align*}
&\eta_t\nabla\cL\rbr{\theta^{(t)}}^{\top}\cdot \Delta \theta^{(t)} + \frac{\eta_t^2}{2} \nbr{\Delta \theta^{(t)}}_{H}^2+\cR^{\ell_1}_{\lambda}\rbr{\theta^{(t)} + \eta_t\Delta \theta^{(t)}} \\
&\ge \nabla\cL\rbr{\theta^{(t)}}^{\top}\cdot \Delta \theta^{(t)} + \frac{1}{2} \nbr{\Delta \theta^{(t)}}_{H}^2+\cR^{\ell_1}_{\lambda}\rbr{\theta^{(t)} + \Delta \theta^{(t)}}
\end{align*}
By the convexity of $\cR^{\ell_1}_{\lambda}$, we have
\begin{align*}
&\eta_t\nabla\cL\rbr{\theta^{(t)}}^{\top}\cdot \Delta \theta^{(t)} + \frac{\eta_t^2}{2} \nbr{\Delta \theta^{(t)}}_{H}^2+ \eta_t \cR^{\ell_1}_{\lambda}\rbr{\theta^{(t)} + \Delta \theta^{(t)}} + (1-\eta_t) \cR^{\ell_1}_{\lambda}(\theta^{(t)}) \\
&\ge \nabla\cL\rbr{\theta^{(t)}}^{\top}\cdot \Delta \theta^{(t)} + \frac{1}{2} \nbr{\Delta \theta^{(t)}}_{H}^2+\cR^{\ell_1}_{\lambda}\rbr{\theta^{(t)} + \Delta \theta^{(t)}}.
\end{align*}
Rearranging the terms, we obtain
\begin{align*}
&(1-\eta_t)\rbr{\nabla\cL\rbr{\theta^{(t)}}^{\top}\cdot \Delta \theta^{(t)} + \cR^{\ell_1}_{\lambda}\rbr{\theta^{(t)}-\Delta \theta^{(t)}} - \cR^{\ell_1}_{\lambda}(\theta^{(t)})} \le - \frac{1-\eta_t^2}{2}\nbr{\Delta \theta^{(t)}}_H^2 
\end{align*}
Canceling the $(1-\eta_t)$ factor from both sides and let $\eta_t\rightarrow 1$, we obtain the desired inequality,
\begin{align*}
\gamma_t \le -\nbr{\Delta \theta^{(t)}}_H^2.\qedhere
\end{align*}

\subsection{Proof of Lemma~\ref{lem:approx_kkt}}\label{pf:lem:approx_kkt}

We first demonstrate an upper bound of the approximate KKT parameter $\omega_{\lambda}$. Given the solution $\theta^{(t-1)}$ from the $(t-1)$-th iteration, the optimal solution at $t$-th iteration satisfies the KKT condition:
\begin{align*}
\nabla^2{\cL}(\theta^{(t-1)})(\theta^{(t)}-\theta^{(t-1)}) + \nabla\cL(\theta^{(t-1)}) + \lambda \xi^{(t)}=0,
\end{align*}
where $\xi^{(t)}\in\partial\norm{\theta^{(t)}}_1$. Then for any vector $v$ with $\norm{v}_2 \leq \norm{v}_1 = 1$ and $\norm{v}_0 \leq s^*+2\tilde{s}$, by taking $\Delta \theta^{(t-1)} = \theta^{(t)} - \theta^{(t-1)}$, we have
{
	\begin{align}
	&(\nabla \cL(\theta^{(t)}) + \lambda \xi^{(t)})^{\top} v \nonumber \\
	&= (\nabla \cL(\theta^{(t)}) - \nabla^2{\cL}(\theta^{(t-1)})\Delta \theta^{(t-1)} - \nabla\cL(\theta^{(t-1)}))^{\top} v \nonumber \\
	& = (\nabla \cL(\theta^{(t)}) - \nabla\cL(\theta^{(t-1)}))^{\top} v - (\nabla^2{\cL}(\theta^{(t-1)})\Delta \theta^{(t-1)} )^{\top} v \nonumber \\
	&\overset{(i)}{\le} \nbr{(\nabla^2{\cL}(\tilde{\theta}))^{1/2}\Delta \theta^{(t-1)}}_2 \hspace{-.07in}\cdot\hspace{-.03in} \nbr{v^{\top} (\nabla^2{\cL}(\tilde{\theta}))^{1/2}}_2 + \nbr{(\nabla^2{\cL}(\theta^{(t-1)}))^{1/2}\Delta \theta^{(t-1)}}_2 \hspace{-.07in}\cdot\hspace{-.03in} \nbr{v^{\top} (\nabla^2{\cL}(\theta^{(t-1)}))^{1/2}}_2 \nonumber \\
	&\overset{(ii)}{\le} 2\rho^+_{s^*+2\tilde{s}} \nbr{ \Delta \theta^{(t-1)} }_2, \label{eqn:approx_kkt_ub1}
	\end{align}
}
where $(i)$ is from mean value theorem with some $\tilde{\theta} = (1-a) \theta^{(t-1)} + a \theta^{(t)}$ for some $a \in [0,1]$ and Cauchy-Schwarz inequality, and $(ii)$ is from the SE properties. Take the supremum of the L.H.S. of \eqref{eqn:approx_kkt_ub1} with respect to $v$, we have
\begin{align}
\nbr{\nabla \cL(\theta^{(t)}) + \lambda \xi^{(t)}}_{\infty} \leq 2\rho^+_{s^*+2\tilde{s}} \nbr{ \Delta \theta^{(t-1)} }_2. \label{eqn:approx_kkt_ub2}
\end{align}
Then from Lemma~\ref{lem:quadratic_rate}, we have
\begin{align*}
\nbr{\theta^{(t+1)} - \overline{\theta}}_2 &\le \rbr{\frac{L_{s^*+2\tilde{s}}}{2\rho^-_{s^*+2\tilde{s}}}}^{1+2+4+\ldots +2^{t-1}}\nbr{\theta^{(0)} - \overline{\theta}}_2^{2^\top } \le \rbr{\frac{L_{s^*+2\tilde{s}}}{2\rho^-_{s^*+2\tilde{s}}}\nbr{\theta^{(0)} - \overline{\theta}}_2}^{2^t }.
\end{align*}
By \eqref{eqn:approx_kkt_ub2} and \eqref{eqn:deltatheta}, we obtain
\begin{align*}
&\omega_{\lambda} \rbr{\theta^{(t)}} \leq 2\rho^+_{s^*+2\tilde{s}} \nbr{ \Delta \theta^{(t-1)} }_2 \leq 3\rho^+_{s^*+2\tilde{s}} \nbr{ \theta^{(t-1)} - \overline{\theta} }_2 \le 3\rho^+_{s^*+2\tilde{s}}\rbr{\frac{L_{s^*+2\tilde{s}}}{2\rho^-_{s^*+2\tilde{s}}}\nbr{\theta^{(0)} - \overline{\theta}}_2}^{2^t }.
\end{align*}
By requiring the R.H.S. equal to $\varepsilon$ we obtain
\begin{align*}
t&=\log \frac{\log\rbr{\frac{3\rho^+_{s^*+2\tilde{s}}}{\varepsilon}}}{\log \rbr{\frac{2\rho^-_{s^*+2\tilde{s}}}{L_{s^*+2\tilde{s}} \nbr{\theta^{(0)} - \overline{\theta}}_2}}}  = \log \log\rbr{\frac{3\rho^+_{s^*+2\tilde{s}}}{\varepsilon}} - \log \log \rbr{\frac{2\rho^-_{s^*+2\tilde{s}}}{L_{s^*+2\tilde{s}} \nbr{\theta^{(0)} - \overline{\theta}}_2}} \\
&\overset{(i)}{\leq} \log \log\rbr{\frac{3\rho^+_{s^*+2\tilde{s}}}{\varepsilon}} - \log \log 4 \leq \log \log\rbr{\frac{3\rho^+_{s^*+2\tilde{s}}}{\varepsilon}} ,
\end{align*}
where $(i)$ is from the fact that $\nbr{\theta^{(0)} - \overline{\theta}}_2 \leq r = \frac{\rho^-_{s^*+2\tilde{s}}}{2 L_{s^*+2\tilde{s}}}$.
\begin{lemma}\label{lem:obj_bd}
	Given $\omega_{\lambda} ({\theta}^{(t)}) \leq \frac{\lambda}{4}$, we have
	\begin{align*}
	\cF_{\lambda}(\theta^{(t)}) \leq \cF_{\lambda}(\theta^*) + \frac{\lambda}{4} \norm{\theta^{(t)} - \theta^*}_1.
	\end{align*}
\end{lemma}
\begin{proof}
	For some $\xi^{(t)} = \argmin_{\xi \in \partial \norm{\theta^{(t)}}_1} \norm{\nabla \cL(\theta^{(t)}) + \lambda \xi}_{\infty}$, we have
	\begin{align*}
	\cF_{\lambda} (\theta^*) &\overset{(i)}{\geq} \cF_{\lambda}(\theta^{(t)}) - (\nabla \cL(\theta^{(t)}) + \lambda \xi^{(t)})^{\top} (\theta^{(t)} - \theta^*) \geq \cF_{\lambda}(\theta^{(t)}) - \norm{\nabla \cL(\theta^{(t)}) + \lambda \xi^{(t)}}_{\infty} \norm{\theta^{(t)} - \theta^*}_{1} \\
	&\overset{(ii)}{\geq} \cF_{\lambda}(\theta^{(t)}) - \frac{\lambda}{4} \norm{\theta^{(t)} - \theta^*}_{1}
	\end{align*}
	where $(i)$ is from the convexity of $\cF_{\lambda}$ and $(ii)$ is from the fact that for all $t \geq 0$, $\cF_{\lambda} (\theta^{(t)}) \leq \cF_{\lambda} (\theta^{(t-1)})$ and $\omega_{\lambda} ({\theta}^{(t)}) \leq \frac{\lambda}{4}$. This finishes the proof.
\end{proof}

\begin{lemma}[Adapted from \cite{fan2015tac}]\label{lem:est_bd}
	Suppose $\norm{\theta^{(t)}_{\Sc}}_0\leq \tilde{s}$ and  $\omega_{\lambda} (\theta^{(t)}) \leq \frac{\lambda}{4}$. Then there exists a generic constant $c_1$ such that $\norm{\theta^{(t)}-\theta^*}_2\leq 
	\frac{c_1\lambda \sqrt{s^*}}{\rho^-_{s^*+2\tilde{s}}}$.
\end{lemma}

\section{Proofs of Intermediate Lemmas in Appendix~\ref{int:thm:intrsc2}}\label{pf:int:thm:intrsc2}

\subsection{Proof of Lemma~\ref{lem:sublin_to_linv2}}\label{pf:lem:sublin_to_linv2}

\textbf{Part 1}. We first show $\norm{\theta - \theta^*}_2^2 \leq r$ by contradiction. Suppose $\norm{\theta - \theta^*}_2 > \sqrt{r}$. Let $\alpha \in [0,1]$ such that $\tilde{\theta} = (1-\alpha)\theta + \alpha \theta^*$ and
\begin{align}
\norm{\tilde{\theta} - \theta^*}_2 = \sqrt{r}. \label{eqn:assuption_theta}
\end{align}

Let $\tilde{g} = \argmin_{g \in \partial \norm{\theta}_1}  \norm{\nabla \cL (\theta) + \lambda g}_{\infty}$ and $\Delta = \theta - \theta^*$, then we have
\begin{align}
\cF_{\lambda}(\theta^*) &\overset{(i)}{\geq} \cF_{\lambda}(\theta) - (\nabla \cL (\theta) + \lambda \tilde{g})^\top \Delta \geq \cF_{\lambda}(\theta) - \norm{\nabla \cL (\theta) + \lambda \tilde{g}}_{\infty} \norm{\Delta}_1 \nonumber \\
&\overset{(ii)}{\geq} \cF_{\lambda}(\theta) - \frac{\lambda}{4} \norm{\Delta}_1, \label{eqn:obj_ub}
\end{align}
where $(i)$ is from the convexity of $\cF_{\lambda}(\theta)$ and $(ii)$ is from the approximate KKT condition. 

Denote $\tilde{\Delta} = \tilde{\theta} - \theta^*$. Combining \eqref{eqn:obj_ub} and \eqref{eqn:assuption_theta}, we have
\begin{align*}
\cF_{\lambda}(\tilde{\theta}) &\overset{(i)}{\leq} (1-\alpha) \cF_{\lambda}(\theta) + \alpha \cF_{\lambda}(\theta^*) \leq (1-\alpha)\cF_{\lambda}(\theta^*) + \frac{(1-\alpha)\lambda}{4} \norm{\Delta}_1 + \alpha \cF_{\lambda}(\theta^*) \\
&\leq \cF_{\lambda}(\theta^*) + \frac{\lambda}{4} \norm{(1-\alpha)(\theta - \theta^*)}_1 = \cF_{\lambda}(\theta^*) + \frac{\lambda}{4} \norm{(1-\alpha)\theta + \alpha \theta^* - \theta^*)}_1 \\
&=  \cF_{\lambda}(\theta^*) + \frac{\lambda}{4} \norm{\tilde{\theta} - \theta^*}_1 = \cF_{\lambda}(\theta^*) + \frac{\lambda}{4} \norm{\tilde{\Delta}}_1.
\end{align*}
where $(i)$ is from the convexity of $\cF_{\lambda}(\theta)$. This indicates 
\begin{align}
\cL(\tilde{\theta}) - \cL(\theta^*) &\leq \lambda (\norm{\theta^*}_1 - \norm{\tilde{\theta}}_1 + \frac{1}{4} \norm{\tilde{\Delta}}_1) \nonumber\\
&= \lambda (\norm{\theta^*_{\cS^*}}_1 - \norm{\tilde{\theta}_{\cS^*}}_1 - \norm{\tilde{\theta}_{\overline{\cS}^*}}_1 + \frac{1}{4} \norm{\tilde{\Delta}_{\cS^*}}_1 + \frac{1}{4} \norm{\tilde{\Delta}_{\overline{\cS}^*}}_1) \nonumber \\
&\leq \lambda(\norm{\theta^*_{\cS^*} - \tilde{\theta}_{\cS^*}}_1 - \norm{\tilde{\theta}_{\overline{\cS}^*} - \theta^*_{\overline{\cS}^*}}_1 + \frac{1}{4} \norm{\tilde{\Delta}_{\cS^*}}_1 + \frac{1}{4} \norm{\tilde{\Delta}_{\overline{\cS}^*}}_1) \nonumber\\
&= \frac{5\lambda}{4} \norm{\tilde{\Delta}_{\cS^*}}_1 - \frac{3\lambda}{4} \norm{\tilde{\Delta}_{\overline{\cS}^*}}_1. \label{eqn:loss_ub}
\end{align}
On the other hand, we have
\begin{align}
\cL(\tilde{\theta}) - \cL(\theta^*) &\overset{(i)}{\geq} \nabla \cL(\theta^*)^\top \tilde{\Delta} \geq - \norm{\nabla \cL(\theta^*)}_{\infty} \norm{\tilde{\Delta}}_1 \overset{(ii)}{\geq} -\frac{\lambda}{6} \norm{\tilde{\Delta}}_1 \nonumber\\
&= -\frac{\lambda}{6} \norm{\tilde{\Delta}_{\cS^*}}_1 - \frac{\lambda}{6} \norm{\tilde{\Delta}_{\overline{\cS}^*}}_1, \label{eqn:loss_lb}
\end{align}
where $(i)$ is from the convexity of $\cL(\theta)$, $(ii)$ is from Lemma~\ref{lem:rsc}. Combining \eqref{eqn:loss_ub} and \eqref{eqn:loss_lb}, we have
\begin{align}
\norm{\tilde{\Delta}_{\overline{\cS}^*}}_1 \leq \frac{5}{2} \norm{\tilde{\Delta}_{\cS^*}}_1. \label{eqn:cone}
\end{align}

Next, we consider the following sequence of sets:
\begin{align*}
\cS_0 &= \left\{ j \in \overline{\cS}^* : \sum_{m \in \overline{\cS}^*} \mathds{1}(|\tilde{\theta}_m| \geq |\tilde{\theta}_j|) \leq \tilde{s} \right\}~~\text{and}~~\\
\cS_i &= \left\{ j \in \overline{\cS}^* \backslash \bigcup_{k<i}\cS_k : \sum_{m \in \overline{\cS}^* \backslash \bigcup_{k<i}\cS_k} \mathds{1}(|\tilde{\theta}_m| \geq |\tilde{\theta}_j|) \leq \tilde{s} \right\}~\text{for all}~i=1,2,\ldots.
\end{align*}
We introduce a result from \cite{buhlmann2011statistics} with its proof provided therein.
\begin{lemma}[Adapted from Lemma 6.9 in \cite{buhlmann2011statistics} by setting $q=2$]\label{lem:orderseq}
	Let $v = [v_1, v_2,\ldots]^\top$ with $v_1 \geq v_2 \geq \ldots \geq 0$. For any $s \in \{ 1,2,\ldots \}$, we have
	\begin{align*}
	\left( \sum_{j\geq s+1} v_j^2 \right)^{1/2} \leq \sum_{k=1}^{\infty} \left( \sum_{j= ks+1}^{(k+1)s} v_j^2 \right)^{1/2} \leq \frac{\norm{v}_1}{\sqrt{s}}.
	\end{align*}
\end{lemma}
Denote $\cA = \cS^* \cup \cS_0$. Then we have
\begin{align}
\sum_{i \geq 1} \norm{\tilde{\Delta}_{\cS_i}}_2 \overset{(i)}{\leq} \frac{1}{\sqrt{\tilde{s}}} \norm{\tilde{\Delta}_{\overline{\cS}^*}}_1 \overset{(ii)}{\leq} \frac{5}{2} \sqrt{\frac{s^*}{\tilde{s}}} \norm{\tilde{\Delta}_{\cS^*}}_2 \leq \frac{5}{2} \sqrt{\frac{s^*}{\tilde{s}}} \norm{\tilde{\Delta}_{\cA}}_2, \label{eqn:sp_l2bd}
\end{align}
where $(i)$ is rom Lemma~\ref{lem:orderseq} with $s=\tilde{s}$ and $(ii)$ is from \eqref{eqn:cone}. Let $\check{\theta} = (1-\beta) \tilde{\theta} + \beta \theta^*$ for any $\beta \in [0,1]$. Then we have
\begin{align*}
\norm{\check{\theta} - \theta^*}_2  = (1-\beta)\norm{\tilde{\theta} - \theta^*}_2 \leq \sqrt{r}, 
\end{align*}
which implies $\cL(\check{\theta})$ satisfies RSC/RSS for $\check{\theta}$ restricted on a sparse set by Lemma~\ref{lem:rsc}. Then we have
\begin{align}
|\tilde{\Delta}_{\overline{\cA}}^\top \nabla_{\overline{\cA},\cA} \cL(\check{\theta}) \tilde{\Delta}_{\cA}| &\leq \sum_{i \geq 1} |\tilde{\Delta}_{\cS_i}^\top \nabla_{\cS_i,\cA} \cL(\check{\theta}) \tilde{\Delta}_{\cA}| \leq \rho^{+}_{s^*+\tilde{s}} \norm{\tilde{\Delta}_{\cA}}_2 \sum_{i \geq 1} \norm{\tilde{\Delta}_{\cS_i}}_2 \nonumber\\
&\overset{(i)}{\leq} \frac{5}{2} \sqrt{\frac{s^*}{\tilde{s}}} \rho^{+}_{s^*+\tilde{s}} \norm{\tilde{\Delta}_{\cA}}_2^2, \label{eqn:rssbd1}
\end{align}
where $(i)$ is from \eqref{eqn:sp_l2bd}. On the other hand, we have from RSC
\begin{align}
\tilde{\Delta}_{\cA}^\top \nabla_{\cA,\cA} \cL(\check{\theta}) \tilde{\Delta}_{\cA} \geq \rho^{-}_{s^*+\tilde{s}} \norm{\tilde{\Delta}_{\cA}}_2^2. \label{eqn:rscbd1}
\end{align}
Then we have w.h.p.
\begin{align*}
\tilde{\Delta} \nabla \cL(\check{\theta}) \tilde{\Delta} &= \tilde{\Delta}_{\cA}^\top \nabla_{\cA,\cA} \cL(\check{\theta}) \tilde{\Delta}_{\cA} + 2\tilde{\Delta}_{\overline{\cA}}^\top \nabla_{\overline{\cA},\cA} \cL(\check{\theta}) \tilde{\Delta}_{\cA} + \tilde{\Delta}_{\overline{\cA}}^\top \nabla_{\overline{\cA},\overline{\cA}} \cL(\check{\theta}) \tilde{\Delta}_{\overline{\cA}} \\
&\geq \tilde{\Delta}_{\cA}^\top \nabla_{\cA,\cA} \cL(\check{\theta}) \tilde{\Delta}_{\cA} - 2|\tilde{\Delta}_{\overline{\cA}}^\top \nabla_{\overline{\cA},\cA} \cL(\check{\theta}) \tilde{\Delta}_{\cA}| \\
&\overset{(i)}{\geq} \left( \rho^{-}_{s^*+\tilde{s}} - 5 \sqrt{\frac{s^*}{\tilde{s}}} \rho^{+}_{s^*+\tilde{s}} \right) \norm{\tilde{\Delta}_{\cA}}_2^2 \overset{(ii)}{\geq} \frac{9}{14} \rho^{-}_{s^*+\tilde{s}} \norm{\tilde{\Delta}_{\cA}}_2^2, 
\end{align*}
where $(i)$ is from \eqref{eqn:rssbd1} and \eqref{eqn:rscbd1}, $(ii)$ is from Lemma~\ref{lem:rsc}. This implies
\begin{align}
\cL(\tilde{\theta}) - \cL(\theta^*) &= \nabla \cL(\theta^*)^\top \tilde{\Delta} + \frac{1}{2} \tilde{\Delta} \nabla \cL(\check{\theta}) \tilde{\Delta} \geq \nabla \cL(\theta^*)^\top \tilde{\Delta} + \frac{9}{28} \rho^{-}_{s^*+\tilde{s}} \norm{\tilde{\Delta}_{\cA}}_2^2 \nonumber\\
&\overset{(i)}{\geq} \frac{9}{28} \rho^{-}_{s^*+\tilde{s}} \norm{\tilde{\Delta}_{\cA}}_2^2 -\frac{\lambda}{6} \norm{\tilde{\Delta}_{\cS^*}}_1 - \frac{\lambda}{6} \norm{\tilde{\Delta}_{\overline{\cS}^*}}_1, \label{eqn:loss_lb2}
\end{align}
where $(i)$ is from $\lambda \geq \lambda_{[N]} \geq 6\norm{\nabla \cL(\theta^*)}_\infty$. Combining \eqref{eqn:loss_ub} and \eqref{eqn:loss_lb2}, we have
\begin{align*}
\rho^{-}_{s^*+\tilde{s}} \norm{\tilde{\Delta}_{\cS^*}}_2^2 \leq \rho^{-}_{s^*+\tilde{s}} \norm{\tilde{\Delta}_{\cA}}_2^2 \leq \frac{8}{3} \lambda \norm{\tilde{\Delta}_{\cS^*}}_1 \leq \frac{8}{3} \lambda \sqrt{s^*} \norm{\tilde{\Delta}_{\cS^*}}_2 \leq \frac{8}{3} \lambda \sqrt{s^*} \norm{\tilde{\Delta}_{\cA}}_2.
\end{align*}
This implies 
\begin{align}
\norm{\tilde{\Delta}_{\cS^*}}_2 \leq \norm{\tilde{\Delta}_{\cA}}_2 \leq \frac{8 \lambda \sqrt{s^*}}{3 \rho^{-}_{s^*+\tilde{s}}}~~\text{and}~~\norm{\tilde{\Delta}_{\cS^*}}_1 \leq \frac{8\lambda s^*}{3 \rho^{-}_{s^*+\tilde{s}}}. \label{esterr1}
\end{align}
Then we have
\begin{align}
\norm{\tilde{\Delta}_{\overline{\cA}}}_2 \leq \sum_{i \geq 1} \norm{\tilde{\Delta}_{\cS_i}}_2 \overset{(i)}{\leq} \frac{1}{\sqrt{\tilde{s}}} \norm{\tilde{\Delta}_{\overline{\cS}^*}}_1  \overset{(ii)}{\leq} \frac{5}{2} \sqrt{\frac{1}{s^*}} \norm{\tilde{\Delta}_{\cS^*}}_1 \overset{(iii)}{\leq} \frac{20 \lambda \sqrt{s^*}}{3 \rho^{-}_{s^*+\tilde{s}}}, \label{esterr2}
\end{align}
where $(i)$ is rom Lemma~\ref{lem:orderseq} with $s=\tilde{s}$, $(ii)$ is from \eqref{eqn:cone} and $\tilde{s} \geq s^*$ and $(iii)$ is from \eqref{esterr1}. Combining \eqref{esterr1} and \eqref{esterr2}, we have 
\begin{align*}
\norm{\tilde{\Delta}}_2 = \sqrt{\norm{\tilde{\Delta}_{\cA}}_2^2 + \norm{\tilde{\Delta}_{\overline{\cA}}}_2^2} \leq \frac{8 \lambda \sqrt{s^*}}{\rho^{-}_{s^*+\tilde{s}}} < \sqrt{r},
\end{align*}
where the last inequality is from the condition $\frac{\rho^{-}_{s^*+\tilde{s}}}{8}\sqrt{\frac{r}{s^*}} > \lambda$. This conflicts with \eqref{eqn:assuption_theta}, which indicates that $\norm{\theta - \theta^*}_2 \leq \sqrt{r}$. 

\noindent\textbf{Part 2}. We next demonstrate the sparsity of $\theta$. From $\lambda > \lambda_{[N]} \geq 6 \norm{\nabla \cL(\theta^*)}_{\infty}$, we have
\begin{align}
\left| \left\{ i \in \overline{\cS}^* : |\nabla_i \cL(\theta^*) | \geq \frac{\lambda}{6} \right\} \right| = 0. \label{eqn:card_set1v2}
\end{align}
Denote $\check{\cS}_1 = \left\{ i \in \overline{\cS}^* : | \nabla_i \cL(\theta) - \nabla_i \cL(\theta^*) | \geq \frac{\lambda}{2} \right\}$ and $\check{s}_1 = |\check{\cS}_1|$. Then there exists some $b \in \RR^d$ such that $\norm{b}_{\infty} = 1$, $\norm{b}_0 \leq \check{s}_1$ and $b^\top (\nabla \cL(\theta) - \nabla \cL(\theta^*)) \geq \frac{\lambda \check{s}_1}{2}$. Then by the mean value theorem, we have for some $\check{\theta} = (1-\alpha) \theta + \alpha \theta^*$ with $\alpha \in [0,1]$, $\nabla \cL(\theta) - \nabla \cL(\theta^*) = \nabla^2 \cL (\check{\theta}) \Delta$, where $\Delta = \theta - \theta^*$. Then we have
\begin{align}
\frac{\lambda \check{s}_1}{2} &\leq b^\top \nabla^2 \cL (\check{\theta}) \Delta \overset{(i)}{\leq} \sqrt{b^\top \nabla^2 \cL (\check{\theta}) b} \sqrt{\Delta^\top \nabla^2 \cL (\check{\theta}) \Delta} \nonumber\\
&\overset{(ii)}{\leq} \sqrt{\check{s}_1 \rho^{+}_{\check{s}_1} } \sqrt{\Delta^\top (\nabla \cL(\theta) - \nabla \cL(\theta^*))}, \label{eqn:sp_bd1v2}
\end{align}
where $(i)$ is from the generalized Cauchy-Schwarz inequality, $(ii)$ is from the definition of RSS and the fact that $\norm{b}_2 \leq \sqrt{\check{s}_1} \norm{b}_{\infty} = \sqrt{\check{s}_1}$. Let $g$ achieve $\min_{g \in \partial \norm{\theta}_1} \cF_{\lambda}(\theta)$. Further, we have
\begin{align}
\Delta^\top (\nabla \cL(\theta) - \nabla \cL(\theta^*)) &\leq \norm{\Delta}_1 \norm{\nabla \cL(\theta) - \nabla \cL(\theta^*)}_{\infty} \nonumber\\
&\leq \norm{\Delta}_1 (\norm{\nabla \cL(\theta^*)}_{\infty} + \norm{\nabla \cL(\theta)}_{\infty}) \nonumber\\
&\leq \norm{\Delta}_1 (\norm{\nabla \cL(\theta^*)}_{\infty} + \norm{\nabla \cL(\theta) + \lambda g}_{\infty} + \lambda\norm{g}_{\infty}) \nonumber\\
&\overset{(i)}{\leq} \frac{28 \lambda s^*}{3 \rho^{-}_{s^*+\tilde{s}}} (\frac{\lambda}{6}+\frac{\lambda}{4} + \lambda) \leq \frac{14 \lambda^2 s^*}{\rho^{-}_{s^*+\tilde{s}}}, \label{eqn:sp_bd2v2}
\end{align}
where $(i)$ is from combining \eqref{eqn:cone} and \eqref{esterr1}, condition on $\lambda$, approximate KKT condition and $\norm{g}_{\infty} \leq 1$. Combining \eqref{eqn:sp_bd1v2} and \eqref{eqn:sp_bd2v2}, we have $\frac{\sqrt{\check{s}_1}}{2} \leq \sqrt{\frac{14 \rho^{+}_{\check{s}_1} s^*}{\rho^{-}_{s^*+\tilde{s}}}}$, which further implies
\begin{align}
\check{s}_1 \leq \frac{56 \rho^{+}_{\check{s}_1} s^*}{\rho^{-}_{s^*+\tilde{s}}} \leq 56 {\kappa}_{s^*+2\tilde{s}} s^* \leq \tilde{s}. 
\end{align}
For any $v \in \RR^d$ that satisfies $\norm{v}_{\infty} \leq 1$, we have
\begin{align*}
\check{\cS}_2 = \left\{ i \in \overline{\cS}^* : \left| \nabla_i \cL (\theta) + \frac{\lambda}{6} v_i \right| \geq \frac{2 \lambda}{3} \right\} \subseteq \left\{ i \in \overline{\cS}^* : |\nabla_i \cL(\theta^*) | \geq \frac{\lambda}{6} \right\} \bigcup \check{\cS}_1 = \check{\cS}_1.
\end{align*}
Then we have $|\check{\cS}_2| \leq |\check{\cS}_1| \leq \tilde{s}$. Since for any $i \in \overline{\cS}^*$ and $\left| \nabla_i \cL (\theta) + \frac{\lambda}{6} v_i \right| < \frac{2 \lambda}{3}$, we can find $g_i$ that satisfies $|g_i| \leq 1$ such that $\nabla_i \cL (\theta) + \frac{\lambda}{6} v_i + \lambda g_i = 0$ which implies $\theta_i = 0$, then we have
\begin{align*}
\left| \left\{ i \in \overline{\cS}^* : \left| \nabla_i \cL (\theta) + \frac{\lambda}{6} v_i \right| \geq \frac{2 \lambda}{3} \right\} \right| \leq \check{s}_1.
\end{align*}
Therefore, we have $\norm{\theta_{\overline{\cS}^*}}_0 \leq |\check{\cS}_2| \leq \tilde{s}$. 

\subsection{Proof of Lemma~\ref{lem:warmst}}\label{pf:lem:warmst}

Since $\omega_{\lambda_{[K-1]}}(\hat{\theta}_{[K-1]}) \leq \lambda_{[K-1]}/4$, there exists some subgradient $g \in \partial \norm{\hat{\theta}_{[K-1]}}_1$ such that 
\begin{align}
\norm{\nabla \cL(\hat{\theta}_{[K-1]}) + \lambda_{[K-1]} g }_{\infty} \leq \lambda_{[K-1]}/4. \label{eqn:optcond1}
\end{align}
By the definition of $\omega_{\lambda_{[K]}}(\cdot)$, we have
\begin{align*}
\omega_{\lambda_{[K]}}(\hat{\theta}_{[K-1]}) &\leq \norm{\nabla \cL(\hat{\theta}_{[K-1]}) + \lambda_{[K]} g }_{\infty} = \norm{\nabla \cL(\hat{\theta}_{[K-1]}) + \lambda_{[K-1]} g + (\lambda_{[K]} - \lambda_{[K-1]}) g }_{\infty} \\
&\leq \norm{\nabla \cL(\hat{\theta}_{[K-1]}) + \lambda_{[K-1]} g }_{\infty} + |\lambda_{[K]} - \lambda_{[K-1]}| \cdot \norm{g}_{\infty} \overset{(i)}{\leq} \lambda_{[K-1]}/4 + (1-\eta_{\lambda})\lambda_{[K-1]} \\
&\overset{(ii)}{\leq} \lambda_{[K]}/2,
\end{align*}
where $(i)$ is from \eqref{eqn:optcond1} and choice of $\lambda_{[K]}$, $(ii)$ is from the condition on $\eta_{\lambda}$.

\end{document}